\documentclass[10pt]{article}

\usepackage[report]{hairline}        %
\usepackage[masthead]{frontmatter}   %
\makeatletter\def\hlPrintReportBodyLead{13.5}\makeatother
\newcommand{\gridGapAdamWThreeEmFourAbs}{0.6}

\newcommand{\gridGapAdamWTwoFourEmThreeAbs}{10.7}

\newcommand{\gridGapAdamWThreeFourEmThreeAbs}{131.1}

\newcommand{\selLrAdamW}{1.2e\ensuremath{-}3}

\newcommand{\gridGapMuonThreeEmFourAbs}{0.1}

\newcommand{\selLrMuon}{1.7e\ensuremath{-}3}

\newcommand{\testMzeroAdamW}{0.9995}

\newcommand{\gpuhMzeroAdamW}{35.4}
\newcommand{\testMzeroMuon}{0.9744}

\newcommand{\gpuhMzeroMuon}{35.6}
\newcommand{\testMthreeAdamW}{0.9905}

\newcommand{\testMthreeMuon}{0.9707}

\newcommand{\testMfourAdamW}{0.9906}

\newcommand{\testMfourMuon}{0.9710}

\newcommand{\testMzeroTwoEMuon}{0.9645}

\newcommand{\testUAdamW}{0.9926}

\newcommand{\gpuhUAdamW}{40.2}
\newcommand{\testUMuon}{0.9702}

\newcommand{\gpuhUMuon}{38.8}
\newcommand{\testRAdamW}{0.9936}

\newcommand{\testMzeroRAdamW}{0.9955}

\newcommand{\testMzeroLoopMuon}{0.9621}

\newcommand{\gapMthreeMzeroAdamW}{\ensuremath{-}9.0}
\newcommand{\gapMthreeMzeroAdamWAbs}{9.0}

\newcommand{\gapMthreeMfourAdamW}{\ensuremath{-}0.1}

\newcommand{\gapMthreeMzeroMuon}{\ensuremath{-}3.7}
\newcommand{\gapMthreeMzeroMuonAbs}{3.7}

\newcommand{\gapMthreeMfourMuon}{\ensuremath{-}0.4}

\newcommand{\gapMfourMzeroAdamWAbs}{8.9}

\newcommand{\gapMfourUAdamW}{\ensuremath{-}2.0}
\newcommand{\gapMfourUAdamWAbs}{2.0}

\newcommand{\gapMfourMzeroMuonAbs}{3.3}

\newcommand{\gapMfourUMuon}{\ensuremath{+}0.9}
\newcommand{\gapMfourUMuonAbs}{0.9}

\newcommand{\gapMsevenMfourAdamWAbs}{3.5}

\newcommand{\gapMsevenMfourMuonAbs}{0.1}

\newcommand{\gapMzeroTwoEMzeroAdamWAbs}{14.3}

\newcommand{\gapMzeroTwoEMzeroMuon}{\ensuremath{-}9.9}
\newcommand{\gapMzeroTwoEMzeroMuonAbs}{9.9}

\newcommand{\gapGMzeroAdamW}{\ensuremath{-}4.2}
\newcommand{\gapGMzeroAdamWAbs}{4.2}

\newcommand{\gapGMzeroMuonAbs}{0.9}

\newcommand{\gapUMzeroAdamW}{\ensuremath{-}6.9}
\newcommand{\gapUMzeroAdamWAbs}{6.9}

\newcommand{\gapUMzeroMuon}{\ensuremath{-}4.2}
\newcommand{\gapUMzeroMuonAbs}{4.2}

\newcommand{\gapRUAdamW}{\ensuremath{+}1.0}

\newcommand{\gapRUMuon}{\ensuremath{+}0.8}

\newcommand{\gapMzeroRMzeroAdamW}{\ensuremath{-}4.0}
\newcommand{\gapMzeroRMzeroAdamWAbs}{4.0}

\newcommand{\gapHuginnMzeroAdamWAbs}{435.6}

\newcommand{\gapSmeltVthreeMthreeAdamW}{\ensuremath{+}2.2}

\newcommand{\gapSmeltVthreeMthreeMuon}{\ensuremath{+}0.2}

\newcommand{\gapSmeltVfourMthreeAdamWAbs}{1.9}

\newcommand{\gapSmeltVfourMthreeMuonAbs}{2.0}

\newcommand{\gapSmeltVfourUMuonAbs}{2.5}

\newcommand{\gapDtwoFourUAdamWAbs}{4.7}

\newcommand{\gapDtwoFourUMuonAbs}{2.7}

\newcommand{\gapDthreeSixUAdamWAbs}{17.3}

\newcommand{\gapDthreeSixUMuonAbs}{7.0}

\newcommand{\gapMzeroLRMzeroAdamW}{\ensuremath{-}2.9}

\newcommand{\gapMzeroLRMzeroRAdamW}{\ensuremath{+}1.2}

\newcommand{\gapMzeroIMzeroAdamW}{\ensuremath{-}2.3}

\newcommand{\gapMzeroLoopMzeroAdamWAbs}{14.8}

\newcommand{\gapMzeroLoopMzeroRAdamWAbs}{10.8}

\newcommand{\gapMzeroLoopMzeroTwoEAdamWAbs}{0.5}

\newcommand{\gapMzeroLoopMzeroMuon}{\ensuremath{-}12.2}
\newcommand{\gapMzeroLoopMzeroMuonAbs}{12.2}

\newcommand{\gapMzeroLoopMzeroRMuonAbs}{12.3}

\newcommand{\gapLatentLoopResidualMthreeAdamWAbs}{1.5}

\newcommand{\gapLatentLoopResidualMthreeMuonAbs}{0.4}

\newcommand{\gapUGAdamWAbs}{2.7}

\newcommand{\gapUGMuonAbs}{3.4}

\newcommand{\gapMfourRAdamWAbs}{3.0}

\newcommand{\gapMfiveMsevenAdamWAbs}{3.9}

\newcommand{\gapMfiveMsevenMuonAbs}{1.7}

\newcommand{\gapMthreeMfiveAdamWAbs}{0.2}

\newcommand{\gapMthreeMfiveMuonAbs}{1.2}

\newcommand{\leadMzero}{\ensuremath{-}25.2}
\newcommand{\leadMzeroAbs}{25.2}

\newcommand{\leadMfourAbs}{19.6}

\newcommand{\leadDthreeSixAbs}{32.8}

\newcommand{\budParamsMthree}{\ensuremath{-}0.27}

\newcommand{\budFlopsMthree}{\ensuremath{+}2.44}

\newcommand{\budKvMthree}{\ensuremath{+}3.12}

\newcommand{\budParamsMfour}{\ensuremath{+}0.83}

\newcommand{\budFlopsMfour}{\ensuremath{+}2.33}

\newcommand{\budKvMfour}{\ensuremath{+}3.12}

\newcommand{\budParamsG}{\ensuremath{+}0.01}

\newcommand{\budFlopsG}{\ensuremath{+}0.02}

\newcommand{\budKvG}{0.00}

\newcommand{\budParamsU}{\ensuremath{-}0.27}

\newcommand{\budFlopsU}{\ensuremath{+}2.33}

\newcommand{\budKvU}{\ensuremath{+}3.12}

\newcommand{\budParamsR}{\ensuremath{+}0.83}

\newcommand{\budFlopsR}{\ensuremath{+}2.33}

\newcommand{\budKvR}{\ensuremath{+}3.12}

\newcommand{\budParamsMzeroR}{0.00}

\newcommand{\budFlopsMzeroR}{0.00}

\newcommand{\budKvMzeroR}{0.00}

\newcommand{\budParamsDtwoFour}{\ensuremath{+}0.33}

\newcommand{\budFlopsDtwoFour}{\ensuremath{-}0.69}

\newcommand{\budKvDtwoFour}{0.00}

\newcommand{\budParamsDthreeSix}{\ensuremath{+}0.45}

\newcommand{\budFlopsDthreeSix}{\ensuremath{-}2.63}

\newcommand{\budKvDthreeSix}{\ensuremath{+}3.12}

\newcommand{\budParamsMzeroTwoE}{\ensuremath{+}96.72}

\newcommand{\budFlopsMzeroTwoE}{\ensuremath{+}0.19}

\newcommand{\budKvMzeroTwoE}{0.00}

\newcommand{\budParamsMzeroLoop}{0.00}

\newcommand{\budFlopsMzeroLoop}{\ensuremath{+}31.46}

\newcommand{\budKvMzeroLoop}{\ensuremath{+}50.00}

\newcommand{\budParamsSmeltVthree}{\ensuremath{+}0.11}

\newcommand{\budFlopsSmeltVthree}{\ensuremath{+}2.55}

\newcommand{\budKvSmeltVthree}{\ensuremath{+}3.12}

\newcommand{\budParamsSmeltVfour}{\ensuremath{+}0.01}

\newcommand{\budFlopsSmeltVfour}{\ensuremath{+}2.66}

\newcommand{\budKvSmeltVfour}{\ensuremath{+}3.12}

\newcommand{\hgapMthreeMzeroAdamWHzZeroPSixTwoFiveAbs}{3.4}

\newcommand{\hstepExpertsAdamWHzZeroPSixTwoFive}{\ensuremath{-}0.5}
\newcommand{\hstepExpertsAdamWHzZeroPSixTwoFiveAbs}{0.5}
\newcommand{\hstepRouterAdamWHzZeroPSixTwoFive}{\ensuremath{+}4.5}
\newcommand{\hstepRouterAdamWHzZeroPSixTwoFiveAbs}{4.5}

\newcommand{\hstepAttnAdamWHzZeroPSixTwoFiveAbs}{3.6}

\newcommand{\hstepAllShareAdamWHzZeroPSixTwoFiveAbs}{7.5}

\newcommand{\hstepExpertsAdamWHzOnePTwoFiveAbs}{0.4}

\newcommand{\hstepRouterAdamWHzOnePTwoFiveAbs}{0.5}

\newcommand{\hstepAttnAdamWHzOnePTwoFiveAbs}{0.8}

\newcommand{\hstepAllShareAdamWHzOnePTwoFiveAbs}{0.7}

\newcommand{\hstepExpertsAdamWHzTwoPFiveAbs}{2.0}

\newcommand{\hstepRouterAdamWHzTwoPFiveAbs}{2.5}

\newcommand{\hstepAttnAdamWHzTwoPFiveAbs}{0.2}

\newcommand{\hstepAllShareAdamWHzTwoPFiveAbs}{0.3}

\newcommand{\hTestMzeroAdamWHzOneZero}{0.9583}

\newcommand{\hTestMthreeAdamWHzOneZero}{0.9491}

\newcommand{\hgapMthreeMzeroAdamWHzOneZero}{\ensuremath{-}9.2}

\newcommand{\hTestMzeroAdamWHzTwoZero}{0.9263}

\newcommand{\hTestMthreeAdamWHzTwoZero}{0.9181}

\newcommand{\hgapMthreeMzeroAdamWHzTwoZero}{\ensuremath{-}8.2}

\newcommand{\hgapMthreeMzeroMuonHzZeroPSixTwoFiveAbs}{5.7}

\newcommand{\hstepExpertsMuonHzZeroPSixTwoFiveAbs}{0.0}

\newcommand{\hstepRouterMuonHzZeroPSixTwoFiveAbs}{0.6}

\newcommand{\hstepAttnMuonHzZeroPSixTwoFiveAbs}{1.4}

\newcommand{\hstepAllShareMuonHzZeroPSixTwoFiveAbs}{2.0}

\newcommand{\hstepShapeMuonHzZeroPSixTwoFiveAbs}{3.7}

\newcommand{\hstepExpertsMuonHzOnePTwoFiveAbs}{0.3}

\newcommand{\hstepRouterMuonHzOnePTwoFiveAbs}{0.3}

\newcommand{\hstepAttnMuonHzOnePTwoFiveAbs}{1.2}

\newcommand{\hstepAllShareMuonHzOnePTwoFiveAbs}{0.6}

\newcommand{\hstepExpertsMuonHzTwoPFiveAbs}{0.3}

\newcommand{\hstepRouterMuonHzTwoPFiveAbs}{1.0}

\newcommand{\hstepAttnMuonHzTwoPFiveAbs}{0.9}

\newcommand{\hstepAllShareMuonHzTwoPFiveAbs}{0.3}

\newcommand{\hTestMzeroMuonHzOneZero}{0.9403}

\newcommand{\hTestMthreeMuonHzOneZero}{0.9357}

\newcommand{\hgapMthreeMzeroMuonHzOneZero}{\ensuremath{-}4.6}

\newcommand{\hTestMzeroMuonHzTwoZero}{0.9135}

\newcommand{\hTestMthreeMuonHzTwoZero}{0.9085}

\newcommand{\hgapMthreeMzeroMuonHzTwoZero}{\ensuremath{-}5.0}
\newcommand{\hgapMthreeMzeroMuonHzTwoZeroAbs}{5.0}

\newcommand{\hleadMzeroHzOneZero}{\ensuremath{-}17.9}

\newcommand{\hleadMzeroHzTwoZero}{\ensuremath{-}12.8}
\newcommand{\hleadMzeroHzTwoZeroAbs}{12.8}

\newcommand{\lrRuleCoef}{0.771}

\newcommand{\szVtxSzZeroPFiveMuonDnThree}{2.23e\ensuremath{-}3}

\newcommand{\selLrSzZeroPFiveAdamW}{1.7e\ensuremath{-}3}
\newcommand{\selLrSzZeroPFiveMuon}{1.7e\ensuremath{-}3}

\newcommand{\lamMfourAdamWStFourSevenSeven}{6.99}

\newcommand{\lamMsevenMuonStFourSevenSeven}{7.17}

\newcommand{\lamMzeroAdamWStFourSevenSeven}{9.36}

\newcommand{\lamMthreeAdamWStFourSevenSeven}{7.26}

\newcommand{\lamMzeroMuonStFourSevenSeven}{7.61}

\newcommand{\lamMfiveMuonStFourSevenSeven}{6.03}

\newcommand{\lamMinAdamWStFourSevenSeven}{6.99}
\newcommand{\lamMaxAdamWStFourSevenSeven}{9.36}
\newcommand{\lamMinAdamWStFourSevenSixNine}{15.80}
\newcommand{\lamMaxAdamWStFourSevenSixNine}{28.12}

\newcommand{\lamMinAllMuon}{5.52}
\newcommand{\lamMaxAllMuon}{9.29}

\newcommand{\cosExpertsMthreeAdamWStFourSevenSeven}{0.633}

\newcommand{\cosExpertsMthreeAdamWStFourSevenSixNine}{0.317}

\newcommand{\cosExpertsMthreeMuonStFourSevenSeven}{0.495}

\newcommand{\cosExpertsMthreeMuonStFourSevenSixNine}{0.143}

\newcommand{\nSnrAbove}{22}

\newcommand{\snrRatioMedian}{0.993}

\newcommand{\cosExpertsMinMseven}{0.023}
\newcommand{\cosExpertsMaxMseven}{0.062}

\newcommand{\roFirstMinusNativeMfiveAdamW}{188.0}

\newcommand{\roFirstMinusNativeMfiveMuon}{274.3}

\newcommand{\roFirstMinusNativeMthreeAdamW}{98.2}

\newcommand{\roFirstMinusNativeMthreeMuon}{120.9}

\newcommand{\roControlMinusMixtureMin}{0.35}
\newcommand{\roControlMinusMixtureMax}{0.71}
\newcommand{\roTopShareMin}{64}
\newcommand{\roTopShareMax}{95}
\newcommand{\roURatioMin}{1.76}
\newcommand{\roURatioMax}{2.61}
\newcommand{\roWeightTwoMin}{0.89}
\newcommand{\roWeightTwoMax}{0.91}

\newcommand{\budParamsMseven}{\ensuremath{+}0.88}
\newcommand{\budFlopsMseven}{\ensuremath{+}2.44}
\newcommand{\budKvMseven}{\ensuremath{+}3.12}
\newcommand{\budParamsMfive}{\ensuremath{+}0.83}
\newcommand{\budFlopsMfive}{\ensuremath{+}2.44}
\newcommand{\budKvMfive}{\ensuremath{+}3.12}

\newcommand{\budFlopsMzeroLoopInt}{\ensuremath{+}31}
\newcommand{\budKvMzeroLoopInt}{\ensuremath{+}50}
\newcommand{\budParamsMzeroTwoEInt}{\ensuremath{+}97}

\newcommand{\roFirstMinusNativeMinOneDec}{98.2}
\newcommand{\roFirstMinusNativeMaxOneDec}{274.3}

\newcommand{\cfgClipNorm}{1.0}

\newcommand{\testMzeroAdamWMuonDiff}{25.2}

\newcommand{\hTestUMuonHzOneZero}{0.9353}

\newcommand{\hTestUMuonHzTwoZero}{0.9083}

\newcommand{\hgapMthreeUMuonHzOneZero}{\ensuremath{+}0.4}

\newcommand{\hgapUMzeroMuonHzOneZero}{\ensuremath{-}5.1}

\newcommand{\hgapMthreeUMuonHzTwoZero}{\ensuremath{+}0.2}

\newcommand{\hgapUMzeroMuonHzTwoZero}{\ensuremath{-}5.2}

\newcommand{\depthDeltaMthreeMuonVisOneAbs}{121.2}

\newcommand{\depthDeltaMthreeMuonVisThreeAbs}{21.4}

\newcommand{\depthDeltaMthreeMuonVisFourAbs}{63.3}

\newcommand{\depthDeltaMthreeMuonVisSixAbs}{150.0}

\newcommand{\depthDeltaMthreeMuonVisEightAbs}{222.1}

\newcommand{\depthDeltaSmeltVthreeMuonVisOneAbs}{285.4}

\newcommand{\depthDeltaSmeltVthreeMuonVisTwoAbs}{46.7}

\newcommand{\depthDeltaSmeltVthreeMuonVisFourAbs}{12.8}

\newcommand{\depthDeltaSmeltVthreeMuonVisSixAbs}{82.8}

\newcommand{\depthDeltaSmeltVthreeMuonVisEightAbs}{167.0}

\newcommand{\depthDeltaSmeltVfourMuonVisOneAbs}{491.5}

\newcommand{\depthDeltaSmeltVfourMuonVisTwoAbs}{141.3}

\newcommand{\depthDeltaSmeltVfourMuonVisThreeAbs}{30.5}

\newcommand{\depthDeltaSmeltVfourMuonVisSixAbs}{38.3}

\newcommand{\depthDeltaSmeltVfourMuonVisEightAbs}{117.0}

\newcommand{\depthDeltaHuginnMuonVisOneAbs}{51.5}

\newcommand{\depthDeltaHuginnMuonVisTwoAbs}{13.8}

\newcommand{\depthDeltaHuginnMuonVisThreeAbs}{3.1}

\newcommand{\depthDeltaHuginnMuonVisSixAbs}{1.2}

\newcommand{\depthDeltaHuginnMuonVisEightAbs}{1.2}

\newcommand{\depthDeltaOuroMuonVisOneAbs}{171.0}

\newcommand{\depthDeltaOuroMuonVisTwoAbs}{5.1}

\newcommand{\depthDeltaOuroMuonVisFourAbs}{5.5}

\newcommand{\depthDeltaOuroMuonVisSixAbs}{39.7}

\newcommand{\depthDeltaOuroMuonVisEightAbs}{91.8}

\newcommand{\depthHuginnBestMinusMthreeAbs}{13.1}

\newcommand{\gapHuginnMzeroAdamWUpdatedAbs}{49.6}

\newcommand{\testMzeroDeepMuon}{0.9639}
\newcommand{\hTestUAdamWHzOneZero}{0.9514}
\newcommand{\hTestMfourAdamWHzOneZero}{0.9490}
\newcommand{\hTestMfourMuonHzOneZero}{0.9361}
\newcommand{\hTestUAdamWHzTwoZero}{0.9201}
\newcommand{\hTestMfourAdamWHzTwoZero}{0.9171}
\newcommand{\hTestMfourMuonHzTwoZero}{0.9090}
\newcommand{\testMzeroAdamWLarge}{0.9143}
\newcommand{\testMzeroMuonLarge}{0.9003}
\newcommand{\testUAdamWLarge}{0.9083}
\newcommand{\testUMuonLarge}{0.8966}
\newcommand{\testMfourAdamWLarge}{0.9067}
\newcommand{\testMfourMuonLarge}{0.8963}
\newcommand{\testMthreeAdamWLarge}{0.9070}
\newcommand{\testMthreeMuonLarge}{0.8950}

\newcommand{\hgapUMzeroAdamWHzOneZero}{\ensuremath{-}6.9}

\newcommand{\hgapMfourUAdamWHzOneZero}{\ensuremath{-}2.4}

\newcommand{\hgapMthreeMfourAdamWHzOneZero}{\ensuremath{+}0.1}

\newcommand{\hgapMfourUMuonHzOneZero}{\ensuremath{+}0.8}

\newcommand{\hgapMthreeMfourMuonHzOneZero}{\ensuremath{-}0.4}

\newcommand{\hgapUMzeroAdamWHzTwoZero}{\ensuremath{-}6.2}

\newcommand{\hgapMfourUAdamWHzTwoZero}{\ensuremath{-}3.0}

\newcommand{\hgapMthreeMfourAdamWHzTwoZero}{\ensuremath{+}1.0}

\newcommand{\hgapMfourUMuonHzTwoZero}{\ensuremath{+}0.6}

\newcommand{\hgapMthreeMfourMuonHzTwoZero}{\ensuremath{-}0.4}

\newcommand{\gapUMzeroAdamWLarge}{\ensuremath{-}6.0}

\newcommand{\gapMfourUAdamWLarge}{\ensuremath{-}1.6}

\newcommand{\gapMthreeMfourAdamWLarge}{\ensuremath{+}0.3}

\newcommand{\gapMthreeMzeroAdamWLarge}{\ensuremath{-}7.4}

\newcommand{\gapUMzeroMuonLarge}{\ensuremath{-}3.8}

\newcommand{\gapMfourUMuonLarge}{\ensuremath{-}0.2}

\newcommand{\gapMthreeMfourMuonLarge}{\ensuremath{-}1.3}
\newcommand{\gapMthreeMfourMuonLargeAbs}{1.3}

\newcommand{\gapMthreeMzeroMuonLarge}{\ensuremath{-}5.3}

\newcommand{\gapMthreeUMuonLarge}{\ensuremath{-}1.6}

\newcommand{\shareUofMthreeAdamW}{76.9}

\newcommand{\shareMfourofMthreeMuonLarge}{75.3}

\newcommand{\leadMzeroLarge}{\ensuremath{-}14.0}

\newcommand{\ratioLeadMzeroGainAdamW}{2.8}

\newcommand{\xgapUMuonMzeroAdamW}{\ensuremath{-}29.4}

\newcommand{\ratioLeadMzeroGainAdamWHzOneZero}{2.0}

\newcommand{\xgapUMuonMzeroAdamWHzOneZero}{\ensuremath{-}23.0}

\newcommand{\ratioLeadMzeroGainAdamWHzTwoZero}{1.6}

\newcommand{\xgapUMuonMzeroAdamWHzTwoZero}{\ensuremath{-}18.0}

\newcommand{\ratioLeadMzeroGainAdamWLarge}{1.9}

\newcommand{\xgapUMuonMzeroAdamWLarge}{\ensuremath{-}17.8}

\newcommand{\gapMfourRMuonAbsTwoDec}{0.03}

\newcommand{\shrinkMfourR}{\ensuremath{+}3.0}

\newcommand{\shrinkMfourRAbsTwoDec}{3.01}

\newcommand{\gapMsevenMfourAdamWTwoDec}{\ensuremath{+}3.55}
\newcommand{\gapMsevenMfourAdamWAbsTwoDec}{3.55}
\newcommand{\gapMfiveMsevenAdamWTwoDec}{\ensuremath{-}3.87}
\newcommand{\gapMfiveMsevenAdamWAbsTwoDec}{3.87}
\newcommand{\gapMthreeMfiveAdamWTwoDec}{\ensuremath{+}0.21}
\newcommand{\gapMthreeMfiveAdamWAbsTwoDec}{0.21}
\newcommand{\gapMfiveMfourAdamWTwoDec}{\ensuremath{-}0.32}

\newcommand{\gapMfiveUAdamWTwoDec}{\ensuremath{-}2.29}

\newcommand{\gapMsevenMfourMuonTwoDec}{\ensuremath{+}0.15}

\newcommand{\gapMfiveMsevenMuonTwoDec}{\ensuremath{-}1.69}
\newcommand{\gapMfiveMsevenMuonAbsTwoDec}{1.69}
\newcommand{\gapMthreeMfiveMuonTwoDec}{\ensuremath{+}1.18}

\newcommand{\gapMfiveMfourMuonTwoDec}{\ensuremath{-}1.55}

\newcommand{\gapMfiveUMuonTwoDec}{\ensuremath{-}0.67}
\newcommand{\gapMfiveUMuonAbsTwoDec}{0.67}

\newcommand{\gapMzeroLRMzeroMuon}{\ensuremath{+}1.0}

\newcommand{\gapMzeroIMzeroMuon}{\ensuremath{-}0.9}

\newcommand{\gridDeltaMzeroAdamWTwoFourEmThree}{\ensuremath{+}8.2}

\newcommand{\gridDeltaMzeroAdamWThreeFourEmThree}{\ensuremath{+}135.9}

\newcommand{\gridDeltaMzeroMuonSixEmFour}{\ensuremath{+}9.3}

\newcommand{\gridDeltaMzeroMuonOneTwoEmThree}{\ensuremath{+}1.1}

\newcommand{\lrHalfGlobalMthreeAdamWAbs}{11.3}

\newcommand{\gapMzeroDeepMzeroMuonAbs}{10.4}

\newcommand{\shareMzeroDeepofMzeroLoopMuon}{85.2}

\newcommand{\rangeShapeMinAbs}{3.8}

\newcommand{\rangeShapeMaxAbs}{6.9}

\newcommand{\rangeMthreeUMuonMaxAbs}{1.6}

\newcommand{\budParamsMzeroDeep}{\ensuremath{+}0.13}
\newcommand{\budFlopsMzeroDeep}{\ensuremath{+}31.36}

\newcommand{\budKvMzeroDeep}{\ensuremath{+}50.00}

\newcommand{\budParamsMzeroDeepVsLoop}{\ensuremath{+}0.13}
\newcommand{\budFlopsMzeroDeepVsLoop}{\ensuremath{-}0.07}

\newcommand{\budParamsULarge}{\ensuremath{+}0.40}

\newcommand{\budParamsMfourLarge}{\ensuremath{+}2.52}

\newcommand{\budParamsMthreeLarge}{\ensuremath{+}0.93}
\newcommand{\budFlopsMfourLarge}{\ensuremath{+}2.55}
\newcommand{\budFlopsULarge}{\ensuremath{+}2.55}
\newcommand{\budFlopsMthreeLarge}{\ensuremath{+}2.67}

\newcommand{\clipStepsMzeroRAdamW}{41}

\newcommand{\clipPctMzeroRAdamWTwoDec}{0.86}
\newcommand{\clipStepsMzeroRMuon}{103}

\newcommand{\clipPctMzeroRMuonTwoDec}{2.16}
\newcommand{\clipStepsRAdamW}{39}

\newcommand{\clipPctRAdamWTwoDec}{0.82}
\newcommand{\clipStepsRMuon}{93}

\newcommand{\clipPctRMuonTwoDec}{1.95}
\newcommand{\clipStepsMfourAdamW}{36}

\newcommand{\clipPctMfourAdamWTwoDec}{0.75}
\newcommand{\clipStepsMfourMuon}{89}

\newcommand{\clipPctMfourMuonTwoDec}{1.87}
\newcommand{\clipStepsMthreeAdamW}{39}

\newcommand{\clipPctMthreeAdamWTwoDec}{0.82}
\newcommand{\clipStepsMthreeMuon}{91}

\newcommand{\clipPctMthreeMuonTwoDec}{1.91}
\newcommand{\clipStepsMzeroLRAdamW}{44}

\newcommand{\clipPctMzeroLRAdamWTwoDec}{0.92}
\newcommand{\clipStepsMzeroLRMuon}{110}

\newcommand{\clipPctMzeroLRMuonTwoDec}{2.31}
\newcommand{\clipStepsMzeroIAdamW}{46}

\newcommand{\clipPctMzeroIAdamWTwoDec}{0.96}
\newcommand{\clipStepsMzeroIMuon}{108}

\newcommand{\clipPctMzeroIMuonTwoDec}{2.26}
\newcommand{\clipPctScaleMinAdamW}{0.75}
\newcommand{\clipPctScaleMaxAdamW}{0.86}

\newcommand{\clipPctScaleMinMuon}{1.87}
\newcommand{\clipPctScaleMaxMuon}{2.16}

\newcommand{\tokPerActiveMain}{24.6}

\newcommand{\tokPerActiveHzTwoZero}{98.3}
\newcommand{\tokPerActiveLarge}{24.1}

\newcommand{\gpuhPctUMzeroAdamW}{\ensuremath{+}13.7}

\newcommand{\gpuhPctUMzeroMuon}{\ensuremath{+}9.0}

\newcommand{\gpuhMzeroDeepMuon}{51.3}

\newcommand{\tokEqUAdamWTwoDec}{1.12}

\newcommand{\tokEqMthreeAdamWTwoDec}{1.16}

\newcommand{\tokEqMfourAdamWTwoDec}{1.16}

\newcommand{\tokEqMzeroLoopAdamWTwoDec}{1.28}

\newcommand{\tokEqMzeroTwoEAdamWTwoDec}{1.27}

\newcommand{\tokGainMzeroVisitFlopsAdamWAbs}{16.3}

\newcommand{\tokGainMzeroVisitFlopsAdamWQuadAbs}{17.4}

\newcommand{\tokEqUMuonTwoDec}{1.09}

\newcommand{\tokEqMthreeMuonTwoDec}{1.08}

\newcommand{\tokEqMfourMuonTwoDec}{1.07}

\newcommand{\tokEqMzeroLoopMuonTwoDec}{1.28}

\newcommand{\tokEqMzeroDeepMuonTwoDec}{1.24}

\newcommand{\tokEqMzeroTwoEMuonTwoDec}{1.22}

\newcommand{\tokGainMzeroVisitFlopsMuonAbs}{13.4}

\newcommand{\tokGainMzeroVisitFlopsMuonQuadAbs}{14.3}

\newcommand{\tokEqMuonSwitchMain}{1.5}
\newcommand{\tokEqMuonSwitchMainTwoDec}{1.53}

\newcommand{\tokEqMuonSwitchHzOneZeroTwoDec}{1.44}

\newcommand{\tokMuonReachesAdamWHzOneZero}{6.9}
\newcommand{\tokEqMuonSwitchHzTwoZero}{1.4}
\newcommand{\tokEqMuonSwitchHzTwoZeroTwoDec}{1.39}

\newcommand{\tokMuonReachesAdamWHzTwoZero}{14.4}

\newcommand{\gpuhSeedMinMthreeAdamW}{40.9}
\newcommand{\gpuhSeedMaxMthreeAdamW}{45.4}

\newcommand{\flopsRatioMzeroLoopTwoDec}{1.31}

\newcommand{\flopsTotalMthreeLarge}{\ensuremath{8.0\times10^{19}}}

\newcommand{\seedSdPooled}{1.03}
\newcommand{\seedSdAdamW}{1.56}
\newcommand{\seedSdMuon}{0.62}
\newcommand{\seedSdGroupMthreeAdamW}{1.20}
\newcommand{\seedTestMthreeAdamWSFortyThree}{0.9925}
\newcommand{\seedTestMthreeAdamWSFortyFour}{0.9904}
\newcommand{\seedTestMeanMthreeAdamW}{0.9912}
\newcommand{\seedSdGroupMfourAdamW}{1.85}
\newcommand{\seedTestMfourAdamWSFortyThree}{0.9923}
\newcommand{\seedTestMfourAdamWSFortyFour}{0.9886}
\newcommand{\seedTestMeanMfourAdamW}{0.9905}
\newcommand{\seedSdGroupMthreeMuon}{0.28}
\newcommand{\seedTestMthreeMuonSFortyThree}{0.9704}
\newcommand{\seedTestMthreeMuonSFortyFour}{0.9710}
\newcommand{\seedTestMeanMthreeMuon}{0.9707}
\newcommand{\seedSdGroupMfourMuon}{0.96}
\newcommand{\seedTestMfourMuonSFortyThree}{0.9701}
\newcommand{\seedTestMfourMuonSFortyFour}{0.9691}
\newcommand{\seedTestMeanMfourMuon}{0.9701}
\newcommand{\seedSdGroupMzeroLoopMuon}{0.62}
\newcommand{\seedTestMzeroLoopMuonSFortyThree}{0.9631}
\newcommand{\seedTestMzeroLoopMuonSFortyFour}{0.9619}
\newcommand{\seedTestMeanMzeroLoopMuon}{0.9624}
\newcommand{\seedSdGroupMzeroDeepMuon}{0.36}
\newcommand{\seedTestMzeroDeepMuonSFortyThree}{0.9633}
\newcommand{\seedTestMzeroDeepMuonSFortyFour}{0.9633}
\newcommand{\seedTestMeanMzeroDeepMuon}{0.9635}
\newcommand{\seedTwoSeOneOne}{2.9}

\newcommand{\seedTwoSeOneOneAdamW}{4.4}
\newcommand{\seedTwoSeOneOneMuon}{1.7}
\newcommand{\seedTwoSeThreeOne}{2.4}

\newcommand{\seedTwoSeThreeOneAdamW}{3.6}
\newcommand{\seedTwoSeThreeOneMuon}{1.4}
\newcommand{\seedTwoSeThreeThree}{1.7}

\newcommand{\seedTwoSeThreeThreeAdamW}{2.5}
\newcommand{\seedTwoSeThreeThreeMuon}{1.0}

\newcommand{\seedSeGapMthreeMfourAdamW}{0.84}

\newcommand{\seedDiffMthreeMfourAdamW}{\ensuremath{+}0.7}
\newcommand{\seedDiffMthreeMfourAdamWAbs}{0.7}

\newcommand{\seedTwoSeMthreeMfourAdamW}{1.7}

\newcommand{\seedVerdictMthreeMfourAdamW}{within noise}

\newcommand{\seedNsMthreeMfourAdamW}{3 vs 3}
\newcommand{\seedPairMthreeMfourAdamWSFortyTwo}{\ensuremath{-}0.1}

\newcommand{\seedPairMthreeMfourAdamWSFortyThree}{\ensuremath{+}0.3}

\newcommand{\seedPairMthreeMfourAdamWSFortyFour}{\ensuremath{+}1.8}

\newcommand{\seedSeGapMthreeMfourMuon}{0.84}

\newcommand{\seedDiffMthreeMfourMuon}{\ensuremath{+}0.6}
\newcommand{\seedDiffMthreeMfourMuonAbs}{0.6}

\newcommand{\seedTwoSeMthreeMfourMuon}{1.7}

\newcommand{\seedVerdictMthreeMfourMuon}{within noise}

\newcommand{\seedNsMthreeMfourMuon}{3 vs 3}
\newcommand{\seedPairMthreeMfourMuonSFortyTwo}{\ensuremath{-}0.4}

\newcommand{\seedPairMthreeMfourMuonSFortyThree}{\ensuremath{+}0.3}

\newcommand{\seedPairMthreeMfourMuonSFortyFour}{\ensuremath{+}1.8}

\newcommand{\seedSeGapMzeroLoopMzeroDeepMuon}{0.84}

\newcommand{\seedDiffMzeroLoopMzeroDeepMuon}{\ensuremath{-}1.1}
\newcommand{\seedDiffMzeroLoopMzeroDeepMuonAbs}{1.1}

\newcommand{\seedTwoSeMzeroLoopMzeroDeepMuon}{1.7}

\newcommand{\seedVerdictMzeroLoopMzeroDeepMuon}{within noise}

\newcommand{\seedNsMzeroLoopMzeroDeepMuon}{3 vs 3}
\newcommand{\seedPairMzeroLoopMzeroDeepMuonSFortyTwo}{\ensuremath{-}1.8}

\newcommand{\seedPairMzeroLoopMzeroDeepMuonSFortyThree}{\ensuremath{-}0.2}

\newcommand{\seedPairMzeroLoopMzeroDeepMuonSFortyFour}{\ensuremath{-}1.4}

\newcommand{\seedSeGapMthreeUAdamW}{1.19}

\newcommand{\seedDiffMthreeUAdamW}{\ensuremath{-}1.5}
\newcommand{\seedDiffMthreeUAdamWAbs}{1.5}

\newcommand{\seedVerdictMthreeUAdamW}{within noise}

\newcommand{\seedNsMthreeUAdamW}{3 vs 1}
\newcommand{\seedPairMthreeUAdamWSFortyTwo}{\ensuremath{-}2.1}

\newcommand{\seedSeGapMthreeUMuon}{1.19}

\newcommand{\seedDiffMthreeUMuon}{\ensuremath{+}0.5}
\newcommand{\seedDiffMthreeUMuonAbs}{0.5}

\newcommand{\seedTwoSeMthreeUMuon}{2.4}

\newcommand{\seedVerdictMthreeUMuon}{within noise}

\newcommand{\seedNsMthreeUMuon}{3 vs 1}
\newcommand{\seedPairMthreeUMuonSFortyTwo}{\ensuremath{+}0.5}

\newcommand{\seedSeGapMfourUAdamW}{1.19}

\newcommand{\seedDiffMfourUAdamW}{\ensuremath{-}2.1}
\newcommand{\seedDiffMfourUAdamWAbs}{2.1}

\newcommand{\seedVerdictMfourUAdamW}{within noise}

\newcommand{\seedNsMfourUAdamW}{3 vs 1}
\newcommand{\seedPairMfourUAdamWSFortyTwo}{\ensuremath{-}2.0}

\newcommand{\seedSeGapMfourUMuon}{1.19}

\newcommand{\seedDiffMfourUMuon}{\ensuremath{-}0.1}

\newcommand{\seedVerdictMfourUMuon}{within noise}

\newcommand{\seedNsMfourUMuon}{3 vs 1}
\newcommand{\seedPairMfourUMuonSFortyTwo}{\ensuremath{+}0.9}

\newcommand{\seedSeGapMzeroLoopMzeroTwoEMuon}{1.19}

\newcommand{\seedDiffMzeroLoopMzeroTwoEMuon}{\ensuremath{-}2.1}

\newcommand{\seedVerdictMzeroLoopMzeroTwoEMuon}{within noise}

\newcommand{\seedNsMzeroLoopMzeroTwoEMuon}{3 vs 1}
\newcommand{\seedPairMzeroLoopMzeroTwoEMuonSFortyTwo}{\ensuremath{-}2.4}

\newcommand{\seedSeGapMfourRAdamW}{1.19}

\newcommand{\seedDiffMfourRAdamW}{\ensuremath{-}3.1}
\newcommand{\seedDiffMfourRAdamWAbs}{3.1}

\newcommand{\seedVerdictMfourRAdamW}{beyond noise}

\newcommand{\seedNsMfourRAdamW}{3 vs 1}
\newcommand{\seedPairMfourRAdamWSFortyTwo}{\ensuremath{-}3.0}

\newcommand{\seedPairMfourRAdamWSFortyThree}{\ensuremath{-}1.4}

\newcommand{\seedPairMfourRAdamWSFortyFour}{\ensuremath{-}5.1}

\newcommand{\seedSeGapMfourRMuon}{1.19}

\newcommand{\seedDiffMfourRMuon}{\ensuremath{-}0.9}

\newcommand{\seedVerdictMfourRMuon}{within noise}

\newcommand{\seedNsMfourRMuon}{3 vs 1}
\newcommand{\seedPairMfourRMuonSFortyTwo}{\ensuremath{+}0.0}

\newcommand{\seedSeGapMzeroDeepMzeroMuon}{1.19}

\newcommand{\seedDiffMzeroDeepMzeroMuon}{\ensuremath{-}10.9}

\newcommand{\seedVerdictMzeroDeepMzeroMuon}{beyond noise}

\newcommand{\seedNsMzeroDeepMzeroMuon}{3 vs 1}
\newcommand{\seedPairMzeroDeepMzeroMuonSFortyTwo}{\ensuremath{-}10.4}

\newcommand{\seedSeGapUMzeroAdamW}{1.46}

\newcommand{\seedDiffUMzeroAdamW}{\ensuremath{-}6.9}

\newcommand{\seedVerdictUMzeroAdamW}{beyond noise}

\newcommand{\seedNsUMzeroAdamW}{1 vs 1}
\newcommand{\seedPairUMzeroAdamWSFortyTwo}{\ensuremath{-}6.9}

\newcommand{\seedSeGapUMzeroMuon}{1.46}

\newcommand{\seedDiffUMzeroMuon}{\ensuremath{-}4.2}

\newcommand{\seedVerdictUMzeroMuon}{beyond noise}

\newcommand{\seedNsUMzeroMuon}{1 vs 1}
\newcommand{\seedPairUMzeroMuonSFortyTwo}{\ensuremath{-}4.2}

\newcommand{\seedSeGapMthreeMzeroAdamW}{1.19}

\newcommand{\seedDiffMthreeMzeroAdamW}{\ensuremath{-}8.4}

\newcommand{\seedVerdictMthreeMzeroAdamW}{beyond noise}

\newcommand{\seedNsMthreeMzeroAdamW}{3 vs 1}
\newcommand{\seedPairMthreeMzeroAdamWSFortyTwo}{\ensuremath{-}9.0}

\newcommand{\seedSeGapMthreeMzeroMuon}{1.19}

\newcommand{\seedDiffMthreeMzeroMuon}{\ensuremath{-}3.7}

\newcommand{\seedVerdictMthreeMzeroMuon}{beyond noise}

\newcommand{\seedNsMthreeMzeroMuon}{3 vs 1}
\newcommand{\seedPairMthreeMzeroMuonSFortyTwo}{\ensuremath{-}3.7}

\newcommand{\seedSeGapMzeroLoopMzeroMuon}{1.19}

\newcommand{\seedDiffMzeroLoopMzeroMuon}{\ensuremath{-}12.0}

\newcommand{\seedVerdictMzeroLoopMzeroMuon}{beyond noise}

\newcommand{\seedNsMzeroLoopMzeroMuon}{3 vs 1}
\newcommand{\seedPairMzeroLoopMzeroMuonSFortyTwo}{\ensuremath{-}12.2}

\newcommand{\seedVerdictOneGMzeroAdamW}{beyond noise}

\newcommand{\seedVerdictOneUGMuon}{beyond noise}

\newcommand{\seedVerdictOneOwnMfourRMuon}{within noise}

\newcommand{\gapLrHalfLocalGlobalAdamWAbs}{8.0}

\newcommand{\seedVerdictOneLrHalfLocalGlobalAdamW}{beyond noise}

\newcommand{\seedGapResidualStepMuonAdamW}{\ensuremath{+}2.2}
\newcommand{\seedGapResidualStepMuonAdamWAbs}{2.2}

\newcommand{\seedTwoSeResidualStepMuonAdamW}{3.4}
\newcommand{\seedTwoSeResidualStepMuonAdamWTwoDec}{3.37}
\newcommand{\seedVerdictResidualStepMuonAdamW}{within noise}

\newcommand{\interimDagger}{}
\newcommand{\interimNote}{}

\newcommand{\gapMfourRAdamWSFortyTwo}{\ensuremath{-}3.0}

\newcommand{\gapMthreeMfourAdamWAbsSFortyTwo}{0.1}

\newcommand{\gapMthreeMfourAdamWTwoDecSFortyTwo}{\ensuremath{-}0.10}

\newcommand{\gapMthreeMfourMuonAbsSFortyTwo}{0.4}

\newcommand{\gapMthreeMfourMuonTwoDecSFortyTwo}{\ensuremath{-}0.37}
\newcommand{\gapMthreeMzeroAdamWSFortyTwo}{\ensuremath{-}9.0}
\newcommand{\gapMthreeMzeroAdamWAbsSFortyTwo}{9.0}

\newcommand{\gapMthreeMzeroMuonSFortyTwo}{\ensuremath{-}3.7}
\newcommand{\gapMthreeMzeroMuonAbsSFortyTwo}{3.7}

\newcommand{\gapMzeroDeepMzeroMuonTwoDecSFortyTwo}{\ensuremath{-}10.43}

\newcommand{\gapMzeroLoopMzeroMuonAbsSFortyTwo}{12.2}

\newcommand{\gapMzeroRMzeroAdamWSFortyTwo}{\ensuremath{-}4.0}

\newcommand{\gapMzeroTwoEMzeroMuonAbsSFortyTwo}{9.9}

\newcommand{\gapUMzeroAdamWAbsSFortyTwo}{6.9}

\newcommand{\gapUMzeroMuonAbsSFortyTwo}{4.2}

\newcommand{\shareMzeroDeepofMzeroLoopMuonSFortyTwo}{85.2}

\renewcommand{\gapMfourRAdamWAbs}{3.6}

\renewcommand{\gapMfourUAdamW}{\ensuremath{-}2.6}
\renewcommand{\gapMfourUAdamWAbs}{2.6}

\renewcommand{\gapMfourUMuon}{\ensuremath{-}0.8}
\renewcommand{\gapMfourUMuonAbs}{0.8}

\renewcommand{\gapMthreeMfourAdamW}{\ensuremath{+}0.7}

\renewcommand{\gapMthreeMfourMuon}{\ensuremath{+}0.6}

\renewcommand{\gapMthreeMzeroAdamW}{\ensuremath{-}6.3}
\renewcommand{\gapMthreeMzeroAdamWAbs}{6.3}

\renewcommand{\gapMthreeMzeroMuon}{\ensuremath{-}3.2}
\renewcommand{\gapMthreeMzeroMuonAbs}{3.2}

\renewcommand{\gapMzeroLoopMzeroMuon}{\ensuremath{-}11.5}
\renewcommand{\gapMzeroLoopMzeroMuonAbs}{11.5}

\renewcommand{\gapMzeroRMzeroAdamW}{\ensuremath{-}1.4}
\renewcommand{\gapMzeroRMzeroAdamWAbs}{1.4}

\renewcommand{\gapMzeroTwoEMzeroMuon}{\ensuremath{-}9.3}
\renewcommand{\gapMzeroTwoEMzeroMuonAbs}{9.3}

\renewcommand{\gapUMzeroAdamW}{\ensuremath{-}4.5}
\renewcommand{\gapUMzeroAdamWAbs}{4.5}

\renewcommand{\gapUMzeroMuon}{\ensuremath{-}3.0}
\renewcommand{\gapUMzeroMuonAbs}{3.0}

\renewcommand{\leadMzero}{\ensuremath{-}23.6}
\renewcommand{\leadMzeroAbs}{23.6}

\renewcommand{\ratioLeadMzeroGainAdamW}{3.7}

\renewcommand{\seedDiffMfourRAdamW}{\ensuremath{-}3.6}
\renewcommand{\seedDiffMfourRAdamWAbs}{3.6}

\renewcommand{\seedDiffMfourUAdamW}{\ensuremath{-}2.6}
\renewcommand{\seedDiffMfourUAdamWAbs}{2.6}

\renewcommand{\seedDiffMfourUMuon}{\ensuremath{-}0.8}

\renewcommand{\seedDiffMthreeMzeroAdamW}{\ensuremath{-}6.3}

\renewcommand{\seedDiffMthreeMzeroMuon}{\ensuremath{-}3.2}

\renewcommand{\seedDiffMthreeUAdamW}{\ensuremath{-}1.9}
\renewcommand{\seedDiffMthreeUAdamWAbs}{1.9}

\renewcommand{\seedDiffMthreeUMuon}{\ensuremath{-}0.2}
\renewcommand{\seedDiffMthreeUMuonAbs}{0.2}

\renewcommand{\seedDiffMzeroDeepMzeroMuon}{\ensuremath{-}10.4}

\renewcommand{\seedDiffMzeroLoopMzeroMuon}{\ensuremath{-}11.5}

\renewcommand{\seedDiffMzeroLoopMzeroTwoEMuon}{\ensuremath{-}2.3}

\renewcommand{\seedDiffUMzeroAdamW}{\ensuremath{-}4.5}

\renewcommand{\seedDiffUMzeroMuon}{\ensuremath{-}3.0}

\renewcommand{\seedGapResidualStepMuonAdamW}{\ensuremath{+}2.7}
\renewcommand{\seedGapResidualStepMuonAdamWAbs}{2.7}

\renewcommand{\seedNsMfourRAdamW}{3 vs 3}
\renewcommand{\seedNsMfourUAdamW}{3 vs 3}
\renewcommand{\seedNsMfourUMuon}{3 vs 3}
\renewcommand{\seedNsMthreeMzeroAdamW}{3 vs 3}
\renewcommand{\seedNsMthreeMzeroMuon}{3 vs 3}
\renewcommand{\seedNsMthreeUAdamW}{3 vs 3}
\renewcommand{\seedNsMthreeUMuon}{3 vs 3}
\renewcommand{\seedNsMzeroDeepMzeroMuon}{3 vs 3}
\renewcommand{\seedNsMzeroLoopMzeroMuon}{3 vs 3}
\renewcommand{\seedNsMzeroLoopMzeroTwoEMuon}{3 vs 3}
\renewcommand{\seedNsUMzeroAdamW}{3 vs 3}
\renewcommand{\seedNsUMzeroMuon}{3 vs 3}
\renewcommand{\seedPairMfourRAdamWSFortyFour}{\ensuremath{-}4.6}

\renewcommand{\seedPairMfourRAdamWSFortyThree}{\ensuremath{-}3.3}

\renewcommand{\seedSdAdamW}{1.31}
\renewcommand{\seedSeGapMfourRAdamW}{1.07}
\renewcommand{\seedSeGapMfourRMuon}{0.71}
\renewcommand{\seedSeGapMfourUAdamW}{1.07}
\renewcommand{\seedSeGapMfourUMuon}{0.50}
\renewcommand{\seedSeGapMthreeMfourAdamW}{1.07}
\renewcommand{\seedSeGapMthreeMfourMuon}{0.50}
\renewcommand{\seedSeGapMthreeMzeroAdamW}{1.07}
\renewcommand{\seedSeGapMthreeMzeroMuon}{0.50}
\renewcommand{\seedSeGapMthreeUAdamW}{1.07}
\renewcommand{\seedSeGapMthreeUMuon}{0.50}
\renewcommand{\seedSeGapMzeroDeepMzeroMuon}{0.50}
\renewcommand{\seedSeGapMzeroLoopMzeroDeepMuon}{0.50}
\renewcommand{\seedSeGapMzeroLoopMzeroMuon}{0.50}
\renewcommand{\seedSeGapMzeroLoopMzeroTwoEMuon}{0.50}
\renewcommand{\seedSeGapUMzeroAdamW}{1.07}
\renewcommand{\seedSeGapUMzeroMuon}{0.50}

\renewcommand{\seedTwoSeMthreeMfourAdamW}{2.1}

\renewcommand{\seedTwoSeMthreeMfourMuon}{1.0}

\renewcommand{\seedTwoSeMthreeUMuon}{1.0}

\renewcommand{\seedTwoSeMzeroLoopMzeroDeepMuon}{1.0}

\renewcommand{\seedTwoSeOneOneAdamW}{3.7}

\renewcommand{\seedTwoSeResidualStepMuonAdamW}{2.6}
\renewcommand{\seedTwoSeResidualStepMuonAdamWTwoDec}{2.57}
\renewcommand{\seedTwoSeThreeOneAdamW}{3.0}
\renewcommand{\seedTwoSeThreeThreeAdamW}{2.1}

\renewcommand{\seedVerdictMfourRAdamW}{beyond noise}
\renewcommand{\seedVerdictMfourRMuon}{within noise}
\renewcommand{\seedVerdictMfourUAdamW}{beyond noise}
\renewcommand{\seedVerdictMfourUMuon}{within noise}
\renewcommand{\seedVerdictMthreeMfourAdamW}{within noise}
\renewcommand{\seedVerdictMthreeMfourMuon}{within noise}
\renewcommand{\seedVerdictMthreeMzeroAdamW}{beyond noise}
\renewcommand{\seedVerdictMthreeMzeroMuon}{beyond noise}
\renewcommand{\seedVerdictMthreeUAdamW}{within noise}
\renewcommand{\seedVerdictMthreeUMuon}{within noise}
\renewcommand{\seedVerdictMzeroDeepMzeroMuon}{beyond noise}
\renewcommand{\seedVerdictMzeroLoopMzeroDeepMuon}{beyond noise}
\renewcommand{\seedVerdictMzeroLoopMzeroMuon}{beyond noise}
\renewcommand{\seedVerdictMzeroLoopMzeroTwoEMuon}{beyond noise}
\renewcommand{\seedVerdictOneGMzeroAdamW}{beyond noise}

\renewcommand{\seedVerdictOneLrHalfLocalGlobalAdamW}{beyond noise}

\renewcommand{\seedVerdictOneOwnMfourRMuon}{within noise}

\renewcommand{\seedVerdictOneUGMuon}{beyond noise}

\renewcommand{\seedVerdictResidualStepMuonAdamW}{beyond noise}
\renewcommand{\seedVerdictUMzeroAdamW}{beyond noise}
\renewcommand{\seedVerdictUMzeroMuon}{beyond noise}
\renewcommand{\shareMzeroDeepofMzeroLoopMuon}{90.2}
\renewcommand{\shareUofMthreeAdamW}{70.1}
\renewcommand{\xgapUMuonMzeroAdamW}{\ensuremath{-}26.6}

\newcommand{\seedDfAdamW}{12}
\newcommand{\seedDfMuon}{14}
\newcommand{\seedDiffMfourMzeroAdamW}{\ensuremath{-}7.0}
\newcommand{\seedDiffMfourMzeroAdamWAbs}{7.0}

\newcommand{\seedDiffMfourMzeroMuon}{\ensuremath{-}3.8}

\newcommand{\seedDiffMzeroHalfLrMzeroAdamWAbs}{7.2}

\newcommand{\seedDiffMzeroIMzeroAdamW}{\ensuremath{-}0.3}

\newcommand{\seedDiffMzeroIMzeroMuon}{\ensuremath{-}0.4}

\newcommand{\seedDiffMzeroIMzeroMuonTwoDec}{\ensuremath{-}0.45}
\newcommand{\seedDiffMzeroLRMzeroAdamW}{\ensuremath{-}0.8}

\newcommand{\seedDiffMzeroLRMzeroMuon}{\ensuremath{+}1.4}

\newcommand{\seedDiffMzeroLRMzeroMuonTwoDec}{\ensuremath{+}1.42}
\newcommand{\seedDiffMzeroRMzeroMuon}{\ensuremath{+}0.6}

\newcommand{\seedSdGroupMzeroAdamW}{1.86}
\newcommand{\seedSdGroupMzeroMuon}{0.49}
\newcommand{\seedSdGroupMzeroRAdamW}{0.52}
\newcommand{\seedSdGroupMzeroTwoEMuon}{0.69}
\newcommand{\seedSdGroupRAdamW}{1.25}
\newcommand{\seedSdGroupUAdamW}{0.46}
\newcommand{\seedSdGroupUMuon}{0.65}

\newcommand{\seedTestMeanMzeroAdamW}{0.9975}
\newcommand{\seedTestMeanMzeroMuon}{0.9739}
\newcommand{\seedTestMeanMzeroTwoEMuon}{0.9647}
\newcommand{\seedTestMeanUAdamW}{0.9930}
\newcommand{\seedTestMeanUMuon}{0.9709}
\newcommand{\seedTestMzeroAdamWSFortyFour}{0.9959}
\newcommand{\seedTestMzeroAdamWSFortyThree}{0.9971}
\newcommand{\seedTestMzeroMuonSFortyFour}{0.9734}
\newcommand{\seedTestMzeroMuonSFortyThree}{0.9740}
\newcommand{\seedTestMzeroTwoEMuonSFortyFour}{0.9654}
\newcommand{\seedTestMzeroTwoEMuonSFortyThree}{0.9640}
\newcommand{\seedTestUAdamWSFortyFour}{0.9930}
\newcommand{\seedTestUAdamWSFortyThree}{0.9935}
\newcommand{\seedTestUMuonSFortyFour}{0.9712}
\newcommand{\seedTestUMuonSFortyThree}{0.9713}

\newcommand{\seedTwoSeMzeroIMzeroAdamW}{3.0}

\newcommand{\seedTwoSeMzeroLRMzeroAdamW}{3.0}

\newcommand{\seedTwoSeMzeroLRMzeroMuon}{1.4}

\newcommand{\seedTwoSeOneOneAdamWTwoDec}{3.71}
\newcommand{\seedTwoSeOneOneMuonTwoDec}{1.74}

\newcommand{\seedTwoSeThreeOneMuonTwoDec}{1.42}

\newcommand{\seedVerdictMzeroHalfLrMzeroAdamW}{beyond noise}
\newcommand{\seedVerdictMzeroIMzeroAdamW}{within noise}

\newcommand{\seedVerdictMzeroLRMzeroAdamW}{within noise}
\newcommand{\seedVerdictMzeroLRMzeroMuon}{within noise}
\newcommand{\seedVerdictMzeroRMzeroMuon}{within noise}

\newcommand{\seGapMfourUAdamW}{1.07}

\newcommand{\verdictGapMfourUAdamW}{beyond noise}
\newcommand{\seGapMfourUMuon}{0.50}

\newcommand{\verdictGapMfourUMuon}{within noise}
\newcommand{\seGapMthreeMfourAdamW}{1.07}

\newcommand{\seGapMthreeMfourMuon}{0.50}

\newcommand{\seGapMthreeMzeroAdamW}{1.07}
\newcommand{\twoSeGapMthreeMzeroAdamW}{2.1}

\newcommand{\seGapMthreeMzeroMuon}{0.50}

\newcommand{\seGapMzeroRMzeroAdamW}{1.07}
\newcommand{\twoSeGapMzeroRMzeroAdamW}{2.1}
\newcommand{\nSeedsGapMzeroRMzeroAdamW}{3 vs 3}
\newcommand{\verdictGapMzeroRMzeroAdamW}{within noise}

\newcommand{\seGapUMzeroAdamW}{1.07}

\newcommand{\seGapUMzeroMuon}{0.50}

\newcommand{\seedVerdictPooledMfourRAdamW}{beyond noise}

\newcommand{\seedVerdictPooledMfourRMuon}{within noise}

\newcommand{\seedVerdictPooledMfourUAdamW}{beyond noise}

\newcommand{\seedVerdictPooledMfourUMuon}{within noise}

\newcommand{\seedVerdictPooledMthreeMfourAdamW}{within noise}

\newcommand{\seedVerdictPooledMthreeMfourMuon}{within noise}

\newcommand{\seedVerdictPooledMthreeMzeroAdamW}{beyond noise}

\newcommand{\seedVerdictPooledMthreeMzeroMuon}{beyond noise}

\newcommand{\seedVerdictPooledMthreeUAdamW}{beyond noise}

\newcommand{\seedVerdictPooledMthreeUMuon}{within noise}

\newcommand{\seedVerdictPooledMzeroDeepMzeroMuon}{beyond noise}

\newcommand{\seedVerdictPooledMzeroLoopMzeroDeepMuon}{within noise}

\newcommand{\seedVerdictPooledMzeroLoopMzeroMuon}{beyond noise}

\newcommand{\seedVerdictPooledMzeroLoopMzeroTwoEMuon}{beyond noise}

\newcommand{\seedVerdictPooledResidualStepMuonAdamW}{within noise}
\newcommand{\seedTwoSePooledResidualStepMuonAdamW}{2.9}
\newcommand{\seedVerdictPooledUMzeroAdamW}{beyond noise}

\newcommand{\seedVerdictPooledUMzeroMuon}{beyond noise}

\newcommand{\seedTestMeanMzeroRAdamW}{0.9961}
\newcommand{\seedTestMeanRAdamW}{0.9941}
\newcommand{\seedTestMzeroRAdamWSFortyFour}{0.9962}
\newcommand{\seedTestMzeroRAdamWSFortyThree}{0.9966}
\newcommand{\seedTestRAdamWSFortyFour}{0.9931}
\newcommand{\seedTestRAdamWSFortyThree}{0.9955}
\newcommand{\testMzeroAdamWLargeLong}{0.8839}
\newcommand{\testMthreeAdamWLargeLong}{0.8771}
\newcommand{\testMfourAdamWLargeLong}{0.8763}
\newcommand{\gapMthreeMfourAdamWLargeLong}{\ensuremath{+}0.8}  %

\newcommand{\gapMfourMzeroAdamWLargeLong}{\ensuremath{-}7.6}  %

\newcommand{\gapMthreeMzeroAdamWLargeLong}{\ensuremath{-}6.8}  %

\newcommand{\testMzeroMuonLargeLong}{0.8734}
\newcommand{\testMthreeMuonLargeLong}{0.8678}
\newcommand{\testMfourMuonLargeLong}{0.8689}
\newcommand{\gapMthreeMfourMuonLargeLong}{\ensuremath{-}1.1}  %

\newcommand{\gapMfourMzeroMuonLargeLong}{\ensuremath{-}4.5}  %

\newcommand{\gapMthreeMzeroMuonLargeLong}{\ensuremath{-}5.7}  %

\newcommand{\shareMfourofMthreeMuonLargeLong}{80}  %
\makeatletter
\ifLuaTeX
\newfontfamily\loopxwordfont{\hl@sans}[Extension=.otf, UprightFont=*-Semibold, ItalicFont=*-SemiboldIt,
  BoldFont=*-Semibold, LetterSpace=-0.5, Scale=MatchLowercase]
\else
\DeclareFontFamily{T1}{hlWordmark}{}
\DeclareFontShape{T1}{hlWordmark}{m}{n}{<-> s*[0.9674107] SourceSans3-Semibold-lf-t1}{}
\DeclareFontShape{T1}{hlWordmark}{m}{it}{<-> s*[0.9674107] SourceSans3-SemiboldIt-lf-t1}{}
\DeclareFontShape{T1}{hlWordmark}{m}{sl}{<-> ssub * hlWordmark/m/it}{}
\SetTracking[context=hlword]{encoding=*}{-5}
\newcommand{\loopxwordfont}{\fontfamily{hlWordmark}\fontseries{m}\selectfont\microtypecontext{tracking=hlword}\lsstyle}
\fi
\makeatother
\newcommand{\ltmark}{\raisebox{-0.17em}{\includegraphics[height=0.92em]{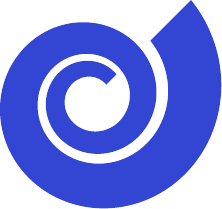}}\kern0.14em}
\newcommand{\moeloop}{\texorpdfstring{\mbox{\ltmark{\loopxwordfont LoopTitan}}}{LoopTitan}}
\newcommand{\MoEtw}{MoE-12}                %
\newcommand{\MoEei}{MoE-18}                %
\newcommand{\MoEloop}{\texorpdfstring{MoE-12\,+\,1 visit}{MoE-12 + 1 visit}}   %
\newcommand{\MoEtwoE}{\texorpdfstring{MoE-12\,+\,2\texttimes{} experts}{MoE-12 + 2x experts}}   %
\newcommand{\Base}{\texorpdfstring{{\loopxwordfont LoopTitan-Base}}{LoopTitan-Base}}         %
\newcommand{\nats}{\,nats}

\newsavebox{\paperfigbox}
\newcommand{\fig}[2]{%
  \IfFileExists{fig/#1.pdf}{%
    \sbox{\paperfigbox}{\includegraphics{fig/#1.pdf}}%
    \ifdim\wd\paperfigbox>\dimexpr\textwidth+1pt\relax
      \begin{wide}\centering\usebox{\paperfigbox}\end{wide}%
    \else
      \centering\usebox{\paperfigbox}\par
    \fi}{%
    \begin{wide}\centering\fcolorbox{hlHair}{white}{\parbox[c][42mm][c]{\dimexpr\linewidth-6mm\relax}{\centering
      \sffamily\small\color{hlSoft}figure in production: fig/\detokenize{#1}.pdf}}\end{wide}}}

\ifcsname proposition\endcsname\else\hltheorems\fi
\providecommand{\normF}[1]{\lVert #1\rVert_{\mathrm{F}}}   %
\providecommand{\half}{\tfrac{1}{2}}                        %

\newcommand{\cvsmURL}{\url{https://github.com/doem97/cvsm}}

\titlespacing*{\paragraph}{0pt}{0.55\baselineskip plus 2pt minus 1pt}{0.4em}

\newcommand{\zc}[1]{\textcolor{red}{[Zichen commented.]}}

\newcommand{\Ctrl}{non-revisit control}             %
\newcommand{\DeepCtrl}{deeper control}  %

\makeatletter
\@ifpackageloaded{tcolorbox}{%
  \newtcolorbox{propbox}{enhanced, unbreakable, frame hidden, interior hidden, coltext=hlInk,
    boxrule=0pt, left=0pt, right=0pt, top=2pt, bottom=6.5pt, boxsep=0pt, before skip=14pt, after skip=14pt,
    borderline north={\hlRuleHair}{0pt}{hlHair}, borderline south={\hlRuleHair}{0pt}{hlHair},
    before upper={\noindent}, after app={\hl@noindentafter\@nobreakfalse}}}{%
  \newenvironment{propbox}{\begin{definitionbox}}{\end{definitionbox}}}
\makeatother
\newcommand{\moeloopname}{\texorpdfstring{{\loopxwordfont LoopTitan}}{LoopTitan}}
\makeatletter
\def\term#1{\par\noindent{\sffamily\bfseries #1.}\enspace}
\makeatother
\titlespacing*{\paragraph}{0pt}{0pt}{0.4em}   %
\newcommand{\recordStudies}{12}
\newcommand{\recordRuns}{482}
\newcommand{\recordDays}{36}
\crefname{figure}{Fig.}{Figs.}\crefname{table}{Tab.}{Tabs.}

\usetikzlibrary{svg.path}
\definecolor{lbGitHub}{HTML}{24292F}
\definecolor{lbHFYellow}{HTML}{FFD21E}
\definecolor{lbHFOrange}{HTML}{FF9D0B}
\definecolor{lbHFDark}{HTML}{3A3B45}
\definecolor{lbHFRed}{HTML}{FF323D}
\colorlet{lbHome}{hlAccent}
\newcommand{\lbUrlHome}{https://datamllab.github.io/LoopTitan/}
\newcommand{\lbUrlHF}{https://huggingface.co/NaturalScale}
\newcommand{\lbUrlGitHub}{https://github.com/datamllab/LoopTitan}
\newcommand{\lbIconGitHub}[1][lbGitHub]{{\tikz[x=0.59pt, y=-0.59pt, sharp corners, baseline={([yshift=1.6pt]current bounding box.south)}]{\fill[#1] (8.000,0.000) .. controls (3.580,0.000) and (0.000,3.580) .. (0.000,8.000) .. controls (0.000,11.540) and (2.290,14.530) .. (5.470,15.590) .. controls (5.870,15.660) and (6.020,15.420) .. (6.020,15.210) .. controls (6.020,15.020) and (6.010,14.390) .. (6.010,13.720) .. controls (4.000,14.090) and (3.480,13.230) .. (3.320,12.780) .. controls (3.230,12.550) and (2.840,11.840) .. (2.500,11.650) .. controls (2.220,11.500) and (1.820,11.130) .. (2.490,11.120) .. controls (3.120,11.110) and (3.570,11.700) .. (3.720,11.940) .. controls (4.440,13.150) and (5.590,12.810) .. (6.050,12.600) .. controls (6.120,12.080) and (6.330,11.730) .. (6.560,11.530) .. controls (4.780,11.330) and (2.920,10.640) .. (2.920,7.580) .. controls (2.920,6.710) and (3.230,5.990) .. (3.740,5.430) .. controls (3.660,5.230) and (3.380,4.410) .. (3.820,3.310) .. controls (3.820,3.310) and (4.490,3.100) .. (6.020,4.130) .. controls (6.660,3.950) and (7.340,3.860) .. (8.020,3.860) .. controls (8.700,3.860) and (9.380,3.950) .. (10.020,4.130) .. controls (11.550,3.090) and (12.220,3.310) .. (12.220,3.310) .. controls (12.660,4.410) and (12.380,5.230) .. (12.300,5.430) .. controls (12.810,5.990) and (13.120,6.700) .. (13.120,7.580) .. controls (13.120,10.650) and (11.250,11.330) .. (9.470,11.530) .. controls (9.760,11.780) and (10.010,12.260) .. (10.010,13.010) .. controls (10.010,14.080) and (10.000,14.940) .. (10.000,15.210) .. controls (10.000,15.420) and (10.150,15.670) .. (10.550,15.590) .. controls (13.798,14.493) and (15.999,11.428) .. (16.000,8.000) .. controls (16.000,3.580) and (12.420,0.000) .. (8.000,0.000) -- cycle;}}}
\newcommand{\lbIconHF}{{\tikz[x=0.108pt, y=-0.108pt, sharp corners, baseline={([yshift=1.7pt]current bounding box.south)}]{%
    \fill[lbHFYellow, even odd rule] (47.210,76.500) .. controls (87.336,76.500) and (87.336,7.000) .. (47.210,7.000) .. controls (7.084,7.000) and (7.084,76.500) .. (47.210,76.500) -- cycle;
    \fill[lbHFOrange, even odd rule] (81.960,41.750) .. controls (81.960,1.624) and (12.460,1.624) .. (12.460,41.750) .. controls (12.460,81.876) and (81.960,81.876) .. (81.960,41.750) -- cycle (8.460,41.750) .. controls (8.460,-2.995) and (85.960,-2.995) .. (85.960,41.750) .. controls (85.960,86.495) and (8.460,86.495) .. (8.460,41.750) -- cycle;
    \fill[lbHFDark, even odd rule] (58.500,32.300) .. controls (59.780,32.740) and (60.280,35.360) .. (61.570,34.680) .. controls (59.148,35.966) and (56.097,35.032) .. (54.810,32.610) .. controls (55.420,33.760) and (57.360,31.890) .. (58.510,32.290) -- (58.500,32.300) -- cycle (34.950,32.300) .. controls (33.670,32.740) and (33.160,35.360) .. (31.880,34.680) .. controls (34.302,35.966) and (37.353,35.032) .. (38.640,32.610) .. controls (38.030,33.760) and (36.080,31.890) .. (34.940,32.290) -- (34.950,32.300) -- cycle;
    \fill[lbHFRed, even odd rule] (46.960,56.290) .. controls (56.790,56.290) and (59.960,47.530) .. (59.960,43.030) .. controls (59.960,40.690) and (58.390,41.430) .. (55.870,42.670) .. controls (53.540,43.820) and (50.410,45.410) .. (46.970,45.410) .. controls (39.780,45.410) and (33.970,38.530) .. (33.970,43.030) .. controls (33.970,47.530) and (37.130,56.290) .. (46.970,56.290) -- (46.960,56.290) -- cycle;
    \fill[lbHFDark, even odd rule] (39.430,54.000) .. controls (40.496,51.833) and (42.418,50.206) .. (44.730,49.510) .. controls (45.130,49.390) and (45.540,50.080) .. (45.970,50.790) .. controls (46.370,51.470) and (46.790,52.160) .. (47.210,52.160) .. controls (47.660,52.160) and (48.110,51.480) .. (48.540,50.810) .. controls (48.990,50.110) and (49.430,49.430) .. (49.860,49.560) .. controls (52.007,50.242) and (53.804,51.740) .. (54.860,53.730) .. controls (58.590,50.790) and (59.960,45.990) .. (59.960,43.030) .. controls (59.960,40.690) and (58.390,41.430) .. (55.870,42.670) -- (55.730,42.740) .. controls (53.420,43.890) and (50.340,45.410) .. (46.960,45.410) .. controls (43.580,45.410) and (40.510,43.890) .. (38.190,42.740) .. controls (35.590,41.450) and (33.960,40.640) .. (33.960,43.030) .. controls (33.960,46.080) and (35.420,51.090) .. (39.430,54.000) -- (39.430,54.000) -- cycle;
    \fill[lbHFOrange, even odd rule] (70.710,37.000) .. controls (74.463,37.000) and (74.463,30.500) .. (70.710,30.500) .. controls (66.957,30.500) and (66.957,37.000) .. (70.710,37.000) -- cycle (24.210,37.000) .. controls (27.963,37.000) and (27.963,30.500) .. (24.210,30.500) .. controls (20.457,30.500) and (20.457,37.000) .. (24.210,37.000) -- cycle (17.520,48.000) .. controls (15.900,48.000) and (14.460,48.660) .. (13.450,49.870) .. controls (12.589,50.934) and (12.119,52.262) .. (12.120,53.630) .. controls (11.490,53.441) and (10.837,53.340) .. (10.180,53.330) .. controls (8.630,53.330) and (7.230,53.920) .. (6.240,54.990) .. controls (4.430,56.870) and (4.101,59.750) .. (5.440,61.990) .. controls (4.555,62.712) and (3.927,63.702) .. (3.650,64.810) .. controls (3.410,65.710) and (3.170,67.610) .. (4.450,69.550) .. controls (3.476,71.048) and (3.336,72.946) .. (4.080,74.570) .. controls (5.100,76.890) and (7.650,78.710) .. (12.600,80.670) .. controls (15.670,81.890) and (18.490,82.670) .. (18.510,82.680) .. controls (22.072,83.667) and (25.744,84.204) .. (29.440,84.280) .. controls (35.300,84.280) and (39.490,82.480) .. (41.900,78.940) .. controls (45.780,73.250) and (45.230,68.040) .. (40.200,63.020) .. controls (37.430,60.240) and (35.580,56.150) .. (35.200,55.250) .. controls (34.420,52.590) and (32.360,49.630) .. (28.950,49.630) .. controls (27.111,49.659) and (25.395,50.577) .. (24.350,52.090) .. controls (23.350,50.830) and (22.370,49.840) .. (21.490,49.270) .. controls (20.316,48.475) and (18.938,48.034) .. (17.520,48.000) -- cycle (17.520,52.000) .. controls (18.030,52.000) and (18.660,52.220) .. (19.340,52.650) .. controls (21.480,54.010) and (25.590,61.080) .. (27.100,63.830) .. controls (27.600,64.750) and (28.470,65.140) .. (29.240,65.140) .. controls (30.790,65.140) and (31.990,63.610) .. (29.390,61.660) .. controls (25.470,58.730) and (26.840,53.940) .. (28.710,53.650) .. controls (28.790,53.630) and (28.880,53.630) .. (28.950,53.630) .. controls (30.650,53.630) and (31.400,56.560) .. (31.400,56.560) .. controls (31.400,56.560) and (33.600,62.080) .. (37.380,65.860) .. controls (41.150,69.630) and (41.350,72.660) .. (38.600,76.690) .. controls (36.720,79.440) and (33.130,80.270) .. (29.440,80.270) .. controls (25.630,80.270) and (21.710,79.370) .. (19.520,78.810) .. controls (19.410,78.780) and (6.070,75.010) .. (7.760,71.810) .. controls (8.040,71.270) and (8.510,71.050) .. (9.100,71.050) .. controls (11.480,71.050) and (15.800,74.590) .. (17.670,74.590) .. controls (18.080,74.590) and (18.370,74.420) .. (18.500,73.990) .. controls (19.290,71.140) and (6.440,69.940) .. (7.520,65.820) .. controls (7.720,65.090) and (8.230,64.800) .. (8.960,64.800) .. controls (12.100,64.800) and (19.160,70.330) .. (20.640,70.330) .. controls (20.750,70.330) and (20.840,70.300) .. (20.880,70.230) .. controls (21.620,69.030) and (21.210,68.190) .. (15.980,65.030) .. controls (10.770,61.870) and (7.100,59.970) .. (9.180,57.700) .. controls (9.420,57.440) and (9.760,57.320) .. (10.180,57.320) .. controls (13.350,57.320) and (20.840,64.140) .. (20.840,64.140) .. controls (20.840,64.140) and (22.860,66.240) .. (24.090,66.240) .. controls (24.370,66.240) and (24.610,66.140) .. (24.770,65.860) .. controls (25.630,64.400) and (16.710,57.640) .. (16.210,54.850) .. controls (15.870,52.950) and (16.450,52.000) .. (17.520,52.000) -- (17.520,52.000) -- cycle;
    \fill[lbHFYellow, even odd rule] (38.600,76.690) .. controls (41.350,72.650) and (41.150,69.620) .. (37.380,65.850) .. controls (33.600,62.080) and (31.400,56.550) .. (31.400,56.550) .. controls (31.400,56.550) and (30.580,53.350) .. (28.710,53.650) .. controls (26.840,53.950) and (25.470,58.730) .. (29.390,61.660) .. controls (33.300,64.590) and (28.610,66.580) .. (27.100,63.830) .. controls (25.600,61.080) and (21.480,54.010) .. (19.340,52.650) .. controls (17.210,51.300) and (15.710,52.050) .. (16.210,54.850) .. controls (16.710,57.640) and (25.640,64.400) .. (24.770,65.850) .. controls (23.900,67.320) and (20.840,64.140) .. (20.840,64.140) .. controls (20.840,64.140) and (11.270,55.430) .. (9.180,57.700) .. controls (7.100,59.970) and (10.770,61.870) .. (15.980,65.030) .. controls (21.210,68.190) and (21.620,69.030) .. (20.880,70.230) .. controls (20.130,71.430) and (8.600,61.700) .. (7.520,65.830) .. controls (6.440,69.940) and (19.290,71.130) .. (18.500,73.980) .. controls (17.700,76.830) and (9.440,68.600) .. (7.760,71.800) .. controls (6.060,75.010) and (19.410,78.780) .. (19.520,78.810) .. controls (23.820,79.930) and (34.770,82.300) .. (38.600,76.690) -- cycle;
    \fill[lbHFOrange, even odd rule] (77.400,48.000) .. controls (79.020,48.000) and (80.470,48.660) .. (81.470,49.870) .. controls (82.331,50.934) and (82.801,52.262) .. (82.800,53.630) .. controls (83.433,53.440) and (84.089,53.339) .. (84.750,53.330) .. controls (86.300,53.330) and (87.700,53.920) .. (88.690,54.990) .. controls (90.500,56.870) and (90.829,59.750) .. (89.490,61.990) .. controls (90.371,62.714) and (90.996,63.703) .. (91.270,64.810) .. controls (91.510,65.710) and (91.750,67.610) .. (90.470,69.550) .. controls (91.444,71.048) and (91.584,72.946) .. (90.840,74.570) .. controls (89.820,76.890) and (87.270,78.710) .. (82.330,80.670) .. controls (79.250,81.890) and (76.430,82.670) .. (76.410,82.680) .. controls (72.848,83.667) and (69.176,84.204) .. (65.480,84.280) .. controls (59.620,84.280) and (55.430,82.480) .. (53.020,78.940) .. controls (49.140,73.250) and (49.690,68.040) .. (54.720,63.020) .. controls (57.500,60.240) and (59.350,56.150) .. (59.730,55.250) .. controls (60.510,52.590) and (62.560,49.630) .. (65.970,49.630) .. controls (67.809,49.659) and (69.525,50.577) .. (70.570,52.090) .. controls (71.570,50.830) and (72.550,49.840) .. (73.440,49.270) .. controls (74.611,48.477) and (75.986,48.036) .. (77.400,48.000) -- cycle (77.400,52.000) .. controls (76.890,52.000) and (76.270,52.220) .. (75.580,52.650) .. controls (73.450,54.010) and (69.330,61.080) .. (67.820,63.830) .. controls (67.405,64.629) and (66.580,65.134) .. (65.680,65.140) .. controls (64.140,65.140) and (62.930,63.610) .. (65.540,61.660) .. controls (69.450,58.730) and (68.080,53.940) .. (66.210,53.650) .. controls (66.131,53.637) and (66.050,53.630) .. (65.970,53.630) .. controls (64.270,53.630) and (63.520,56.560) .. (63.520,56.560) .. controls (63.520,56.560) and (61.320,62.080) .. (57.550,65.860) .. controls (53.770,69.630) and (53.570,72.660) .. (56.330,76.690) .. controls (58.200,79.440) and (61.800,80.270) .. (65.480,80.270) .. controls (69.300,80.270) and (73.210,79.370) .. (75.410,78.810) .. controls (75.510,78.780) and (88.860,75.010) .. (87.170,71.810) .. controls (86.880,71.270) and (86.420,71.050) .. (85.830,71.050) .. controls (83.450,71.050) and (79.120,74.590) .. (77.260,74.590) .. controls (76.840,74.590) and (76.550,74.420) .. (76.430,73.990) .. controls (75.630,71.140) and (88.480,69.940) .. (87.400,65.820) .. controls (87.210,65.090) and (86.700,64.800) .. (85.960,64.800) .. controls (82.820,64.800) and (75.760,70.330) .. (74.280,70.330) .. controls (74.180,70.330) and (74.090,70.300) .. (74.050,70.230) .. controls (73.310,69.030) and (73.710,68.190) .. (78.930,65.030) .. controls (84.160,61.870) and (87.830,59.970) .. (85.730,57.700) .. controls (85.500,57.440) and (85.160,57.320) .. (84.750,57.320) .. controls (81.570,57.320) and (74.080,64.140) .. (74.080,64.140) .. controls (74.080,64.140) and (72.060,66.240) .. (70.840,66.240) .. controls (70.560,66.253) and (70.296,66.105) .. (70.160,65.860) .. controls (69.290,64.400) and (78.210,57.640) .. (78.710,54.850) .. controls (79.050,52.950) and (78.470,52.000) .. (77.400,52.000) -- (77.400,52.000) -- cycle;
    \fill[lbHFYellow, even odd rule] (56.330,76.690) .. controls (53.580,72.650) and (53.770,69.620) .. (57.550,65.850) .. controls (61.320,62.080) and (63.520,56.550) .. (63.520,56.550) .. controls (63.520,56.550) and (64.340,53.350) .. (66.220,53.650) .. controls (68.080,53.950) and (69.450,58.730) .. (65.540,61.660) .. controls (61.620,64.590) and (66.320,66.580) .. (67.820,63.830) .. controls (69.330,61.080) and (73.450,54.010) .. (75.580,52.650) .. controls (77.710,51.300) and (79.220,52.050) .. (78.710,54.850) .. controls (78.210,57.640) and (69.290,64.400) .. (70.160,65.850) .. controls (71.020,67.320) and (74.080,64.140) .. (74.080,64.140) .. controls (74.080,64.140) and (83.660,55.430) .. (85.740,57.700) .. controls (87.820,59.970) and (84.160,61.870) .. (78.940,65.030) .. controls (73.710,68.190) and (73.310,69.030) .. (74.040,70.230) .. controls (74.790,71.430) and (86.320,61.700) .. (87.400,65.830) .. controls (88.480,69.940) and (75.640,71.130) .. (76.430,73.980) .. controls (77.230,76.830) and (85.480,68.600) .. (87.170,71.800) .. controls (88.860,75.010) and (75.520,78.780) .. (75.410,78.810) .. controls (71.100,79.930) and (60.150,82.300) .. (56.330,76.690) -- (56.330,76.690) -- cycle;}}}
\newcommand{\lbIconHome}{\raisebox{-1.6pt}{\includegraphics[height=9.4pt]{brand/variants/mark-text-color.pdf}}}
\newcommand{\lbButton}[6]{\href{#1}{\tikz[baseline=(lb.base)]\node[fill=#4, draw=#5, line width=0.45pt,
    rounded corners=6.4pt, inner xsep=6.5pt, inner ysep=3.1pt, text=#6, font=\sffamily\fontsize{8.5}{10}\selectfont]
    (lb){#2\hspace{4.2pt}#3\vphantom{Hg}};}}

\newcommand{\lbButtonsSolid}{{\noindent
  \lbButton{\lbUrlHome}{\lbIconHome}{Homepage}{white}{lbHome}{lbHome!85!black}\hspace{6pt}%
  \lbButton{\lbUrlHF}{\lbIconHF}{Hugging Face}{lbHFYellow}{lbHFYellow}{lbHFDark}\hspace{6pt}%
  \lbButton{\lbUrlGitHub}{\lbIconGitHub[white]}{GitHub}{lbGitHub}{lbGitHub}{white}\par}}

\makeatletter
\newcommand{\heroLogos}{\raisebox{\dimexpr4.85pt-0.5\height\relax}{\includegraphics[height=5.5mm]{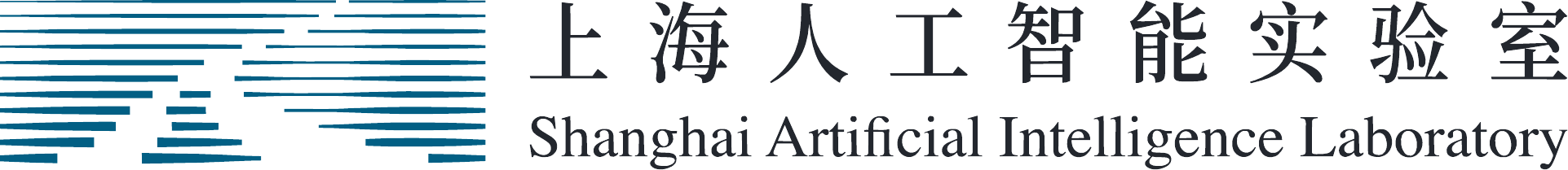}}\hspace{6pt}}
\renewcommand{\fm@firsthead}{\fm@lockup{color}{17.5}{hlInk}}
\def\ps@fmfirst{%
  \def\@oddhead{\lower\heroHeadDrop\hbox to\textwidth{\fm@firsthead\hfil\heroLogos}}%
  \def\@oddfoot{\hbox to\textwidth{\hfil\sffamily\hl@headsize\color{hl-running-head}\hltabnums\hl@folio}}%
  \let\@evenhead\@oddhead \let\@evenfoot\@oddfoot}
\newcommand{\heroHeadDrop}{6pt}\newcommand{\heroTop}{22.8pt}\newcommand{\heroTitleLead}{35}\newcommand{\heroTitlePeople}{20pt}
\newcommand{\heroPeopleLinks}{16pt}\newcommand{\heroLinksRule}{22.5pt}\newcommand{\heroRuleAbstract}{22.5pt}
\renewcommand{\fm@authors}{{\sffamily\hl@size{10.5}{14}\color{hlInk}\RaggedRight\def\and{\unskip,\hspace{0.6em}}\@author\par}%
  \vspace{3pt}{\sffamily\small\color{hlMid}\RaggedRight\hl@affil\par}%
  \vspace{\heroPeopleLinks}\lbButtonsSolid}
\renewcommand{\fm@masthead}{\nointerlineskip\vbox{\hsize\textwidth\linewidth\textwidth\parindent\z@
  \vspace*{\heroTop}\fm@title{30}{\heroTitleLead}\vspace{\heroTitlePeople}\fm@authors
  \par\nointerlineskip\vskip\heroLinksRule{\color{hlHair}\hrule height\hlRuleHair}}\nointerlineskip\vskip\dimexpr\heroRuleAbstract-\hlPrintReportBodyLead pt/2\relax}   %
\renewcommand{\needspace}[1]{\par}
\newcommand{\hl@internnote}{}
\newcommand{\internnote}[1]{\renewcommand{\hl@internnote}{#1}}
\renewcommand{\maketitle}{\par\hl@titlestart\begingroup\setlength{\parindent}{0pt}\setlength{\parskip}{0pt}%
  \hl@pdfinfo\thispagestyle{fmfirst}%
  \csname fm@\fm@dir\endcsname
  \endgroup
  \begingroup\renewcommand{\thefootnote}{}%
    \ifx\hl@internnote\@empty\else\footnotetext{\textsuperscript{\dag}\,\hl@internnote}\fi
    \ifx\hl@corr\@empty\else\footnotetext{\textsuperscript{*}\,Corresponding author: \hl@corr.}\fi
  \endgroup
  \hl@titleend}
\makeatother

\hairlinesetup{short-title={LoopTitan: What Does Parameter Revisit Buy a Mixture-of-Experts?}, kind={},
  keywords={parameter revisit, looped transformer, mixture of experts}}
\projectname{LoopTitan}                 %
\title{LoopTitan: What Does Parameter Revisit Buy a Mixture-of-Experts?}
\displaytitle{\fmflush{L}oopTitan: What Does Parameter \\\fmflush[0.082]{R}evisit Buy a Mixture-of-Experts?}
\author{Zichen Tian\affilmark{1,2}\textsuperscript{\,\dag} \and Yudong Lu\affilmark{1} \and Xuansheng Wu\affilmark{1} \and Na Zou\affilmark{1}
  \and Xia Hu\affilmark{1} \and Guanchu Wang\affilmark{1}\textsuperscript{*} \and Qianru Sun\affilmark{2}}
\internnote{Work done during an internship at Shanghai Artificial Intelligence Laboratory.}
\corresponding{Guanchu Wang}
\affiliations{\affilmark{1}Shanghai Artificial Intelligence Laboratory\\\affilmark{2}Singapore Management University}

\begin{document}
\maketitle
\makeatletter\let\mainfb@saved\@fb@secFB \let\@fb@secFB\relax\makeatother
\begin{abstract}
Looped Mixture-of-Experts (MoE) architectures revisit layer parameters across depths.
This design increases effective depth without adding parameters.
Recent looped MoE studies report substantial gains over the standard MoE at matched budgets (parameters, FLOPs and KV cache).
However, the reported gains still confound parameter revisit with other design choices, \emph{e.g.}, network shape and residual scale.
We ask, \textbf{\emph{what does parameter revisit itself buy?}}
\par\smallskip\noindent%
Using matched non-revisit controls, we find that \textbf{parameter revisit itself buys almost nothing}.
We argue that \textbf{revisit only offers a freedom an MoE already has}, \emph{i.e.}, trading experts for depth (more layers, fewer experts each) at fixed parameters.
First, at matched budgets, we decompose the state-of-the-art looped recipe into key design choices, with revisit last.
\textbf{A non-revisit control retains all the gain.}
Most of the gain comes from network shape (a deeper, narrower stack with finer experts), and the rest from residual scale.
We prove this scale acts as a lower learning rate on revisited layers.
This helps under AdamW's uneven layer step sizes, but disappears under the Muon optimizer, which keeps them nearly equal.
Second, beyond matched budgets, we evaluate revisit by adding compute at fixed parameters.
One extra visit gains more than doubling the experts.
Yet without revisit, a deeper standard MoE at equal parameters and compute obtains \textbf{90.2\%} of that gain under Muon.
\par\smallskip\noindent
Overall, the gains come from network shape, the optimizer and added compute, which a standard MoE can adopt.
Accordingly, we propose \textbf{\Base} as a stronger MoE baseline for evaluating looped architectures. Under Muon, it matches the state-of-the-art looped recipe without revisit or residual scale.
\par\smallskip\noindent%
We open-source our infrastructure \moeloop{} for budget-matched looped MoE research and release 17 pretrained models (including \textbf{\Base} and the non-revisit control).
Our full experiment record is public as GitHub issues, with predictions pre-registered and the full exploration history kept.
This is our Phase 1 technical report, laying the foundation for follow-up studies.
\end{abstract}

\section{Introduction}
\label{sec:intro}

Mixture-of-Experts (MoE) architectures~\citep{shazeer2017,fedus2022} power leading language models~\citep{deepseekv3,qwen3}. By activating only a few experts per token, MoEs scale stored parameters at nearly constant compute (expert scaling)~\citep{abnar}.
Conversely, looped Transformers~\citep{huginn,parcae,ouro} revisit layer parameters across depths~\citep{ut,albert}, scaling effective depth and compute without adding stored parameters.
Combining both designs, a growing line of work develops looped MoEs~\citep{moeut,dreamer,sparselayers,loopmoe,tying,loopie,smelt,foil,metaloop,loom,elr,latentloop} and treats visits (passes over revisited layers) as a new scaling axis~\citep{loom,metaloop,foil,elr}.

\begin{figure}[t]
\fig{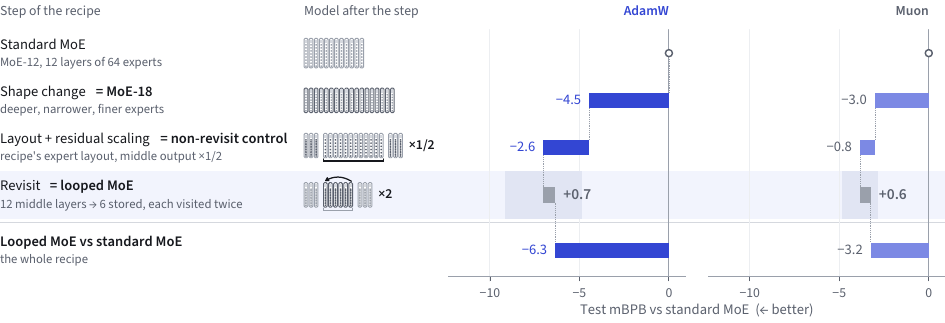}{Waterfall chart with one row per step of the looped recipe and a schematic of the model after each step, AdamW on the left and Muon on the right. Each bar starts where the bar of the row above ended, so the bars descend like a staircase from zero to the total. The axis is test mBPB relative to the standard MoE, and lower is better. Gray bars raise the loss. Every value is a difference of three-seed means. Standard MoE, MoE-12, 12 layers of 64 experts, the zero point. Shape change, which gives MoE-18, AdamW \gapUMzeroAdamW{} and Muon \gapUMzeroMuon. Layout and residual scaling, more experts at the ends and fewer in the middle and the middle layers' output halved, which gives the non-revisit control, AdamW \gapMfourUAdamW{} and Muon \gapMfourUMuon. Revisit, which gives the looped MoE, AdamW \gapMthreeMfourAdamW{} and Muon \gapMthreeMfourMuon, short gray bars inside a light band centred on the non-revisit control's level, plus or minus \seedTwoSeThreeThreeAdamW{} under AdamW and \seedTwoSeThreeThreeMuon{} under Muon. The whole recipe, looped MoE against standard MoE, one bar from zero, AdamW \gapMthreeMzeroAdamW{} and Muon \gapMthreeMzeroMuon.}
\caption{\lead{At matched budgets, all of the looped recipe's gain precedes parameter revisit, arising from network shape and, under AdamW, residual scaling.}
Each row cumulatively applies the next component of the recipe to the model above at matched budgets, with each bar starting where the previous row ends.
Bars report three-seed mean changes in test loss (mBPB, 0.001 bits per byte), where negative values denote gains and gray bars mark costs.
The non-revisit control matches the looped MoE's computed depth using independent parameters (\cref{fig:results-ladder} decomposes each grouped row into individual steps).
Shaded bands on the revisit row span $\pm$\seedTwoSeMthreeMfourAdamW{} (AdamW) and $\pm$\seedTwoSeMthreeMfourMuon\mbpb{} (Muon) around the non-revisit control (two standard errors).
All models have 1.3B stored parameters and train on 5B tokens.}
\label{fig:gain}
\end{figure}

Recent studies report substantial gains for looped MoEs over standard MoEs even at matched stored parameters and compute~\citep{loopmoe,dreamer,tying}.
The state-of-the-art recipes further match the KV cache~\citep{smelt,latentloop} and retain revisit gains across sparsity levels and scales up to 54B non-embedding parameters.
However, these reported gains still confound parameter revisit with accompanying design choices (\cref{fig:gain}), \textit{i.e.}, matching budgets force network shape into a deeper, narrower stack with finer experts. They also apply residual scaling (by $1/2$~\citep{smelt} or learned~\citep{latentloop}).
We therefore ask, \textbf{\emph{what does parameter revisit itself buy an MoE?}}

Using matched non-revisit controls both at matched budgets and with added compute, we find that \textbf{parameter revisit itself buys almost nothing}. We argue that \textbf{revisit offers a freedom an MoE already has}: unlike dense models where revisit decouples per-token compute and stored parameters~\citep{kaplan}, MoE already decouples the two~\citep{abnar}, allowing an MoE to trade experts for depth (more layers, fewer experts each) at fixed stored parameters.

First, at matched budgets (stored parameters, compute and KV cache), we decompose the state-of-the-art looped recipe from a standard MoE into sequential key design choices, \textit{i.e.}, the network shape change, expert layout, residual scaling, and parameter revisit last (\cref{fig:gain}).
The model just before revisit is our \Ctrl{}. It matches the looped MoE in computed layers, residual scaling and budgets, but every layer has its own parameters, so each of its 12 middle layers holds half the experts.
As \cref{fig:gain} shows, \textbf{the \Ctrl{} retains all of the recipe's gain}.
Under AdamW~\citep{loshchilov2019adamw}, the looped MoE lowers test loss by \gapMthreeMzeroAdamWAbs\mbpb{} over the standard MoE and the \Ctrl{} lowers it by \seedDiffMfourMzeroAdamWAbs\mbpb{}, whereas adding revisit slightly raises loss (\textbf{\gapMthreeMfourAdamW\mbpb{}}, within the $\pm$\seedTwoSeThreeThreeAdamW\mbpb{} noise band).
Most of the gain comes from the shape change (\gapUMzeroAdamWAbs\mbpb{}), and the expert layout and residual scaling add \gapMfourUAdamWAbs\mbpb{}.
Across training lengths (5B to 20B tokens) and model sizes (1.3B to 3.0B stored parameters), the shape change remains the largest contributor (\rangeShapeMinAbs{} to \rangeShapeMaxAbs\mbpb{}), and revisit stays within noise up to 24B tokens (\cref{tab:settings}).

To explain the remaining gain, we prove that residual scaling is equivalent to an optimizer setting (a joint rescaling of initialization, learning rate and weight decay, details in \cref{prop:residual}), an instance of a known parametrization symmetry~\citep{yang2022tp5,blake2024umup}.
Consequently, residual scaling does not need parameter revisit, \textit{e.g.}, Even on the \Ctrl{}, it lowers the loss by \seedDiffMfourRAdamWAbs\mbpb{} under AdamW (\cref{fig:residual}a).
Because AdamW normalizes updates entrywise~\citep{bernstein2024oldnorm}, a single global learning rate induces uneven layer step sizes (\cref{tab:results-stepsize}), which a lower effective learning rate on the scaled layers can offset. By contrast, the Muon optimizer~\citep{jordanmuon,moonlight} normalizes whole-matrix updates and equalizes step sizes across layers, so we hypothesize that it needs no per-layer rescaling.
As \cref{fig:gain} confirms, under Muon, residual scaling's gain disappears and the recipe's total gain drops from \gapMthreeMzeroAdamWAbs{} to \gapMthreeMzeroMuonAbs\mbpb{}, all of which the \Ctrl{} still retains (revisit itself, \gapMthreeMfourMuon\mbpb{}).
Moreover, switching the standard MoE from AdamW to Muon alone outperforms the entire looped recipe under AdamW by \textbf{\ratioLeadMzeroGainAdamW{} times} (\leadMzeroAbs{} against \gapMthreeMzeroAdamWAbs\mbpb{}).
These findings hold across our tested scales up to 3.0B stored parameters and 24B tokens.

Second, beyond matched budgets, we evaluate revisit by adding compute at fixed stored parameters. Applying one extra visit to a standard MoE's middle layers (+31\% compute) gains more under Muon than doubling the experts at fixed compute (\gapMzeroLoopMzeroMuonAbs{} against \gapMzeroTwoEMzeroMuonAbs\mbpb{}).
Yet without revisit, a \DeepCtrl{} at equal parameters and compute obtains \textbf{\shareMzeroDeepofMzeroLoopMuon\%} of that gain (\gapMzeroDeepMzeroMuonAbs\mbpb{}, \cref{fig:compute}).
Like the \Ctrl{}, the \DeepCtrl{} uses independent parameters and trades experts per layer (42 instead of 64) for additional layers, leaving revisit itself with only \seedDiffMzeroLoopMzeroDeepMuonAbs\mbpb{} (about one-tenth of the gain).

Overall, the gains come from network shape, the optimizer and added compute, all of which a standard MoE can adopt.
Accordingly, we propose \Base{} as a stronger MoE baseline for evaluating looped architectures (\cref{fig:base}).
Without revisit or residual scaling, \Base{} already matches the state-of-the-art looped MoE under Muon (\seedDiffMthreeUMuonAbs\mbpb{} difference).

We open-source our infrastructure \moeloop{} for budget-matched looped MoE research and release 17 pretrained models, including \Base{} and the \Ctrl{}.
\moeloopname{} unifies nine published looped recipes as configurations of a single model and verifies parameter, compute and KV-cache budgets before training (details in \cref{sec:looptitan-config,sec:system-app}).
Future looped designs can thus be specified as one configuration and benchmarked directly against \Base{} and a matched non-revisit control.
Alongside the codebase, we publish our full experiment record (\recordRuns{} runs, \recordStudies{} studies) as GitHub issues from an auditable collaboration of researchers and AI agents, with pre-registered criteria and the full exploration history, including failed hypotheses (details in \cref{sec:system-protocol}).
This is our Phase 1 technical report, establishing the controlled baseline and open infrastructure for our follow-up studies.

\section{Isolating revisit}
\label{sec:looptitan}

The loss difference between a looped model and its non-revisit control measures the value of revisit.
Beyond matched budgets, where revisit adds compute, we construct a non-revisit deeper control at equal parameters and compute.
All models are configurations of \moeloop{}, our open looped-MoE infrastructure (details in \cref{sec:looptitan-config}).

\subsection{The looped recipe}
\label{sec:looptitan-recipe}

A looped MoE computes its middle layers repeatedly with the same parameters (parameter revisit).
Each pass is a visit.
A standard MoE computes each layer once with independent parameters.
In both models, each layer adds attention and MoE branches to the residual stream $h$,
\begin{equation}
h \leftarrow h + a\,\mathrm{Attn}\bigl(\mathrm{Norm}(h)\bigr), \qquad h \leftarrow h + a\,\mathrm{MoE}\bigl(\mathrm{Norm}(h)\bigr),
\label{eq:setup-block}
\end{equation}
with residual scale $a=1$ by default.
Each branch ends in an output projection.

Matched budgets keep stored parameters, compute and inference cache each within 5\% of the standard MoE at the same scale.
Stored parameters exclude the embedding and output head and count once across visits.
We measure compute as training FLOPs and inference cache as KV cache, both per token.
Like SMELT~\citep{smelt}, we match all three resources (\cref{sec:setup-app-budgets}).
A revisited layer counts toward compute and cache at each visit.
Revisit therefore saves only stored parameters, and extra visits cost compute.

Extra visits increase depth, and the state-of-the-art recipe chooses its depth by ablation~\citep{smelt}.
The recipe narrows the model to match compute and restores stored parameters with finer experts (more in total, each smaller).
Our looped MoE applies this recipe to \MoEtw{}, a 12-layer standard MoE of width 1280 with 64 experts of width 432 per layer (\cref{tab:setup-app-arch}).
SMELT~\citep{smelt} trains a standard MoE at 85\% sparsity with 0.2B active parameters per token.
\MoEtw{} matches that model in layers, width, expert count, top-$k$ and stored parameters (\cref{tab:setup-app-smelt}).
The recipe also sets residual scale $a=1/2$ on revisited layers at every visit (residual scaling).
Following the Loop-µP analysis, the recipe uses this scale to keep shared layers' correlated outputs from inflating the residual stream~\citep{smelt,loopmup}.
Comparing directly with \MoEtw{} therefore cannot separate revisit from the shape change and residual scaling.

\begin{figure}[t]
\fig{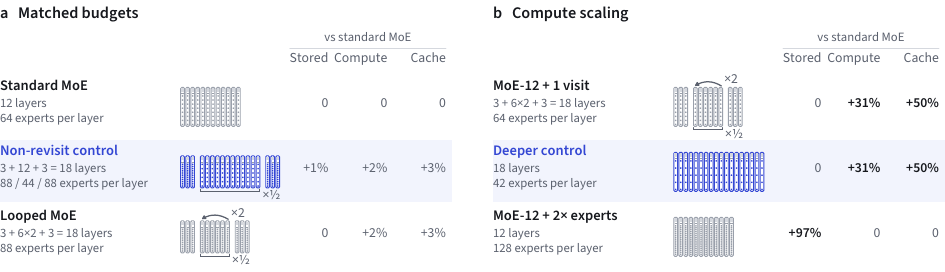}{Two panels of layer schematics. Each model is listed with its layers, its experts per layer and its resources relative to the standard MoE with 1.3B stored parameters, in the order stored parameters, compute, inference cache. Panel a, matched budgets. Standard MoE, 12 layers, 64 experts per layer, 0, 0, 0. Non-revisit control, 3 + 12 + 3 = 18 layers, 88 / 44 / 88 experts per layer, residual scale one half on the middle 12 layers, +1\%, +2\%, +3\%. Looped MoE, 3 + 6 x 2 + 3 = 18 layers, 88 experts per layer, middle 6 layers computed twice with residual scale one half, 0, +2\%, +3\%. Panel b, compute scaling. MoE-12 + 1 visit, 3 + 6 x 2 + 3 = 18 layers, 64 experts per layer, middle 6 layers computed twice with residual scale one half, 0, +31\%, +50\%. Deeper control, 18 layers, 42 experts per layer, 0, +31\%, +50\%. MoE-12 + 2x experts, 12 layers, 128 experts per layer, +97\%, 0, 0.}
\caption{\lead{Matched non-revisit controls remove parameter revisit by trading experts per layer for independent layers, both at matched budgets (\textbf{a}) and with added compute (\textbf{b}).}
\textbf{a}~At matched budgets, the \Ctrl{} matches the looped MoE in resources, computed depth and residual scaling, using independent parameters. \textbf{b}~With added compute, the \DeepCtrl{} matches an extra visit's compute via additional layers with independent parameters, fewer experts each and no residual scaling.
Bars depict layers (height indicates hidden width), and each dot represents eight experts.
$\times2$ denotes revisit and $\times1/2$ denotes residual scaling.
Resource changes (stored parameters, compute and inference cache) are reported relative to \MoEtw.}
\label{fig:control}
\end{figure}

\subsection{A non-revisit control}
\label{sec:looptitan-control}

At matched budgets, we apply the recipe to \MoEtw{} in five steps, with revisit last (\cref{tab:setup-arms}).
The first two steps form the shape change (finer experts and a deeper, narrower stack).
They yield \MoEei{} (18 layers of width 1056, each with 58 experts of width 384).
The third step gives \MoEei{} the recipe's expert layout (more experts in the outer layers than in the middle ones), with six outer layers of 88 experts and 12 middle layers of 44.
The fourth adds residual scaling to these middle layers.
The fifth, revisit, replaces them with six 88-expert layers computed twice.
The model before revisit is the \Ctrl{}.

The \Ctrl{} matches the looped MoE in resources and computed layers, but uses independent parameters (\cref{fig:control}a).
The Foil study names such a control as missing~\citep{foil}.
The looped MoE and the \Ctrl{} compute 18 layers of width 1056 with identical attention, expert width, top-8 routing, residual scaling and training.
They differ in whether middle layers share parameters.
Without revisit, the same stored experts span twice as many middle layers, so each holds 44 experts instead of 88.
The looped MoE reaches this depth by revisit and the \Ctrl{} by spreading experts over more layers.
Therefore, the looped MoE's loss minus the \Ctrl{}'s measures revisit itself.
Six extra attention blocks give the \Ctrl{} 1.1\% more parameters, favoring it.

\subsection{A deeper control for added compute}
\label{sec:looptitan-compute}

\MoEloop{} computes the middle six layers of \MoEtw{} twice with residual scaling (\cref{fig:control}b).
The extra visit adds \budFlopsMzeroLoopInt\% compute and \budKvMzeroLoopInt\% KV cache at fixed stored parameters.
The \DeepCtrl{} matches this compute and cache without revisit, using 18 layers of width 1280 with independent parameters.
Each layer holds 42 experts instead of 64, so stored parameters stay fixed.
Unlike the \Ctrl{}, the \DeepCtrl{} has neither the recipe's expert layout nor residual scaling.
Doubling the experts is the reference for expert scaling.
\MoEtwoE{} doubles experts per layer, with \budParamsMzeroTwoEInt\% stored parameters at fixed compute and cache.

\subsection{Training and how to read a difference}
\label{sec:looptitan-setup}

\term{Optimizers and learning rates}
We run every matched-budget comparison under both AdamW~\citep{loshchilov2019adamw} and Muon~\citep{jordanmuon,moonlight}.
Muon runs optimize attention and expert matrices with Muon and other parameters with AdamW (\cref{tab:setup-app-train}).
Grid searches over full 5B-token runs of \MoEtw{} and the looped MoE calibrate one peak learning rate per optimizer (\cref{sec:results-lr}).
The searches on both models select \selLrAdamW{} under AdamW and \selLrMuon{} under Muon.
All other models (including the \Ctrl{} and \Base{}) reuse these rates without search (one exception, \cref{sec:results-recipes}), so any tuning advantage favors \MoEtw{} and the looped MoE.
The 10B and 20B runs keep these rates.
The larger 3.0B model uses \selLrSzZeroPFiveAdamW{} under both optimizers (\cref{sec:setup-app-train,sec:results-lr}).

\term{Main setting}
The main setting uses 1.3B stored parameters (0.2B active) and 5B tokens.
For longer training, we continue the 5B runs of \MoEtw{}, \MoEei{}, the \Ctrl{} and the looped MoE to 10B and 20B tokens.
The larger model (3.0B stored parameters, 0.5B active) trains separately on 12B tokens (\cref{tab:settings}).
All runs share data (DCLM-baseline~\citep{dclm}), tokenizer (Qwen3~\citep{qwen3}), batch size and schedule shape (\cref{tab:setup-app-train}).

\term{Reading a difference}
We report test loss in bits per byte (BPB) at the last update, using the validation split only to select learning rates.
A gain is a test-loss reduction and a cost an increase, in mBPB (0.001 BPB).
A difference $A-B$ is negative when $A$ has lower loss.
\MoEtw{} with AdamW reaches \testMzeroAdamW{} BPB (seed 42), so 1\mbpb{} is about 0.1\% of the loss.

All models use seed 42.
In the main setting, nine models also use seeds 43 and 44 (\cref{sec:results-seeds}).
Between two of these models, we report differences of three-seed means.
Other differences use seed 42 for both models unless stated.
A difference below two standard errors ($2\sqrt{s_A^2/n_A+s_B^2/n_B}$ for $n_A$ and $n_B$ seeds) is within noise, and we say the two models match.
Otherwise, the difference is beyond noise.
Here $s_A$ and $s_B$ are the run-to-run standard deviations under each model's optimizer, \seedSdAdamW\mbpb{} for AdamW and \seedSdMuon{} for Muon.
We estimate them separately because they differ by a factor of two (\cref{sec:results-seeds}).
With one or three seeds each, two standard errors are therefore \seedTwoSeOneOneAdamW{} or \seedTwoSeThreeThreeAdamW\mbpb{} under AdamW and \seedTwoSeOneOneMuon{} or \seedTwoSeThreeThreeMuon{} under Muon.

\section{Revisit at matched budgets}
\label{sec:matched}

At matched budgets, parameter revisit itself contributes almost nothing.
The recipe's gain comes from the shape change and, under AdamW, residual scaling.

\begin{table}[ht]
\caption{\lead{Across longer training horizons and a larger 3.0B model, gains continue to arise primarily from network shape rather than parameter revisit.}
Entries report test-loss changes in mBPB (negative values denote gains) at each optimizer's calibrated learning rate.
Step columns correspond to the rows of \cref{fig:gain}.
Total reports the looped MoE minus the standard MoE, from unrounded values.
At 24B tokens, MoE-18 was not trained, so the merged cell gives the non-revisit control minus the standard MoE.}
\label{tab:settings}
\newcommand{\pmse}[1]{\,$\pm$\,\num[round-mode=places,round-precision=1]{#1}}
\newcommand{\nose}{\phantom{\pmse{0.84}}}
\begin{tabular*}{\linewidth}{@{\extracolsep{\fill}}lllNNNN@{\hphantom{\interimDagger}}}
\toprule
Stored parameters & Tokens & Optimizer & Shape change & Layout + residual scaling & \best{Revisit} & Total\\
\midrule
1.3B & 5B & AdamW
 & \gapUMzeroAdamW\pmse{\seGapUMzeroAdamW}\rlap{\interimDagger}
 & \gapMfourUAdamW\pmse{\seGapMfourUAdamW}\rlap{\interimDagger}
 & \best{\gapMthreeMfourAdamW}\pmse{\seGapMthreeMfourAdamW}
 & \gapMthreeMzeroAdamW\pmse{\seGapMthreeMzeroAdamW}\rlap{\interimDagger}\\
 & & Muon
 & \gapUMzeroMuon\pmse{\seGapUMzeroMuon}
 & \gapMfourUMuon\pmse{\seGapMfourUMuon}
 & \best{\gapMthreeMfourMuon}\pmse{\seGapMthreeMfourMuon}
 & \gapMthreeMzeroMuon\pmse{\seGapMthreeMzeroMuon}\\
\addlinespace
 & 10B & AdamW
 & \hgapUMzeroAdamWHzOneZero\nose & \hgapMfourUAdamWHzOneZero\nose & \best{\hgapMthreeMfourAdamWHzOneZero}\nose & \hgapMthreeMzeroAdamWHzOneZero\nose\\
 & & Muon
 & \hgapUMzeroMuonHzOneZero\nose & \hgapMfourUMuonHzOneZero\nose & \best{\hgapMthreeMfourMuonHzOneZero}\nose & \hgapMthreeMzeroMuonHzOneZero\nose\\
\addlinespace
 & 20B & AdamW
 & \hgapUMzeroAdamWHzTwoZero\nose & \hgapMfourUAdamWHzTwoZero\nose & \best{\hgapMthreeMfourAdamWHzTwoZero}\nose & \hgapMthreeMzeroAdamWHzTwoZero\nose\\
 & & Muon
 & \hgapUMzeroMuonHzTwoZero\nose & \hgapMfourUMuonHzTwoZero\nose & \best{\hgapMthreeMfourMuonHzTwoZero}\nose & \hgapMthreeMzeroMuonHzTwoZero\nose\\
\addlinespace
3.0B & 12B & AdamW
 & \gapUMzeroAdamWLarge\nose & \gapMfourUAdamWLarge\nose & \best{\gapMthreeMfourAdamWLarge}\nose & \gapMthreeMzeroAdamWLarge\nose\\
 & & Muon
 & \gapUMzeroMuonLarge\nose & \gapMfourUMuonLarge\nose & \best{\gapMthreeMfourMuonLarge}\nose & \gapMthreeMzeroMuonLarge\nose\\
\addlinespace
 & 24B & AdamW
 & \multicolumn{2}{c}{\hltabnums\gapMfourMzeroAdamWLargeLong} & \best{\gapMthreeMfourAdamWLargeLong}\nose & \gapMthreeMzeroAdamWLargeLong\nose\\
 & & Muon
 & \multicolumn{2}{c}{\hltabnums\gapMfourMzeroMuonLargeLong} & \best{\gapMthreeMfourMuonLargeLong}\nose & \gapMthreeMzeroMuonLargeLong\nose\\
\bottomrule
\end{tabular*}
\tablenote{At 1.3B, the shape change yields \MoEei{}.
At 3.0B, it turns 16 layers of width 2048 into 24 of width 1672.
Layout and residual scaling yield the \Ctrl{} (4 + 16 + 4 outer, middle and outer layers at 3.0B).
Revisit yields the looped MoE.
The 10B and 20B runs continue the 5B runs, and 3.0B runs train separately (learning rates in \cref{sec:looptitan-setup}).
In the 5B rows, cells give differences of three-seed means $\pm$ one standard error (each optimizer's own $s$).
Other rows use seed 42.\interimNote}
\end{table}

\term{The non-revisit control retains all of the recipe's gain}
As in prior work~\citep{smelt}, the looped MoE has substantially lower test loss than \MoEtw{}, by \gapMthreeMzeroAdamWAbs\mbpb{} under AdamW and \gapMthreeMzeroMuonAbs\mbpb{} under Muon (\cref{fig:gain}).
The \Ctrl{} (without revisit) matches the looped MoE.
The \Ctrl{}'s three-seed mean loss is lower by \seedDiffMthreeMfourAdamWAbs\mbpb{} under AdamW and \seedDiffMthreeMfourMuonAbs{} under Muon (\cref{tab:settings}).
Both differences lie within two standard errors and change sign across seeds (\cref{tab:seeds}).
Revisit adds only this small cost (\cref{fig:gain}), so the \Ctrl{} retains all of the recipe's gain.
Yet the trained looped MoE uses its second visit (\cref{sec:obs-readout}).

\term{Most of the gain comes from the shape change}
At matched budgets, the shape change turns \MoEtw{} into \MoEei{} (a deeper, narrower standard MoE with finer experts).
This change gains \gapUMzeroAdamWAbs\mbpb{} under AdamW (\shareUofMthreeAdamW\% of the recipe's gain) and \gapUMzeroMuonAbs\mbpb{} under Muon (almost all of it, \cref{fig:gain}).
Under Muon, most of this gain comes from the deeper, narrower stack (seed 42, \cref{fig:results-ladder}).
Across four depths (12 to 36 layers, one seed each), 18 layers give the lowest loss under both optimizers (\cref{fig:results-depth}).

In SMELT's depth sweep at \MoEtw{}'s scale, the 18-layer standard MoE has 0.5\% higher validation loss than the 12-layer model, whereas the looped model is best with 18 computed layers~\citep{smelt}.
SMELT proposes that shared layers may be easier to optimize than an equally deep independent stack~\citep{smelt} (our curvature measurements in \cref{fig:obs-curvature}).
Over three seeds, \MoEei{} has \gapUMzeroAdamWAbs\mbpb{} lower loss than \MoEtw{} under AdamW and \gapUMzeroMuonAbs{} under Muon (the shape change in \cref{tab:settings}).
The sweeps differ in data, learning-rate schedule and training length (up to about 1,000 tokens per active parameter in SMELT, against \tokPerActiveMain{} for \MoEtw{} in ours).
Why a standard MoE's best depth differs is open.

\term{Under AdamW, the remaining gain comes from residual scaling}
Adding the recipe's expert layout and residual scaling to \MoEei{} yields the \Ctrl{} (\cref{fig:gain}).
Together they gain \gapMfourUAdamWAbs\mbpb{} under AdamW (\verdictGapMfourUAdamW) and \gapMfourUMuonAbs{} under Muon (\verdictGapMfourUMuon).
We separate both steps with seed 42 (\cref{fig:results-ladder}).
Residual scaling gains \seedDiffMfourRAdamWAbs\mbpb{} under AdamW and nothing under Muon (\cref{sec:residual}).

\term{With longer training and a larger model, the result holds}
Across all settings and both optimizers, the shape change gains \gapUMzeroMuonAbs{} to \rangeShapeMaxAbs\mbpb{}, more than any other group of steps (\cref{tab:settings,tab:results-horizon}).
In the 10B-token, 20B-token and both 3.0B settings (one seed each), the revisit step lies between \hgapMthreeMfourAdamWHzTwoZero{} and \gapMthreeMfourMuonLarge\mbpb{}, within two standard errors (\cref{tab:settings}).
The \gapMthreeMfourMuonLarge\mbpb{} step comes from the larger model under Muon, where the \Ctrl{} obtains most of the recipe's gain (\shareMfourofMthreeMuonLarge\%).
Doubling the 3.0B model's training from 12B to 24B tokens leaves the revisit step within noise under both optimizers (\gapMthreeMfourMuonLargeLong{} under Muon and \gapMthreeMfourAdamWLargeLong\mbpb{} under AdamW), and the \Ctrl{} still obtains all of the recipe's gain under AdamW and \shareMfourofMthreeMuonLargeLong\% under Muon.
Whether the revisit step grows at larger model scales is open (\cref{sec:discussion}).

\section{Residual scaling as an optimizer setting}
\label{sec:residual}

We prove residual scaling equivalent to an optimizer setting for the scaled layers' output projections, so it requires no loop.
In \cref{eq:setup-block}, each branch ends in an output projection ($W_{\mathrm{att}}$ for attention, $W_{2,j}$ for expert $j$) without bias or normalization after it, so scaling a branch's output scales its output projections.

\begin{propbox}
\begingroup\renewcommand{\thetheorem}{\arabic{theorem}}%
\begin{proposition}[Residual scaling is an optimizer setting]
\label{prop:residual}
For $a > 0$, training with either tuple below gives the same network function after every update.
\begin{equation*}
\underbrace{\bigl(a;\ W_0,\ \eta_t,\ \lambda\bigr)}_{\text{residual scaling}}
\quad\Longleftrightarrow\quad
\underbrace{\bigl(1;\ aW_0,\ a\,\eta_t,\ \lambda/a\bigr)}_{\text{optimizer setting}}
\end{equation*}
Each tuple lists a layer's residual scale and its output projections' initial value, learning rate and decoupled weight-decay coefficient.
After $t$ updates, the projections $W_t$ (left) and $V_t$ (right) satisfy $V_t = aW_t$.
The equivalence holds on the same batches, with all other settings equal and no gradient-norm clipping.
Under Muon, it holds while the lower bound in Muon's normalization is not reached.
Under AdamW, it holds when the optimizer setting replaces $\varepsilon$ with $\varepsilon/a$.
\end{proposition}
\endgroup
\end{propbox}

\term{Proof sketch}
The network depends on $a$ and an output projection $W$ only through $V = aW$, so the gradient for $W$ is $a$ times that for $V$.
Both optimizers remove this factor from the update direction, except for AdamW's $\varepsilon$ (\cref{lem:proof-scale}).
At learning rate $a\,\eta_t$, $V$ then receives $a$ times $W$'s update at $\eta_t$, matching $aW$.
The coefficient $\lambda/a$ keeps the decay per update ($\eta_t\lambda$) unchanged.

\Cref{prop:residual} instantiates a known parametrization symmetry that equates a weight multiplier with rescaling that weight's initialization and learning rate~\citep{yang2022tp5,blake2024umup}.
Tensor Programs V states this symmetry for Adam without $\varepsilon$ or weight decay~\citep{yang2022tp5}.
We extend it to AdamW with decoupled weight decay and to Muon (\cref{prop:proof-full}).
Residual scaling changes how the scaled output projections are trained, not what the model can represent.
Our runs do not replace AdamW's $\varepsilon$ with $\varepsilon/a$.
The two tuples' updates then differ by a relative amount that vanishes as $\varepsilon \to 0$ (\cref{cor:proof-eps}).
Our gradient-norm clipping also breaks exact equivalence (\cref{sec:proof-breaks}).

\begin{figure}[t]
\fig{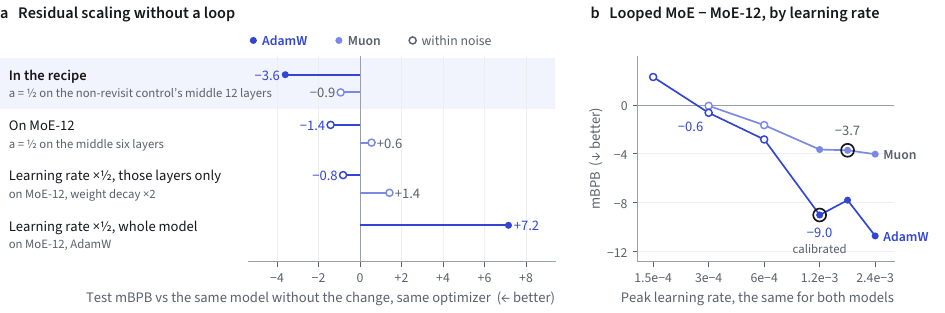}{Two panels. Panel a, lollipop chart with four rows, AdamW above and Muon below in each row, in test mBPB relative to the same model without the change under the same optimizer. A filled dot is beyond noise and a hollow dot within noise. In the recipe, residual scaling with a one half on the non-revisit control's middle 12 layers, AdamW minus 3.6, beyond noise, Muon minus 0.9, within noise. On MoE-12, a one half on the middle six layers, AdamW minus 1.4, Muon plus 0.6, both within noise. Learning rate halved on those layers only, AdamW minus 0.8, Muon plus 1.4, both within noise. Learning rate halved for the whole model under AdamW, plus 7.2, beyond noise. Panel b, line chart of the looped MoE minus MoE-12 in test mBPB against the peak learning rate that both share. AdamW, plus 2.3 at 1.5e-4, minus 0.6 at 3e-4, minus 2.8 at 6e-4, minus 9.0 at the calibrated 1.2e-3, minus 7.8 at 1.7e-3, minus 10.7 at 2.4e-3. Muon, minus 0.1 at 3e-4, minus 1.6 at 6e-4, minus 3.6 at 1.2e-3, minus 3.7 at the calibrated 1.7e-3, minus 4.0 at 2.4e-3.}
\caption{\lead{Even without revisit, residual scaling reduces test loss under AdamW yet remains within noise under Muon (\textbf{a}), while the looped recipe's gain varies with the shared peak learning rate, especially under AdamW (\textbf{b}).}
\textbf{a}~Test-loss change relative to each unmodified model under the same optimizer.
The top row applies residual scaling to the \Ctrl{}'s 12 middle layers, and the lower rows modify the middle six layers or all layers of \MoEtw{}.
\textbf{b}~Test-loss difference between the looped MoE and \MoEtw{} across shared peak learning rates (one seed per point, with rings marking calibrated rates).
\Cref{tab:results-lrgrid} includes \lr{3.4e-3}, where \MoEtw{}'s loss rises sharply under AdamW.
Both panels report test mBPB at 1.3B stored parameters and 5B tokens.
Filled and hollow dots denote differences beyond and within noise (two standard errors for each difference's seeds, \cref{sec:results-seeds}).}
\label{fig:residual}
\end{figure}

\term{Residual scaling requires no loop}
When every visit uses the same scale, \cref{prop:residual} holds whether or not a layer is revisited.
The \Ctrl{} revisits no layer, and residual scaling lowers its loss by \textbf{\seedDiffMfourRAdamWAbs\mbpb{}} under AdamW (three seeds, \cref{fig:residual}a).
On \MoEtw{}, where the scale covers six layers, the change is \gapMzeroRMzeroAdamWAbs\mbpb{} and within noise.

\term{Muon equalizes step sizes across layers, and AdamW does not}
\Cref{prop:residual} holds under both optimizers, so it cannot explain a gain under only one.
The two optimizers differ in step size (how far an update moves a weight matrix in spectral norm).
Adam normalizes each update entry, whereas Muon orthogonalizes each matrix's update, bounding its spectral norm~\citep{bernstein2024oldnorm,yang2023spectral,moonlight}.
We prove that under Muon the ratio of a matrix's step size to the learning rate lies in an interval set by shape alone.
Under AdamW, this ratio depends on the gradient (\cref{lem:proof-step}).
We measure both in training (\cref{tab:results-stepsize}).
Under Muon, every output projection takes nearly the same step per unit learning rate (the 90th percentile is at most 1.06 times the 10th).
Under AdamW, the step differs by up to a factor of two across matrices of one type and between types.
With residual scaling, AdamW's step on the scaled layers keeps its size, so the scale halves what each update contributes, as \cref{prop:residual} states.

We hypothesize that one global learning rate cannot suit layers with uneven step sizes, so AdamW gains from a lower effective rate on the scaled layers, whereas Muon's uniform steps require no such rescaling (\cref{fig:residual}a, \cref{sec:results-residual}).
On \MoEtw{} under AdamW, halving the learning rate of the middle layers' output projections alone, with their weight-decay coefficient doubled as in \cref{prop:residual}, does not raise the loss (\seedDiffMzeroLRMzeroAdamW\mbpb{}, within noise).
Halving the whole model's learning rate, with the weight-decay coefficient kept, raises the loss by \seedDiffMzeroHalfLrMzeroAdamWAbs\mbpb{}.
These layers therefore tolerate a smaller step, whereas the whole model does not.

Under both optimizers, the recipe's overall gain depends on the common learning rate of the looped MoE and \MoEtw{} (\cref{fig:residual}b).
At every rate, the recipe gains less under Muon (\gapMthreeMzeroMuonAbsSFortyTwo{} against \gapMthreeMzeroAdamWAbsSFortyTwo\mbpb{} at calibrated rates, seed 42).

\term{Under Muon, the gain of residual scaling disappears}
Under Muon, residual scaling changes the \Ctrl{}'s loss by \seedDiffMfourRMuon\mbpb{} and \MoEtw{}'s by \seedDiffMzeroRMzeroMuon\mbpb{}, both within noise (\cref{fig:residual}a).
We recorded our Muon predictions publicly before these runs (\cref{sec:system-app-trace}).
By \cref{prop:residual}, the gain of residual scaling comes from an optimizer setting, not revisit.

\section{Trading experts for depth}
\label{sec:depth}

We argue that revisit adds so little at matched budgets because an MoE can already trade experts for depth (spread experts over more layers at fixed stored parameters).
If so, a deeper standard MoE without revisit should capture the gain of an extra visit at equal added compute.

\begin{figure}[t]
\fig{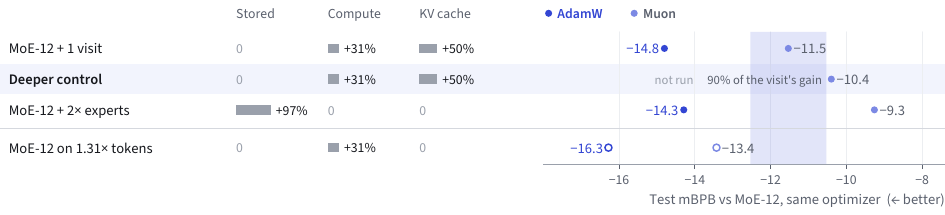}{Table-like chart with four rows, the ways to spend more on MoE-12. Each row has three cost cells with small bars (stored parameters, compute and KV cache, in percent over MoE-12) and a pair of dots for its gain in test mBPB relative to MoE-12 under the same optimizer, AdamW dark and Muon light. Muon dots are three-seed means, AdamW dots one seed. MoE-12 plus one visit, stored 0, compute plus 31\%, cache plus 50\%, AdamW minus \gapMzeroLoopMzeroAdamWAbs, Muon minus \gapMzeroLoopMzeroMuonAbs. Deeper control, emphasised, 0, plus 31\%, plus 50\%, Muon minus \gapMzeroDeepMzeroMuonAbs{}, labeled 90\% of the visit's gain, and not run under AdamW. MoE-12 with twice the experts, plus 97\%, 0, 0, AdamW minus \gapMzeroTwoEMzeroAdamWAbs, Muon minus \gapMzeroTwoEMzeroMuonAbs. Set apart below, MoE-12 on 1.31 times the tokens, 0, plus 31\%, 0, with hollow interpolated points, AdamW minus \tokGainMzeroVisitFlopsAdamWAbs, Muon minus \tokGainMzeroVisitFlopsMuonAbs. A band of plus or minus \seedTwoSeMzeroLoopMzeroDeepMuon{} around the visit's Muon value ends just short of the deeper control's Muon point, and the Muon points of twice the experts and of more tokens lie outside it on either side.}
\caption{\lead{Under Muon, one extra visit outperforms doubling the experts, yet a non-revisit deeper control at equal parameters and compute captures \shareMzeroDeepofMzeroLoopMuon\% of that gain.}
Each row scales \MoEtw{} (1.3B stored parameters, 5B tokens) and reports added stored parameters, training compute and KV cache.
Points show test-loss changes in mBPB (negative values denote gains) relative to \MoEtw{} under the same optimizer.
The bottom row allocates the extra visit's compute to additional training tokens, with interpolated points (\cref{tab:results-tokens}).
The \DeepCtrl{} is trained under Muon only, and only \MoEloop{} uses residual scaling.
Filled Muon points report three-seed means, and all other points use seed 42.
The shaded band spans $\pm$\seedTwoSeMzeroLoopMzeroDeepMuon\mbpb{} around the extra visit's Muon mean (two standard errors, three seeds each), and the \DeepCtrl{} lies just outside it.}
\label{fig:compute}
\end{figure}

\term{An MoE can already trade experts for depth}
In a dense model, every token uses every parameter, so compute grows with stored parameters~\citep{kaplan,abnar}.
Revisit decouples the two by adding compute without adding stored parameters.
In an MoE, sparsity already decouples them~\citep{abnar}, so holding fewer experts per layer frees parameters for more layers.
At SMELT's matched budgets, narrowing the looped MoE offsets the extra visit's compute.
Sparsity then lets the looped MoE restore stored parameters with more experts~\citep{smelt}.

The \Ctrl{} trades experts for depth (\cref{sec:looptitan-control}, \cref{fig:control}a).
Therefore, an MoE can reach a looped MoE's depth without revisit.
This trade needs enough experts per layer.
Each middle layer of the \Ctrl{} still holds 44 experts for top-8 routing.

\term{One extra visit outperforms doubling the experts}
We test this argument with three models starting from \MoEtw{} (\cref{sec:looptitan-compute}, \cref{fig:control}b).
\MoEloop{} adds compute by revisit, and the \DeepCtrl{} adds equal compute by trading experts for depth.
\MoEtwoE{} is the expert-scaling reference.

\MoEtwoE{} gains \gapMzeroTwoEMzeroMuonAbs\mbpb{} over \MoEtw{} under Muon and \gapMzeroTwoEMzeroAdamWAbs{} under AdamW (\cref{fig:compute}).
\MoEloop{} gains \gapMzeroLoopMzeroMuonAbs\mbpb{} under Muon and \gapMzeroLoopMzeroAdamWAbs{} under AdamW.
Under Muon, one extra visit outperforms doubling the experts (\seedVerdictMzeroLoopMzeroTwoEMuon).
Under AdamW, the two gains match (one seed each).
\MoEloop{} also has residual scaling.
Residual scaling alone changes \MoEtw{}'s loss within noise under both optimizers (\cref{fig:residual}a).
With the visit's added compute, \MoEtw{} can train on \flopsRatioMzeroLoopTwoDec{} times the tokens.
Under Muon, this longer training would gain at least as much as the visit (\tokGainMzeroVisitFlopsMuonAbs\mbpb{} interpolated, seed 42, \cref{fig:compute,tab:results-tokens}).

\term{Under Muon, the deeper control captures most of the extra visit's gain}
The \DeepCtrl{} gains \gapMzeroDeepMzeroMuonAbs\mbpb{} over \MoEtw{}, \shareMzeroDeepofMzeroLoopMuon\% of the visit's gain (\cref{fig:compute}).
Besides having no revisit, the \DeepCtrl{} has 42 experts per layer (64 in \MoEloop{}) and no residual scaling.
Over three seeds, the loss of \MoEloop{} is \seedDiffMzeroLoopMzeroDeepMuonAbs\mbpb{} below the \DeepCtrl{}'s and lower in every seed (\cref{tab:seeds}).
This difference is \seedVerdictMzeroLoopMzeroDeepMuon{}, about a tenth of the visit's gain.
With added compute, most of the gain therefore requires no revisit (\cref{sec:results-compute}).

\section{\Base}
\label{sec:base}

\Base{} combines the recipe's network shape with Muon into a stronger MoE baseline for evaluating looped architectures.
Under Muon, \Base{} matches the state-of-the-art looped MoE without revisit or residual scaling.

\Base{} is \MoEei{} (no loop or residual scaling) trained with Muon at learning rate \selLrMuon{} (\cref{tab:setup-app-arch,tab:setup-app-train}).
This baseline is at matched budgets with \MoEtw{}.
At matched compute, \MoEei{} takes longer to train than \MoEtw{} (GPU hours in \cref{tab:results-5b}).
We release 17 pretrained models, including \Base{} and the \Ctrl{}, with their \moeloop{} configurations (\cref{sec:system-app,tab:system-app-models}).

\begin{figure}[t]
\fig{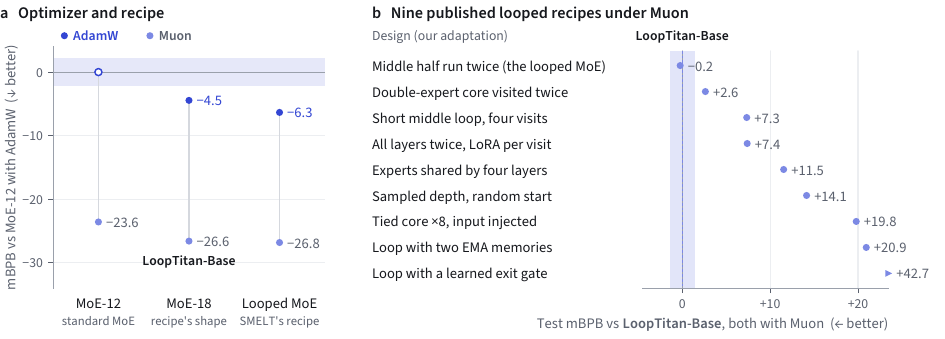}{Two panels. Panel a, three vertical pairs of points, AdamW above and Muon below, in test mBPB relative to MoE-12 trained with AdamW, each a difference of three-seed means. MoE-12, the standard MoE, AdamW 0, Muon \leadMzero. MoE-18, the recipe's shape, AdamW \gapUMzeroAdamW, Muon \xgapUMuonMzeroAdamW, the Muon point labeled LoopTitan-Base. Looped MoE, SMELT's recipe, AdamW \gapMthreeMzeroAdamW, Muon \seedDiffMthreeUMuon{} from LoopTitan-Base. Panel b, one row per published looped recipe in our adaptation, each named by its design and trained with Muon, in test mBPB relative to LoopTitan-Base (three-seed mean), ordered from the closest. Middle half run twice (the looped MoE, three seeds) \seedDiffMthreeUMuon. Double-expert core visited twice plus 2.6. Short middle loop with four visits plus 7.3. All layers twice with LoRA per visit plus 7.4. Experts shared by four layers plus 11.5. Sampled depth from a random start plus 14.1. Tied core run eight times with the input injected plus 19.8. Loop with two EMA memories plus 20.9. Loop with a learned exit gate plus 42.7, off scale. No recipe is below the band.}
\caption{\lead{Switching a standard MoE from AdamW to Muon yields a larger gain than the entire looped recipe under AdamW (\textbf{a}), and under Muon, \Base{} matches the state-of-the-art looped MoE without revisit or residual scaling (\textbf{b}).}
\textbf{a}~Test-loss changes for each model under AdamW and Muon, relative to \MoEtw{} under AdamW.
\textbf{b}~Nine published looped designs implemented in \moeloop{}, adapted to matched scale and budgets (\cref{tab:results-recipes-config}) and evaluated relative to \Base{} under Muon (reflecting our unified adaptations rather than each paper's original setting).
Values report test mBPB at calibrated learning rates (1.3B stored parameters, 5B tokens), using three-seed means for \MoEtw{}, \MoEei{} (\Base{}) and the looped MoE, and one seed for the remaining eight designs.
Shaded bands around zero span two standard errors, $\pm$\twoSeGapMthreeMzeroAdamW\mbpb{} in panel~\textbf{a} (AdamW points, three-seed model against \MoEtw{} under AdamW) and $\pm$\seedTwoSeThreeOneMuon\mbpb{} in panel~\textbf{b} (one-seed recipe against three-seed \Base{}).}
\label{fig:base}
\end{figure}

\term{Switching to Muon outperforms the entire looped recipe}
In the main setting, switching \MoEtw{} from AdamW to Muon (at calibrated learning rates) gains \leadMzeroAbs\mbpb{}, \ratioLeadMzeroGainAdamW{} times the recipe's gain under AdamW (\cref{fig:base}a).
In every setting, switching the standard MoE to Muon gains more than the recipe under AdamW.
This margin shrinks with longer training.
At 20B tokens (our longest 1.3B runs), Muon still gains \hleadMzeroHzTwoZeroAbs\mbpb{} on \MoEtw{}, \ratioLeadMzeroGainAdamWHzTwoZero{} times the recipe's gain under AdamW (\cref{tab:results-base-horizon}).
By interpolation, Muon is worth about \tokEqMuonSwitchMain{} times the training tokens at 5B and \tokEqMuonSwitchHzTwoZero{} times at 20B (seed 42, \cref{tab:results-tokens}).
In dense models, gains from matrix optimizers such as Muon over fully tuned AdamW shrink with model size~\citep{wen2025}.
Architectural gains also shrink when the baseline is well trained~\citep{melis2018,kaddour2023}.

\term{Under Muon, \Base{} matches the looped MoE}
Our looped MoE follows SMELT~\citep{smelt}, the state-of-the-art recipe.
\Cref{fig:base}b shows the nine published designs that \moeloop{} implements, each adapted and trained under one protocol (\cref{sec:results-recipes}).
These runs record the present state of our implementations, not the designs in their original settings.

Under Muon, the looped MoE minus \Base{} is \seedDiffMthreeUMuon\mbpb{} in the main setting (three seeds each, \cref{fig:base}b, \cref{sec:results-seeds}).
In the larger model, this difference is \gapMthreeUMuonLarge\mbpb{} (one seed each).
Both are within two standard errors (\seedTwoSeMthreeUMuon{} and \seedTwoSeOneOneMuon\mbpb{}), as are the differences at 10B and 20B tokens (\cref{tab:results-base-horizon}).
Under AdamW, the looped MoE gains \seedDiffMthreeUAdamWAbs\mbpb{} over \MoEei{} (\seedVerdictMthreeUAdamW).
The \Ctrl{} (without revisit) gains \seedDiffMfourUAdamWAbs\mbpb{} over \MoEei{} (\seedVerdictMfourUAdamW), so the looped MoE's gain does not come from revisit.

\section{\texorpdfstring{\moeloop}{LoopTitan}: one model, many configurations}
\label{sec:looptitan-config}

\begin{figure}[t]
  \fig{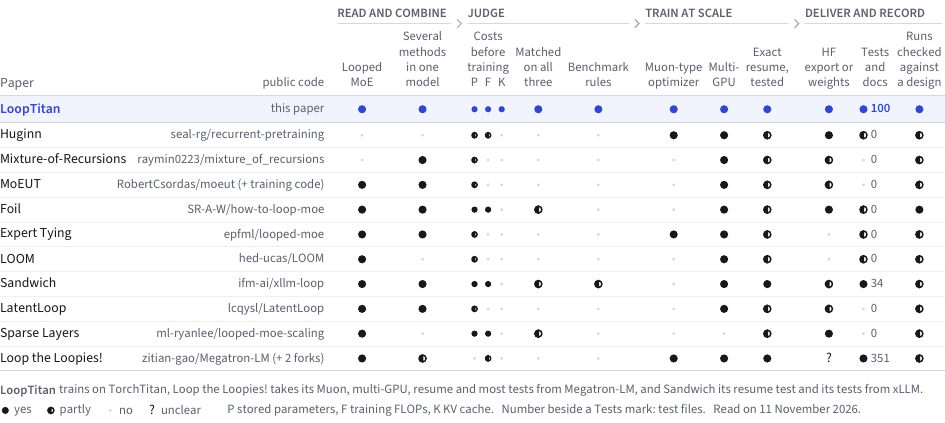}{A dot matrix. Rows: LoopTitan, tinted, then the public code of ten related papers (Huginn,
    Mixture-of-Recursions, MoEUT, Foil, Expert Tying, LOOM, Sandwich, LatentLoop, Sparse Layers, Loop the
    Loopies!), each with its repository. Columns in four groups, joined by small chevrons from left to right: read and
    combine (looped MoE; several methods in one model), judge (costs before training, with separate marks for
    parameters, FLOPs and KV cache; matched on all three; benchmark rules), train at scale (Muon-type optimizer;
    multi-GPU; exact resume, tested), deliver and record (Hugging Face export or weights; tests and docs, with the
    number of test files; runs checked against a design). Marks are yes, partly, no or unclear. Only the LoopTitan row
    is filled in every column. A line below credits the frameworks the code bases build on, TorchTitan for LoopTitan,
    Megatron-LM for Loop the Loopies! and xLLM for Sandwich.}
  \caption{\lead{\moeloop{} unifies looped-MoE configuration, pre-run budget verification, scalable training and reproducible release in one codebase.}
    Rows compare \moeloopname{} with the ten related papers that provide public training code (six additional papers provide none), based on repositories inspected on 11 November 2026 (commits in \cref{tab:system-app-repos}).
    Marks evaluate publicly released code as of that date rather than the papers' experiments (\cref{sec:system-app-public}).
    \moeloopname{} and two other codebases build on larger frameworks, credited under the matrix.}
  \label{fig:system-capabilities}
\end{figure}

One model class implements every configuration in this paper, including both controls, \Base{} and the nine published looped recipes (\cref{sec:results-recipes}).
\moeloopname{} computes each configuration's stored parameters, compute and KV cache before training (\cref{sec:system-code}).
A checker compares each run's recorded configuration with its declared design and flags undeclared differences (\cref{sec:system-app-checker}).
\Cref{fig:system-capabilities} compares \moeloopname{} with the public training code of related papers as inspected on 11 November 2026, not with the papers' experiments.
The rows below \moeloopname{} show the ten related papers with public training code (six more have none).
Of the nine recipes we implement, five come from these ten and four from the other six.
Every study, run and decision is a public issue or record in the \moeloopname{} repository (\cref{sec:system-protocol}).

\section{Related work}
\label{sec:related}

\paragraph{Looped Transformers}
In dense models, gains from looping come from extra compute, not parameter sharing.
Looped Transformers apply one block repeatedly~\citep{ut,albert,huginn,ouro}.
At equal parameters, more visits help~\citep{kaplan,parcae,loopedright}.
At equal compute, unshared models (storing more parameters) achieve lower loss~\citep{kaplan,isodepth,exponents,mor}.
SMELT~\citep{smelt} notes that the looped model in one such study~\citep{isodepth} may underperform due to having fewer parameters rather than looping.

\paragraph{Looped MoEs}
Besides SMELT~\citep{smelt}, looped MoEs share layer groups~\citep{moeut}, share experts across layers~\citep{tying,metaexpert} or flatten experts~\citep{foil}.
Flattening halves the number of expert layers and doubles both experts per layer and visits.
Expert parameters and compute stay fixed~\citep{foil}.
Foil finds gains of the same order from doubling either visits or experts per layer~\citep{foil}.
Foil states that its study has no non-looped control at equal parameters and compute.
As Foil notes, in its setting this control would activate half of each layer's experts, confounding sharing with sparsity~\citep{foil}.
We train such controls at matched budgets.
Top-8 routing activates 8 of 44 experts in the \Ctrl{}'s middle layers and 8 of 42 in the \DeepCtrl{} (\cref{sec:looptitan-control,sec:depth}).

At matched parameters, MoEUT and Expert Tying find gains over unshared MoEs~\citep{moeut,tying}.
LoopMoE and Dreamer report gains at matched parameters and FLOPs~\citep{loopmoe,dreamer}, and LatentLoop reports them at four matched budgets~\citep{latentloop}.
Concurrently, GLM-5.4 loops part of a pre-trained base and reports gains at equal parameters with about 10\% more training compute~\citep{glm54}.
The GLM-5.4 authors note that looped models trained from scratch still underperform unshared models at equal pre-training compute.

\paragraph{Residual scaling and optimizers}
Depth-µP and related work propose a $1/\sqrt{L}$ residual scale for $L$-layer unshared networks, so that learning rates transfer across depth~\citep{yang2023tp6,bordelon2024depthwise}.
For $N$ visits to shared weights, Loop-µP derives a $1/N$ residual scale, which SMELT adopts~\citep{loopmup,smelt}.
Prior work establishes the symmetry behind \cref{prop:residual}~\citep{yang2021tp4,yang2022tp5,blake2024umup}.
This symmetry holds under Adam when $\varepsilon$ is negligible or set per layer~\citep{everett2024scaling}.
Most looped MoEs train with AdamW~\citep{smelt,foil,loom,metaloop,moeut}, whereas Expert Tying uses Muon~\citep{tying,jordanmuon,moonlight}.

\section{Discussion}
\label{sec:discussion}

Parameter revisit adds depth at fixed stored parameters, yet an MoE can already achieve this depth by trading experts for layers.
At equal stored parameters and compute, both routes yield nearly identical loss.

Therefore, revisit offers an MoE a freedom the MoE already has.
Across the revisit column of \cref{tab:settings} and the \DeepCtrl{} of \cref{fig:compute}, the two routes differ by at most \gapMthreeMfourMuonLargeAbs\mbpb{}.
No difference exceeds a quarter of the recipe's or the visit's gain.
An MoE gains from the shape change, the optimizer and added compute.
We measure each on a standard MoE (\MoEei{} for the shape change, \MoEtw{} for the optimizer and the \DeepCtrl{} under Muon for added compute).

\term{Recommendations}
We recommend three practices for testing a new looped design.
First, build a non-revisit control with the same resources.
The looped model's loss minus the non-revisit control's measures revisit itself (\cref{sec:looptitan-control}).
Second, test under more than one optimizer, because apparent architectural gains can stem from an optimizer setting (\cref{sec:residual}).
Third, evaluate against a well-tuned standard MoE baseline, because switching from AdamW to Muon alone outperforms the entire looped recipe under AdamW (\cref{sec:base}).
\Base{} is such a baseline and matches the looped MoE under Muon.

\term{Limitations}
Our conclusions hold for MoEs with a fixed visit count.
We train from scratch on one dataset and compare models by test loss.
Our comparisons with a \Ctrl{} span two scales (1.3B stored parameters with 5B to 20B tokens, and 3.0B with 12B tokens), from \tokPerActiveLarge{} tokens per active parameter (the larger model) to \tokPerActiveHzTwoZero{} (\MoEtw{} at 20B tokens).
We build the \Ctrl{} only for the state-of-the-art recipe.
In compute scaling, we test one extra visit and one doubling of the experts.
The \DeepCtrl{} trains with Muon only.

Only nine models have more than one seed.
All nine are in the main setting (\cref{sec:looptitan-setup}).
Every other result rests on seed 42.

\term{Open questions}
Whether parameter revisit helps at larger scales remains open.
Under Muon, the revisit step is \gapMthreeMfourMuon\mbpb{} in the main setting and negative in every other comparison.
At one seed each, it is \hgapMthreeMfourMuonHzOneZero{} (10B tokens), \hgapMthreeMfourMuonHzTwoZero{} (20B) and \gapMthreeMfourMuonLarge{} (the larger model).
In the larger model, the looped MoE matches the \Ctrl{} (\cref{tab:settings}).
With added compute, \MoEloop{} keeps a small advantage over the \DeepCtrl{} (\seedVerdictMzeroLoopMzeroDeepMuon, about a tenth of the visit's gain, \cref{sec:depth}).
These two comparisons use more training compute than the main setting.

SMELT reports that its looped MoE's compute saving grows with training compute~\citep{smelt}.
Every run we compare with a \Ctrl{} stays below $10^{20}$ training FLOPs (at most \flopsTotalMthreeLarge{}), the smallest budget at which this saving is reported~\citep{smelt}.
From 5B to 20B tokens, the recipe's gain under Muon grows from \gapMthreeMzeroMuonAbsSFortyTwo{} to \hgapMthreeMzeroMuonHzTwoZeroAbs\mbpb{} (seed 42), whereas the revisit step stays within noise (\cref{fig:results-measurable,tab:settings}).

Trading experts for depth needs enough experts per layer.
Whether revisit matters more in layers with few experts is open.
Our matched-budget runs test one sparsity level (about 85\% of stored parameters inactive per token).
Prior work finds that greater sparsity favors more visits~\citep{metaloop}.
We leave open the value of revisit for converting trained models into looped ones, for sharing the KV cache across visits and for downstream tasks.

\section{Conclusion}
\label{sec:conclusion}

Across matched budgets and compute scaling, parameter revisit itself buys an MoE almost nothing, offering only a depth-for-experts freedom that MoE sparsity already provides. Instead, gains come from network shape, the optimizer and added compute, all of which a standard MoE can adopt.
We propose \Base{} as a stronger MoE baseline that matches the state-of-the-art looped recipe under Muon without revisit or residual scaling, and open-source \moeloop{} alongside 17 pretrained models and our full experiment record to support rigorous evaluation against \Base{} and matched non-revisit controls.

\section*{Acknowledgements}
This study stands on the work that it examines. The authors of the looped recipes that we reimplemented published their designs, their ablations and their own open questions, and several of them named the confounds that we separate here. We thank them, and we regard every difference between their measurements and ours as a question for further work. Our code builds on TorchTitan~\citep{torchtitan}, and every run uses the Qwen3 tokenizer~\citep{qwen3}, the DCLM-baseline data~\citep{dclm} and, for half of the runs, Muon~\citep{jordanmuon,moonlight}.

\makeatletter\let\@fb@secFB\mainfb@saved\makeatother   %
\bibliographystyle{plainnat}
\bibliography{references}

\appendix
\section*{Appendix}

\begin{table}[H]
\caption*{\lead{Where each part of the main text is supported.}}
\begin{tabularx}{\linewidth}{@{}>{\RaggedRight\arraybackslash}X>{\RaggedRight\arraybackslash}p{0.36\linewidth}@{}}
\toprule
In the main text & In the appendix\\
\midrule
Matched budgets, with the configuration and the three resources of every model (\cref{sec:looptitan-recipe,sec:looptitan-control,sec:looptitan-compute}) & \cref{sec:setup-app-arms,sec:setup-app-budgets}\\
Training, evaluation, data and the learning rates (\cref{sec:looptitan-setup}) & \cref{sec:setup-app-train,sec:results-lr}\\
\MoEtw{} matches SMELT's standard MoE in configuration (\cref{sec:looptitan-recipe}) & \cref{sec:setup-app-smelt}\\
Within noise and beyond noise, from the seed runs (\cref{sec:looptitan-setup}) & \cref{sec:results-seeds}\\
\grouprule
Every model at every training length and scale (\cref{tab:settings}) & \cref{sec:results-5b,sec:results-horizon}\\
Most of the gain comes from the shape change (\cref{sec:matched}) & \cref{sec:results-ladder}\\
The trained looped MoE uses its second visit (\cref{sec:matched}) & \cref{sec:obs-readout,sec:obs-parts}\\
Whether shared layers are easier to optimize than independent ones (\cref{sec:matched}) & \cref{sec:obs-curvature}\\
\grouprule
Residual scaling is an optimizer setting, with the proofs and the step size of one update, proved and measured (\cref{sec:residual}) & \cref{sec:proof-app}\\
Residual scaling measured, and the predictions registered before the runs (\cref{sec:residual}) & \cref{sec:results-residual,sec:system-app-trace}\\
\grouprule
One more visit, the \DeepCtrl{} and the same compute spent on tokens (\cref{sec:depth}) & \cref{sec:results-compute,sec:results-tokens}\\
\Base{} and nine published recipes under one protocol (\cref{sec:base}) & \cref{sec:results-base,sec:results-recipes,sec:system-app-models}\\
\grouprule
One model for every recipe, the checker and the capability matrix (\cref{sec:looptitan-config}) & \cref{sec:system-code,sec:system-app-checker,sec:system-app-public}\\
The collaboration of researchers and AI agents (\cref{sec:intro}) & \cref{sec:system-protocol}\\
\bottomrule
\end{tabularx}
\end{table}

\makeatletter
\newcommand{\setupKeepHead}[1]{\par}
\newcommand{\setupKeep}[1]{\par}
\makeatother
\setupKeepHead{13}
\section{Implementation details}
\label{sec:setup-app}

\newcommand{\setupM}[1]{\num[exponent-mode=fixed,fixed-exponent=6,drop-exponent=true,round-mode=places,round-precision=0]{#1}}
\newcommand{\setupKvULarge}{\budKvU}
\newcommand{\setupKvMfourLarge}{\budKvMfour}
\newcommand{\setupKvMthreeLarge}{\budKvMthree}

One model class implements every configuration (\cref{sec:looptitan-config}).

\subsection{Configurations and resources}
\label{sec:setup-app-arms}

Matched budgets keep each resource within 5\% of the same-scale standard MoE (\cref{sec:looptitan-recipe}).
Every model of \cref{tab:setup-app-arch} compared at matched budgets stays within 3.125\%, with the largest difference in KV cache (6,336 against 6,144 bytes per token in the main setting, \cref{sec:setup-app-budgets}).
Configurations of published looped recipes stay within 5\% (\cref{tab:results-recipes-config}).
The three compute-scaling and expert-scaling models leave one or two resources unmatched by design (\cref{sec:looptitan-compute}).

\begin{table}[tp]
\caption{\lead{Every model compared at matched budgets stays within 3.125\% of its standard MoE per resource.} Steps 1 and 2 are the shape change, 3 the recipe's expert layout, 4 residual scaling and 5 revisit (\cref{sec:looptitan-control}). Layers are prefix\,+\,core\,$\times$\,visits\,+\,suffix, heads are query heads\,$\times$\,head dimension and experts are per layer (prefix\,/\,core\,/\,suffix). Resources are relative to each scale's reference row, and bold marks resources unmatched by design.}
\label{tab:setup-app-arch}
\label{tab:setup-arms}
\label{tab:setup-app-budgets}
\begin{fittabular}{12}{NllNllNlNNNN}
\toprule
 & & & & & & Expert & & Active & \multicolumn{3}{c}{Resources, relative \colunit{\%}}\\
\cmidrule(l){10-12}
\multicolumn{2}{l}{Model} & Layers & Width & Heads & Experts & width & $a$ & \colunit{M} & Stored & FLOPs & KV\\
\midrule
\multicolumn{12}{l}{\textit{The five steps, 1.3B stored (0.2B active)}}\\
 & \MoEtw & 12 & 1280 & 10\,$\times$\,128 & 64 & 432 & 1 & \setupM{203521280} & \multicolumn{3}{c}{\color{hlMid}reference}\\
1 & \MoEtw{} with finer experts & 12 & 1280 & 10\,$\times$\,128 & 72 (top-9) & 384 & 1 & \setupM{203644160} & \budParamsG & \budFlopsG & \budKvG\\
2 & \MoEei{} (\Base{} with Muon) & 18 & 1056 & 12\,$\times$\,88 & 58 & 384 & 1 & \setupM{219809568} & \budParamsU & \budFlopsU & \budKvU\\
3 & \begin{tabular}[t]{@{}l@{}}\MoEei{} with the recipe's\\\hspace{0.8em}expert layout\end{tabular} & 3\,+\,12\,+\,3 & 1056 & 12\,$\times$\,88 & 88\,/\,44\,/\,88 & 384 & 1 & \setupM{219822240} & \budParamsR & \budFlopsR & \budKvR\\
4 & Non-revisit control & 3\,+\,12\,+\,3 & 1056 & 12\,$\times$\,88 & 88\,/\,44\,/\,88 & 384 & 1/2 & \setupM{219822240} & \budParamsMfour & \budFlopsMfour & \budKvMfour\\
\oursrow 5 & \ours{Looped MoE} & 3\,+\,6\,$\times$\,2\,+\,3 & 1056 & 12\,$\times$\,88 & 88 & 384 & 1/2 & \setupM{220379808} & \budParamsMthree & \budFlopsMthree & \budKvMthree\\
\grouprule
\multicolumn{12}{l}{\textit{Variants of the looped MoE}}\\
 & Experts revisited & 3\,+\,6\,$\times$\,2\,+\,3 & 1056 & 12\,$\times$\,88 & 88 & 384 & 1/2 & \setupM{220379808} & \budParamsMseven & \budFlopsMseven & \budKvMseven\\
 & Experts and router revisited & 3\,+\,6\,$\times$\,2\,+\,3 & 1056 & 12\,$\times$\,88 & 88 & 384 & 1/2 & \setupM{220379808} & \budParamsMfive & \budFlopsMfive & \budKvMfive\\
 & 3 visits & 3\,+\,4\,$\times$\,3\,+\,3 & 1056 & 12\,$\times$\,88 & 88\,/\,134\,/\,88 & 384 & 1/3 & \setupM{220962720} & \budParamsSmeltVthree & \budFlopsSmeltVthree & \budKvSmeltVthree\\
 & 4 visits & 3\,+\,3\,$\times$\,4\,+\,3 & 1056 & 12\,$\times$\,88 & 88\,/\,179\,/\,88 & 384 & 1/4 & \setupM{221532960} & \budParamsSmeltVfour & \budFlopsSmeltVfour & \budKvSmeltVfour\\
\grouprule
\multicolumn{12}{l}{\textit{Residual scaling on \MoEtw{}, and standard MoEs of 24 and 36 layers}}\\
 & \MoEtw{} with residual scaling & 3\,+\,6\,+\,3 & 1280 & 10\,$\times$\,128 & 64 & 432 & 1/2 & \setupM{203521280} & \budParamsMzeroR & \budFlopsMzeroR & \budKvMzeroR\\
 & MoE-24 & 24 & 896 & 14\,$\times$\,64 & 62 & 320 & 1 & \setupM{207815552} & \budParamsDtwoFour & \budFlopsDtwoFour & \budKvDtwoFour\\
 & MoE-36 & 36 & 704 & 16\,$\times$\,44 & 66 & 256 & 1 & \setupM{195352256} & \budParamsDthreeSix & \budFlopsDthreeSix & \budKvDthreeSix\\
\grouprule
\multicolumn{12}{l}{\textit{Compute scaling and expert scaling}}\\
 & \MoEloop & 3\,+\,6\,$\times$\,2\,+\,3 & 1280 & 10\,$\times$\,128 & 64 & 432 & 1/2 & \setupM{305281280} & \budParamsMzeroLoop & \textbf{\budFlopsMzeroLoop} & \textbf{\budKvMzeroLoop}\\
 & Deeper control & 18 & 1280 & 10\,$\times$\,128 & 42 & 432 & 1 & \setupM{304774400} & \budParamsMzeroDeep & \textbf{\budFlopsMzeroDeep} & \textbf{\budKvMzeroDeep}\\
 & \MoEtwoE & 12 & 1280 & 10\,$\times$\,128 & 128 & 432 & 1 & \setupM{204504320} & \textbf{\budParamsMzeroTwoE} & \budFlopsMzeroTwoE & \budKvMzeroTwoE\\
\midrule
\multicolumn{12}{l}{\textit{The 3.0B setting (0.5B active)}}\\
 & Standard MoE, 16 layers & 16 & 2048 & 16\,$\times$\,128 & 64 & 448 & 1 & \setupM{497092608} & \multicolumn{3}{c}{\color{hlMid}reference}\\
 & Standard MoE, 24 layers & 24 & 1672 & 19\,$\times$\,88 & 60 & 392 & 1 & \setupM{521264392} & \budParamsULarge & \budFlopsULarge & \setupKvULarge\\
 & Non-revisit control & 4\,+\,16\,+\,4 & 1672 & 19\,$\times$\,88 & 92\,/\,46\,/\,92 & 392 & 1/2 & \setupM{521317896} & \budParamsMfourLarge & \budFlopsMfourLarge & \setupKvMfourLarge\\
\oursrow & \ours{Looped MoE} & 4\,+\,8\,$\times$\,2\,+\,4 & 1672 & 19\,$\times$\,88 & 92 & 392 & 1/2 & \setupM{522548488} & \budParamsMthreeLarge & \budFlopsMthreeLarge & \setupKvMthreeLarge\\
\bottomrule
\end{fittabular}
\tablenote{Residual scale $a$ applies to the core. Routing is top-8 unless marked, and top-9 in step 1 preserves evaluated expert width per token (\mbox{$9\times384=8\times432$}). Active counts parameters applied per token. \MoEtw{} has 1,318,288,640 stored parameters, 3,140,728,320 training FLOPs per token and 6,144 bytes of KV cache per token, and the 16-layer standard MoE has 2,963,343,360, 6,456,471,552 and 8,192. In the two partial-revisit variants, unshared core parameters have separate copies per visit (\cref{sec:obs-parts}). Residual scaling's two parts use \MoEtw{} ($a=1$) and modify output projections in its middle six layers. The learning-rate part halves their learning rate and doubles their weight-decay coefficient, and the initialization part halves their initial standard deviation with the same random draws (\cref{sec:results-residual}). In the 3.0B setting, \Base{} is the 24-layer standard MoE under Muon (\cref{sec:base}). Run-record names are in \cref{tab:system-app-models}.}
\end{table}

\setupKeep{8}
\term{Architecture}
Every model uses \cref{eq:setup-block} with RMSNorm before each branch, rotary position embeddings and bias-free attention and expert projections.
A linear router scores each layer's SwiGLU experts and mixes the top-$k$ outputs with renormalized softmax weights, without shared experts or dropped tokens.
Attention is multi-query in every model, so each layer's query heads share one key--value head.
This head's dimension shrinks with depth to match the KV cache, except in \MoEloop{} and the \DeepCtrl{}, which keep 128 (\cref{tab:setup-app-arch}).
All models omit QK-norm and use one default initialization standard deviation for all matrices (\cref{tab:setup-app-train} names the one exception).

\setupKeep{5}
\term{Active parameters}
Active parameters are the non-embedding parameters applied per token (attention, router, norms and top-$k$ experts in every computed layer, plus the final norm).
Revisited parameters count once per visit, so a model with revisit applies more parameters per token than it holds distinct ones.
Per token, the looped MoE applies \setupM{220379808}M parameters (\setupM{146920224}M distinct) and \MoEloop{} applies \setupM{305281280}M (\setupM{203521280}M distinct).

\setupKeep{8}
\term{Deeper models and the 3.0B setting}
Relative to \MoEloop{}, the \DeepCtrl{} differs by \budParamsMzeroDeepVsLoop\% in stored parameters and \budFlopsMzeroDeepVsLoop\% in FLOPs per token, with equal KV cache.
MoE-24 and MoE-36 trade width for depth at \MoEtw{}'s resources.
Smaller head dimensions keep their KV cache within budget, and their ratio of expert to model width stays near \MoEei{}'s.
In the 3.0B setting (3.0B stored parameters), the 16-layer standard MoE stores 2.2 times \MoEtw{}'s parameters, and the 24-layer standard MoE corresponds to \MoEei.
Relative to the 16-layer standard MoE, this setting's looped MoE keeps the main-setting proportions (width ratio 0.82, head dimension 88 and 1.4 times the experts at 0.9 times expert width).

\subsection{How the three resources are counted}
\label{sec:setup-app-budgets}

Stored parameters exclude the input embedding and output head.
Training FLOPs per token equal three times the forward matrix FLOPs per position (one forward pass plus a backward pass charged at twice the forward), giving
\begin{equation}
F_{\mathrm{train}} \;=\; 3\,\Bigl[\;\sum_{\ell\,\in\,\text{computed layers}} \bigl(2P^{\mathrm{attn}}_{\ell} + 4wL + 2wm_{\ell} + 6wf_{\ell}k_{\ell}\bigr) \;+\; 2wn_{\mathrm{voc}}\Bigr].
\label{eq:setup-app-flops}
\end{equation}
Here $w$ is model width, $P^{\mathrm{attn}}_{\ell}$ the parameter count of the four attention projections, $L=4096$ the keys charged per position (full sequence length) and $n_{\mathrm{voc}}$ the vocabulary size.
Layer $\ell$'s router scores $m_{\ell}$ experts and evaluates $k_{\ell}$ of hidden width $f_{\ell}$.
The sum runs over computed layers, so revisited layers count once per visit.
The output head is charged once.
We exclude normalization, position embeddings, softmax, expert selection and the optimizer.
Forward terms sum to 1,046,909,440 for \MoEtw.

Each computed layer caches one key and one value vector of its head dimension per context token (2 bytes per element).
Each visit uses its own cache slot, because revisiting parameters does not merge cache entries.
\MoEtw{} caches $12\times2\times128\times2=6{,}144$ bytes per token and the looped MoE $18\times2\times88\times2=6{,}336$ ($3.125\%$ more).

\subsection{Training and evaluation}
\label{sec:setup-app-train}

All models of one scale share one training and evaluation protocol (\cref{tab:setup-app-train}).
The seed alone fixes the initial weights, drawn on CPU in a fixed parameter order, and the data order.
A model's first training loss matches in every printed digit across its learning-rate grid.
We calibrate one peak learning rate per optimizer and scale.
Two cases depart from the protocol by design.
The two parts of residual scaling change the learning rate and weight decay, or the initialization, of six layers (\cref{tab:setup-app-arch}).
One configuration of a published recipe trains at \lr{3e-4} under AdamW (\cref{sec:results-recipes}).

\begin{table}[tp]
\caption{\lead{All models of one scale share one training and evaluation protocol.} In the 3.0B setting, only the listed rows change. Token counts are target tokens scored by the loss. Exceptions are residual scaling's two parts (\cref{tab:setup-app-arch}) and one published recipe under AdamW (\cref{sec:results-recipes}).}
\label{tab:setup-app-train}
\begin{tabularx}{\linewidth}{@{}l>{\RaggedRight\arraybackslash}X@{}}
\toprule
\multicolumn{2}{@{}l}{\textit{Data}}\\
Source & DCLM-baseline 1.0~\citep{dclm}, 19,624,997 documents scanned, 10,762 duplicates removed by document identifier and by the hash of the exact content\\
Splits & the first 8 bytes of the content hash modulo 100 assign a document to test (0), validation (1) or training (2 to 99), so no document text is shared between splits\\
Training split & 24,000,067,249 tokens, of which a 5B run uses 5.0B\\
Validation split & 10,001,497 tokens in 44,351,437 bytes, for learning-rate selection\\
Test split & 10,009,889 tokens in 44,371,521 bytes, for every reported BPB\\
Tokenizer & Qwen3, vocabulary 151,669, with one token as beginning, end and padding\\
Sequences & 4096 tokens of packed documents, each wrapped in begin and end tokens, with attention masked to the token's document and positions restarting in each document\\
\grouprule
\multicolumn{2}{@{}l}{\textit{Training in the main setting, 1.3B stored (0.2B active)}}\\
Batch & 256 sequences, 1,048,576 tokens per update\\
Training length & 4,769 updates (5.0B tokens), continued to 9,540 updates (10.0B) and 19,080 updates (20.0B)\\
Schedule & linear warm-up over 95 updates, constant, linear decay to 0 over the last 20\%\\
Peak learning rate & \selLrAdamW{} under AdamW, \selLrMuon{} under Muon\\
AdamW & betas 0.9 and 0.95, epsilon $10^{-8}$, weight decay 0.1 with none on norms, embedding, head and routers, gradient clipping at norm 1.0\\
Muon & momentum 0.95 with Nesterov, 5 Newton--Schulz iterations, weight decay 0.1, learning rate multiplied by $0.2\sqrt{\max(p,q)}$ for a $p\times q$ matrix to match the root-mean-square update of AdamW\\
Muon acts on & attention projections and expert matrices, each expert orthogonalized separately, with AdamW as above on embedding, head, routers and norms\\
Attention & one key--value head, no QK-norm, head dimension as in \cref{tab:setup-app-arch}\\
Initialization & normal with standard deviation 0.02 for every matrix, output projections, routers, embedding and head included, with no depth-dependent rescaling and norm scales 1\\
Loss & cross-entropy on the final output plus 0.01 times the router load-balance loss, no router z-loss\\
Precision & bfloat16 computation, float32 master weights, float32 router softmax\\
Seed & 42 for every model, and also 43 and 44 for the main-setting models of \cref{tab:results-seeds}\\
Hardware & 8 H200 GPUs per run, gradient accumulation 2 (GPU hours in \cref{tab:results-5b})\\
\grouprule
\multicolumn{2}{@{}l}{\textit{Training in the 3.0B setting (0.5B active)}}\\
Training length & 11,440 updates (12.0B tokens), in runs that branch from runs of 22,888 updates (24.0B tokens, one pass), with shorter branches ending at 1,430, 2,860 and 5,720 updates\\
Schedule & linear warm-up over 458 updates, constant, linear decay to 0 over the last 20\% of each run\\
Peak learning rate & \selLrSzZeroPFiveAdamW{} under both optimizers, from a sweep of shorter runs (\cref{sec:setup-app-train})\\
Hardware & 16 H200 GPUs per run, no accumulation\\
\midrule
\multicolumn{2}{@{}l}{\textit{Evaluation}}\\
Reported metric & BPB on the complete test split from the final checkpoint, at the trained number of visits, in batches of 4 sequences of 4096 tokens with document-masked attention\\
BPB & summed cross-entropy over $\ln 2$ times the number of bytes of the target tokens, with special tokens excluded from both\\
Learning-rate selection & lowest validation BPB on the complete validation split, with ties to the smaller learning rate\\
\bottomrule
\end{tabularx}
\end{table}

\setupKeep{15}
\term{Learning rates}
In the main setting, full 5B-token grid runs on \MoEtw{} and the looped MoE select the peak learning rate minimizing validation BPB, with ties to the smaller rate (\cref{tab:results-lrgrid}).
The 3.0B setting takes its rate from a sweep at one eighth of the length of its 12B-token runs.
Under AdamW it keeps the sweep's best rate, because it treats AdamW's optimum as independent of training length.
Under Muon it multiplies the fitted optimum (the vertex of a parabola in log learning rate fitted to validation BPB) by $8^{-1/8}\approx\lrRuleCoef$.
We adopted this rule from main-setting fitted optima at 0.625B to 5B tokens (\cref{sec:results-lr}).

The sweep trains the 16-layer standard MoE on 1.5B tokens, one eighth of the 12B-token runs (\cref{tab:results-lrsweep}).
Under AdamW, validation BPB is lowest at \selLrSzZeroPFiveAdamW{}, which the rule keeps.
Under Muon, validation BPB is lowest at \lr{2.4e-3}, with fitted optimum \szVtxSzZeroPFiveMuonDnThree{}.
Multiplying by \lrRuleCoef{} gives \lr{1.72e-3}, rounded to the nearest grid point, \selLrSzZeroPFiveMuon{}.
Every model in the 3.0B setting therefore trains at \selLrSzZeroPFiveAdamW{} under both optimizers.

\setupKeep{8}
\term{Training lengths}
Longer or shorter training lengths use a fork (a run branching from a checkpoint with only the schedule's end changed).
The 20B run branches from the 5B run's constant phase at update 3,339, holds the peak rate to update 15,264 and decays over 3,816 updates.
The 10B run branches from update 7,632 of the 20B run, decaying over 1,908 updates.
Every other fork (main-setting runs below 5B tokens and 3.0B-setting runs up to 12B) branches from the constant phase and decays over its last 20\% of updates.
A fork is accepted only if it changes solely the learning-rate schedule and matches its parent's schedule through the branch point.

\subsection{Relation to SMELT's recipe}
\label{sec:setup-app-smelt}

The looped MoE keeps SMELT's loop, sharing and residual scale~\citep{smelt} (\cref{tab:setup-app-smelt}).
Its 88 experts of width 384 per layer match the three budgets, against 96 experts in the published model.
We compare at SMELT's ablation scale (0.2B active parameters).

\begin{table}[!ht]
\caption{\lead{\MoEtw{} matches SMELT's standard MoE in configuration, and the looped MoE keeps SMELT's loop with 88 experts per layer instead of 96.} Published values are from \citet{smelt} at 0.2B active parameters.}
\label{tab:setup-app-smelt}
\begin{tabularx}{\linewidth}{@{}l>{\RaggedRight\arraybackslash}X>{\RaggedRight\arraybackslash}X@{}}
\toprule
 & As published~\citep{smelt} & This paper\\
\midrule
\multicolumn{3}{@{}l}{\textit{Standard MoE, in this paper \MoEtw}}\\
Layers\,$\times$\,width & 12\,$\times$\,1280 & 12\,$\times$\,1280\\
Experts per layer & 64, top-8 & 64, top-8\\
Stored parameters & 1.32B & 1.318B\\
Active parameters & 0.207B (sparsity about 85\%) & 0.204B\\
Expert width & not reported & 432\\
Attention & not reported & 10 query heads of dimension 128\\
\grouprule
\multicolumn{3}{@{}l}{\textit{Looped MoE}}\\
Layers computed twice & middle half of the stored layers & middle 6 of 12 (18 computed layers)\\
Shared between visits & all core parameters & attention, router, norms and experts\\
Residual scale & 1/2 on both core branches, 1 outside & the same\\
Visits & fixed number, gradients through both & the same\\
Loss & cross-entropy on the final output & the same\\
Experts per layer & 96 & 88 of width 384, at which the three budgets match\\
\bottomrule
\end{tabularx}
\end{table}

\section{Further results}
\label{sec:results-app}

Unless stated otherwise, values use seed 42 in the main setting (1.3B stored parameters, 0.2B active, 5B tokens, 4,769 updates) with AdamW at \selLrAdamW{} or Muon at \selLrMuon.
The main text reports three-seed means for nine models (\cref{sec:results-seeds}), and \cref{tab:results-5b,tab:results-seeds} give seed-42 values and three-seed means.
Values are last-update bits per byte on the complete held-out validation and test splits.
Differences $A-B$ are in test mBPB unless stated otherwise.
They come from unrounded values, so they can differ by 0.1 from differences of rounded entries.
Two standard errors, below which a difference is within noise, are \seedTwoSeOneOneAdamW\mbpb{} under AdamW and \seedTwoSeOneOneMuon{} under Muon with one seed each, and \seedTwoSeThreeThreeAdamW{} and \seedTwoSeThreeThreeMuon{} with three seeds each (\cref{sec:results-seeds}).

\subsection{Seed repeats and run-to-run variation}
\label{sec:results-seeds}

We judge each difference by its optimizer's run-to-run standard deviation $s$, \seedSdAdamW\mbpb{} under AdamW and \seedSdMuon{} under Muon (\cref{tab:results-seeds}).
We first repeated four main-setting models with seeds 43 and 44, which gives six model and optimizer pairs of three runs each.
These six groups are the \Ctrl{} and the looped MoE under both optimizers, plus \MoEloop{} and the \DeepCtrl{} under Muon.
We then repeated \MoEtw{} and \MoEei{} under both optimizers, \MoEtwoE{} under Muon, and \MoEei{} with the recipe's expert layout and \MoEtw{} with residual scaling under AdamW, nine models in all.
The seed sets initialization and document order, with every other setting as in the seed-42 run.
Each optimizer's $s$ pools all of its three-run groups, six under AdamW (\seedDfAdamW{} degrees of freedom) and seven under Muon (\seedDfMuon).

\begin{table}[!tp]
\caption{\lead{The run-to-run standard deviation is \seedSdAdamW\mbpb{} under AdamW and \seedSdMuon{} under Muon.} Test BPB in the main setting. SD is each row's three-run standard deviation, pooled by optimizer in the tinted rows.}
\label{tab:results-seeds}
{%
\begin{fittabular}{7}{llNNNNN}
\toprule
Model & Optimizer & Seed 42 & Seed 43 & Seed 44 & Mean & SD \colunit{mBPB}\\
\midrule
Looped MoE & AdamW & \testMthreeAdamW & \seedTestMthreeAdamWSFortyThree & \seedTestMthreeAdamWSFortyFour & \seedTestMeanMthreeAdamW & \seedSdGroupMthreeAdamW\\
Non-revisit control & AdamW & \testMfourAdamW & \seedTestMfourAdamWSFortyThree & \seedTestMfourAdamWSFortyFour & \seedTestMeanMfourAdamW & \seedSdGroupMfourAdamW\\
\MoEtw & AdamW & \testMzeroAdamW & \seedTestMzeroAdamWSFortyThree & \seedTestMzeroAdamWSFortyFour & \seedTestMeanMzeroAdamW & \seedSdGroupMzeroAdamW\\
\MoEei & AdamW & \testUAdamW & \seedTestUAdamWSFortyThree & \seedTestUAdamWSFortyFour & \seedTestMeanUAdamW & \seedSdGroupUAdamW\\
\MoEei{} with the recipe's expert layout & AdamW & \testRAdamW & \seedTestRAdamWSFortyThree & \seedTestRAdamWSFortyFour & \seedTestMeanRAdamW & \seedSdGroupRAdamW\\
\MoEtw{} with residual scaling & AdamW & \testMzeroRAdamW & \seedTestMzeroRAdamWSFortyThree & \seedTestMzeroRAdamWSFortyFour & \seedTestMeanMzeroRAdamW & \seedSdGroupMzeroRAdamW\\
\oursrow \multicolumn{2}{l}{$s$ under AdamW, six groups} & & & & & \seedSdAdamW\\
\grouprule
Looped MoE & Muon & \testMthreeMuon & \seedTestMthreeMuonSFortyThree & \seedTestMthreeMuonSFortyFour & \seedTestMeanMthreeMuon & \seedSdGroupMthreeMuon\\
Non-revisit control & Muon & \testMfourMuon & \seedTestMfourMuonSFortyThree & \seedTestMfourMuonSFortyFour & \seedTestMeanMfourMuon & \seedSdGroupMfourMuon\\
\MoEloop & Muon & \testMzeroLoopMuon & \seedTestMzeroLoopMuonSFortyThree & \seedTestMzeroLoopMuonSFortyFour & \seedTestMeanMzeroLoopMuon & \seedSdGroupMzeroLoopMuon\\
Deeper control & Muon & \testMzeroDeepMuon & \seedTestMzeroDeepMuonSFortyThree & \seedTestMzeroDeepMuonSFortyFour & \seedTestMeanMzeroDeepMuon & \seedSdGroupMzeroDeepMuon\\
\MoEtw & Muon & \testMzeroMuon & \seedTestMzeroMuonSFortyThree & \seedTestMzeroMuonSFortyFour & \seedTestMeanMzeroMuon & \seedSdGroupMzeroMuon\\
\MoEei & Muon & \testUMuon & \seedTestUMuonSFortyThree & \seedTestUMuonSFortyFour & \seedTestMeanUMuon & \seedSdGroupUMuon\\
\MoEtwoE & Muon & \testMzeroTwoEMuon & \seedTestMzeroTwoEMuonSFortyThree & \seedTestMzeroTwoEMuonSFortyFour & \seedTestMeanMzeroTwoEMuon & \seedSdGroupMzeroTwoEMuon\\
\oursrow \multicolumn{2}{l}{$s$ under Muon, seven groups} & & & & & \seedSdMuon\\
\grouprule
\multicolumn{2}{l}{Registered $s$, the first six groups} & & & & & \seedSdPooled\\
\bottomrule
\end{fittabular}\par}
\vspace{16pt}
\caption{\lead{Over three seeds, the looped MoE matches the \Ctrl{} under both optimizers, and \MoEloop{} is \seedDiffMzeroLoopMzeroDeepMuonAbs\mbpb{} below the \DeepCtrl{} (\seedVerdictMzeroLoopMzeroDeepMuon{}, lower in every seed).} Differences $A-B$ of seed means in test mBPB, with one standard error and per-seed values in tinted rows. Seeds gives $n_A$ vs $n_B$. Verdict is within noise when Mean (not a single seed) is below two standard errors with its optimizer's $s$, and beyond noise otherwise. Registered uses the pooled $s$ and is bold where it differs.}
\label{tab:seeds}
{%
\newcommand{\reg}[1]{{\color{hlMid}#1}}%
\newcommand{\regc}[2]{\ifx#1#2\reg{#2}\else\best{#2}\fi}%
\begin{fittabular}{9}{lNNNNNNll}
\toprule
 & \multicolumn{3}{c}{Seed} & & & & &\\
\cmidrule(lr){2-4}
Difference $A-B$ \colunit{mBPB} & 42 & 43 & 44 & Mean & SE & Seeds & Verdict & \reg{Registered}\\
\midrule
\multicolumn{9}{l}{\textit{Under AdamW}}\\
\oursrow Looped MoE $-$ \Ctrl & \seedPairMthreeMfourAdamWSFortyTwo & \seedPairMthreeMfourAdamWSFortyThree & \seedPairMthreeMfourAdamWSFortyFour & \seedDiffMthreeMfourAdamW & \seedSeGapMthreeMfourAdamW & \seedNsMthreeMfourAdamW & \seedVerdictMthreeMfourAdamW & \regc{\seedVerdictMthreeMfourAdamW}{\seedVerdictPooledMthreeMfourAdamW}\\
Looped MoE $-$ \MoEei & \seedPairMthreeUAdamWSFortyTwo & & & \seedDiffMthreeUAdamW & \seedSeGapMthreeUAdamW & \seedNsMthreeUAdamW & \seedVerdictMthreeUAdamW & \regc{\seedVerdictMthreeUAdamW}{\seedVerdictPooledMthreeUAdamW}\\
Non-revisit control $-$ \MoEei & \seedPairMfourUAdamWSFortyTwo & & & \seedDiffMfourUAdamW & \seedSeGapMfourUAdamW & \seedNsMfourUAdamW & \seedVerdictMfourUAdamW & \regc{\seedVerdictMfourUAdamW}{\seedVerdictPooledMfourUAdamW}\\
\oursrow Non-revisit control $-$ \MoEei{} with the recipe's expert layout & \seedPairMfourRAdamWSFortyTwo & \seedPairMfourRAdamWSFortyThree & \seedPairMfourRAdamWSFortyFour & \seedDiffMfourRAdamW & \seedSeGapMfourRAdamW & \seedNsMfourRAdamW & \seedVerdictMfourRAdamW & \regc{\seedVerdictMfourRAdamW}{\seedVerdictPooledMfourRAdamW}\\
\MoEtw{} with residual scaling $-$ \MoEtw & \gapMzeroRMzeroAdamWSFortyTwo & & & \gapMzeroRMzeroAdamW & \seGapMzeroRMzeroAdamW & \nSeedsGapMzeroRMzeroAdamW & \verdictGapMzeroRMzeroAdamW & \reg{within noise}\\
\MoEei{} $-$ \MoEtw & \seedPairUMzeroAdamWSFortyTwo & & & \seedDiffUMzeroAdamW & \seedSeGapUMzeroAdamW & \seedNsUMzeroAdamW & \seedVerdictUMzeroAdamW & \regc{\seedVerdictUMzeroAdamW}{\seedVerdictPooledUMzeroAdamW}\\
Looped MoE $-$ \MoEtw & \seedPairMthreeMzeroAdamWSFortyTwo & & & \seedDiffMthreeMzeroAdamW & \seedSeGapMthreeMzeroAdamW & \seedNsMthreeMzeroAdamW & \seedVerdictMthreeMzeroAdamW & \regc{\seedVerdictMthreeMzeroAdamW}{\seedVerdictPooledMthreeMzeroAdamW}\\
\grouprule
\multicolumn{9}{l}{\textit{Under Muon}}\\
\oursrow Looped MoE $-$ \Ctrl & \seedPairMthreeMfourMuonSFortyTwo & \seedPairMthreeMfourMuonSFortyThree & \seedPairMthreeMfourMuonSFortyFour & \seedDiffMthreeMfourMuon & \seedSeGapMthreeMfourMuon & \seedNsMthreeMfourMuon & \seedVerdictMthreeMfourMuon & \regc{\seedVerdictMthreeMfourMuon}{\seedVerdictPooledMthreeMfourMuon}\\
Looped MoE $-$ \MoEei & \seedPairMthreeUMuonSFortyTwo & & & \seedDiffMthreeUMuon & \seedSeGapMthreeUMuon & \seedNsMthreeUMuon & \seedVerdictMthreeUMuon & \regc{\seedVerdictMthreeUMuon}{\seedVerdictPooledMthreeUMuon}\\
Non-revisit control $-$ \MoEei & \seedPairMfourUMuonSFortyTwo & & & \seedDiffMfourUMuon & \seedSeGapMfourUMuon & \seedNsMfourUMuon & \seedVerdictMfourUMuon & \regc{\seedVerdictMfourUMuon}{\seedVerdictPooledMfourUMuon}\\
Non-revisit control $-$ \MoEei{} with the recipe's expert layout & \seedPairMfourRMuonSFortyTwo & & & \seedDiffMfourRMuon & \seedSeGapMfourRMuon & \seedNsMfourRMuon & \seedVerdictMfourRMuon & \regc{\seedVerdictMfourRMuon}{\seedVerdictPooledMfourRMuon}\\
\MoEei{} $-$ \MoEtw & \seedPairUMzeroMuonSFortyTwo & & & \seedDiffUMzeroMuon & \seedSeGapUMzeroMuon & \seedNsUMzeroMuon & \seedVerdictUMzeroMuon & \regc{\seedVerdictUMzeroMuon}{\seedVerdictPooledUMzeroMuon}\\
Looped MoE $-$ \MoEtw & \seedPairMthreeMzeroMuonSFortyTwo & & & \seedDiffMthreeMzeroMuon & \seedSeGapMthreeMzeroMuon & \seedNsMthreeMzeroMuon & \seedVerdictMthreeMzeroMuon & \regc{\seedVerdictMthreeMzeroMuon}{\seedVerdictPooledMthreeMzeroMuon}\\
\addlinespace[3pt]
\oursrow \MoEloop{} $-$ \DeepCtrl & \seedPairMzeroLoopMzeroDeepMuonSFortyTwo & \seedPairMzeroLoopMzeroDeepMuonSFortyThree & \seedPairMzeroLoopMzeroDeepMuonSFortyFour & \seedDiffMzeroLoopMzeroDeepMuon & \seedSeGapMzeroLoopMzeroDeepMuon & \seedNsMzeroLoopMzeroDeepMuon & \seedVerdictMzeroLoopMzeroDeepMuon & \regc{\seedVerdictMzeroLoopMzeroDeepMuon}{\seedVerdictPooledMzeroLoopMzeroDeepMuon}\\
\MoEloop{} $-$ \MoEtwoE & \seedPairMzeroLoopMzeroTwoEMuonSFortyTwo & & & \seedDiffMzeroLoopMzeroTwoEMuon & \seedSeGapMzeroLoopMzeroTwoEMuon & \seedNsMzeroLoopMzeroTwoEMuon & \seedVerdictMzeroLoopMzeroTwoEMuon & \regc{\seedVerdictMzeroLoopMzeroTwoEMuon}{\seedVerdictPooledMzeroLoopMzeroTwoEMuon}\\
\MoEloop{} $-$ \MoEtw & \seedPairMzeroLoopMzeroMuonSFortyTwo & & & \seedDiffMzeroLoopMzeroMuon & \seedSeGapMzeroLoopMzeroMuon & \seedNsMzeroLoopMzeroMuon & \seedVerdictMzeroLoopMzeroMuon & \regc{\seedVerdictMzeroLoopMzeroMuon}{\seedVerdictPooledMzeroLoopMzeroMuon}\\
Deeper control $-$ \MoEtw & \seedPairMzeroDeepMzeroMuonSFortyTwo & & & \seedDiffMzeroDeepMzeroMuon & \seedSeGapMzeroDeepMzeroMuon & \seedNsMzeroDeepMzeroMuon & \seedVerdictMzeroDeepMzeroMuon & \regc{\seedVerdictMzeroDeepMzeroMuon}{\seedVerdictPooledMzeroDeepMzeroMuon}\\
\grouprule
\multicolumn{9}{l}{\textit{Difference of two differences, with two standard errors of \seedTwoSeResidualStepMuonAdamWTwoDec\mbpb}}\\
Residual-scaling step under Muon $-$ under AdamW & \shrinkMfourR & & & \seedGapResidualStepMuonAdamW & & -- &\seedVerdictResidualStepMuonAdamW & \regc{\seedVerdictResidualStepMuonAdamW}{\seedVerdictPooledResidualStepMuonAdamW}\\
\bottomrule
\end{fittabular}}
\tablenote{Tinted rows list each seed's difference, and other rows list seed 42 and the mean. For 1 vs 1, 3 vs 1 and 3 vs 3 seeds, two standard errors are \seedTwoSeOneOneAdamW, \seedTwoSeThreeOneAdamW{} and \seedTwoSeThreeThreeAdamW\mbpb{} under AdamW, \seedTwoSeOneOneMuon, \seedTwoSeThreeOneMuon{} and \seedTwoSeThreeThreeMuon{} under Muon, and \seedTwoSeOneOne, \seedTwoSeThreeOne{} and \seedTwoSeThreeThree{} with pooled $s$. Under Muon, \MoEei{} with the recipe's expert layout has seed 42 only. The residual-scaling step (\cref{sec:residual}) is the \Ctrl{} minus \MoEei{} with the recipe's expert layout. The last row combines both optimizers' $s$ (two standard errors \seedTwoSePooledResidualStepMuonAdamW\mbpb{} with pooled $s$), with three seeds against one for the Muon step and three seeds each for the AdamW step.}
\end{table}

\term{Within noise and beyond noise}
Each optimizer's $s$ pools its three-run groups via $s^2 = \sum_{g}\sum_{i}(x_{gi}-\bar{x}_g)^2/(2G)$, for test BPB $x_{gi}$ of seed $i$ in group $g$ across $G$ groups (six under AdamW, seven under Muon).
For models with $n_A$ and $n_B$ seeds (seed 42 alone has $n=1$), a difference has standard error $\sqrt{s_A^2/n_A+s_B^2/n_B}$ with the $s$ of each model's optimizer.
A difference below two standard errors is within noise, and we say the two models match.
Otherwise, the difference is beyond noise.
Two standard errors are \seedTwoSeOneOneAdamW\mbpb{} under AdamW and \seedTwoSeOneOneMuon{} under Muon (one seed each), or \seedTwoSeThreeThreeAdamW{} and \seedTwoSeThreeThreeMuon{} (three seeds each).

We registered one pooled $s$ (\seedSdPooled\mbpb{}).
The two optimizers' spreads differ (\seedSdAdamW{} against \seedSdMuon\mbpb), so every verdict in this paper uses the $s$ of the difference's optimizer, and \cref{tab:seeds} gives the registered verdict beside it.
The registered value was fixed before the results and pools the six groups of the first repeats.
It lies between the two optimizers' $s$, so for a difference within one optimizer the registered verdict can differ in two ways only.
It can place beyond noise a difference that is within noise under AdamW, and within noise a difference that is beyond noise under Muon.
For example, the looped MoE minus \MoEei{} under AdamW (\seedDiffMthreeUAdamWAbs\mbpb{}, three seeds each) is \seedVerdictMthreeUAdamW{} with AdamW's $s$ (two standard errors \seedTwoSeThreeThreeAdamW) and \seedVerdictPooledMthreeUAdamW{} with pooled $s$ (\seedTwoSeThreeThree).
Under Muon, \MoEloop{} minus the \DeepCtrl{} is \seedVerdictMzeroLoopMzeroDeepMuon{} (\seedVerdictPooledMzeroLoopMzeroDeepMuon{} as registered).

\subsection{Learning-rate calibration}
\label{sec:results-lr}

Across complete 5B-token grid runs, \MoEtw{} and the looped MoE select the same peak learning rate, \selLrAdamW{} under AdamW and \selLrMuon{} under Muon (\cref{tab:results-lrgrid}).
Each rate minimizes validation BPB at an interior grid point.
Muon's optimum is flat and AdamW's is narrow for \MoEtw.
From \lr{1.2e-3} to \lr{3.4e-3}, either model's test BPB changes by at most 2.0\mbpb{} under Muon.
Under AdamW, \MoEtw{}'s loss changes by \gridDeltaMzeroAdamWTwoFourEmThree\mbpb{} at \lr{2.4e-3} (twice the selected rate) and \gridDeltaMzeroAdamWThreeFourEmThree\mbpb{} at \lr{3.4e-3}.

\begin{table}[!ht]
\caption{\lead{\MoEtw{} and the looped MoE select the same learning rate under each optimizer, with interior minima.} Upper blocks use complete 5B-token runs, and their tinted rows mark the lowest validation BPB. The lower block sweeps the 16-layer standard MoE (width 2048, 3.0B stored parameters) for 1.5B tokens (constant to update 1,144, decay over 286 updates). There, bold marks the lowest validation BPB, and the tinted row gives the long-run rate under both optimizers.}
\label{tab:results-lrgrid}
\label{tab:results-lrsweep}
\begin{fittabular}{6}{lNNNNN}
\toprule
\textbf{Main setting, 5B tokens} & \multicolumn{2}{c}{\MoEtw} & \multicolumn{2}{c}{Looped MoE} & Looped MoE $-$ \MoEtw\\
\cmidrule(lr){2-3}\cmidrule(lr){4-5}
Peak learning rate & Val.\ BPB & Test BPB & Val.\ BPB & Test BPB & Test \colunit{mBPB}\\
\midrule
\multicolumn{6}{l}{\textbf{AdamW}}\\
\lr{1.5e-4} & 1.0837 & 1.0814 & 1.0857 & 1.0837 & $+$2.29\\
\lr{3e-4} & 1.0320 & 1.0299 & 1.0311 & 1.0293 & $-$0.63\\
\lr{6e-4} & 1.0064 & 1.0047 & 1.0037 & 1.0018 & $-$2.81\\
\oursrow \ours{\lr{1.2e-3}} & 1.0013 & 0.9995 & 0.9922 & 0.9905 & $-$8.99\\
\lr{1.7e-3} & 1.0014 & 0.9998 & 0.9940 & 0.9920 & $-$7.78\\
\lr{2.4e-3} & 1.0095 & 1.0077 & 0.9988 & 0.9970 & $-$10.70\\
\lr{3.4e-3} & 1.1384 & 1.1354 & 1.0061 & 1.0044 & $-$131.05\\
\grouprule
\multicolumn{6}{l}{\textbf{Muon}}\\
\lr{3e-4} & 1.0080 & 1.0064 & 1.0081 & 1.0063 & $-$0.07\\
\lr{6e-4} & 0.9854 & 0.9837 & 0.9837 & 0.9820 & $-$1.64\\
\lr{1.2e-3} & 0.9770 & 0.9755 & 0.9737 & 0.9719 & $-$3.64\\
\oursrow \ours{\lr{1.7e-3}} & 0.9761 & 0.9744 & 0.9724 & 0.9707 & $-$3.70\\
\lr{2.4e-3} & 0.9763 & 0.9747 & 0.9725 & 0.9707 & $-$4.02\\
\lr{3.4e-3} & 0.9782 & 0.9764 & 0.9732 & 0.9714 & $-$4.97\\
\midrule
\textbf{3.0B setting, 1.5B tokens} & \multicolumn{2}{c}{Standard MoE, AdamW} & \multicolumn{2}{c}{Standard MoE, Muon} & \\
\cmidrule(lr){2-3}\cmidrule(lr){4-5}
Peak learning rate & Val.\ BPB & Test BPB & Val.\ BPB & Test BPB &\\
\midrule
\lr{6e-4} & 1.1041 & 1.1013 & 1.0743 & 1.0718 &\\
\lr{8.5e-4} & 1.0918 & 1.0890 & 1.0613 & 1.0587 &\\
\lr{1.2e-3} & 1.0841 & 1.0813 & 1.0519 & 1.0496 &\\
\oursrow \ours{\lr{1.7e-3}} & \textbf{1.0823} & 1.0797 & 1.0476 & 1.0452 &\\
\lr{2.4e-3} & 1.0945 & 1.0916 & \textbf{1.0466} & 1.0443 &\\
\lr{3.4e-3} & -- & -- & 1.0492 & 1.0466 &\\
\bottomrule
\end{fittabular}
\end{table}

\term{Shared learning rate}
The recipe's gain depends on the shared learning rate (\cref{tab:results-lrgrid}, \cref{fig:residual}b).
At \lr{3e-4} (a quarter of AdamW's calibrated rate), the two models match, \gridGapAdamWThreeEmFourAbs\mbpb{} apart under AdamW and \gridGapMuonThreeEmFourAbs\mbpb{} under Muon.
At twice the calibrated rate under AdamW, the looped MoE is \gridGapAdamWTwoFourEmThreeAbs\mbpb{} lower because \MoEtw{}'s loss rises faster.
At \lr{3.4e-3} under AdamW, \MoEtw{} reaches 1.135 test BPB and the looped MoE is \gridGapAdamWThreeFourEmThreeAbs\mbpb{} lower.

\term{Training length and the 3.0B setting}
The optimum also depends on training length (\cref{tab:results-lroptima}).
Shorter runs branch from the grid runs during the peak-rate phase and decay over their final 20\% of updates.
The fitted optimum is the vertex of a parabola in log learning rate fitted to validation BPB.
Under AdamW, it stays between \lr{1.27e-3} and \lr{1.46e-3} at every length, inside one grid step (\lr{1.2e-3} to \lr{1.7e-3}), and under Muon it falls.
By the rule in \cref{sec:setup-app-train}, a run one eighth as long predicts the 5B optimum within a factor of $\sqrt{2}$ (one grid step).
In the 3.0B setting, the 16-layer standard MoE at 1.5B tokens has its minimum at \lr{1.7e-3} under AdamW and at \lr{2.4e-3} under Muon, where the fitted optimum is \lr{2.23e-3} (\cref{tab:results-lrgrid}).
From these values, every model of the 3.0B setting uses \selLrSzZeroPFiveAdamW{} under both optimizers (\cref{sec:setup-app-train}).

\begin{table}[!ht]
\caption{\lead{With longer training, the fitted learning-rate optimum falls under Muon and moves by less than one grid step under AdamW.} Parabola vertex in log learning rate fitted to main-setting validation BPB.}
\label{tab:results-lroptima}
\centering
\begin{minipage}{96mm}
\begin{fittabular}{5}{lNNNN}
\toprule
 & \multicolumn{2}{c}{AdamW} & \multicolumn{2}{c}{Muon}\\
\cmidrule(lr){2-3}\cmidrule(l){4-5}
Training tokens & \MoEtw & Looped MoE & \MoEtw & Looped MoE\\
\midrule
0.625B & \lr{1.46e-3} & \lr{1.41e-3} & \lr{2.55e-3} & \lr{3.0e-3}\\
1.25B & \lr{1.45e-3} & \lr{1.35e-3} & \lr{2.25e-3} & \lr{2.86e-3}\\
2.5B & \lr{1.45e-3} & \lr{1.32e-3} & \lr{2.07e-3} & \lr{2.40e-3}\\
5B & \lr{1.43e-3} & \lr{1.27e-3} & \lr{1.90e-3} & \lr{1.97e-3}\\
\bottomrule
\end{fittabular}
\end{minipage}
\end{table}

\needspace{5\baselineskip}
\subsection{Every model at 5B tokens}
\label{sec:results-5b}

Every model trained with both optimizers has lower loss under Muon than under AdamW (\cref{tab:results-5b}).
Between three-seed means, Muon lowers \MoEtw{}'s loss by \leadMzeroAbs\mbpb{} (\cref{tab:results-base-horizon}).

\begin{table}[!ht]
\caption{\lead{Every model trained with both optimizers is lower under Muon than under AdamW, by \leadMfourAbs{} to \leadDthreeSixAbs\mbpb{} (last column).} Test BPB and GPU hours at seed 42. $\Delta$ and 3 seeds give test mBPB relative to \MoEtw{} under the same optimizer, at seed 42 and between three-seed means. The last column gives test mBPB under Muon minus under AdamW at seed 42.}
\label{tab:results-5b}
{%
\begin{fittabular}{10}{lNNNNNNNN>{\bfseries}N}
\toprule
 & \multicolumn{4}{c}{AdamW} & \multicolumn{4}{c}{Muon} & Muon $-$\\
\cmidrule(lr){2-5}\cmidrule(lr){6-9}
Model & Test & $\Delta$ & 3 seeds & GPU h & Test & $\Delta$ & 3 seeds & GPU h & AdamW\\
\midrule
\multicolumn{10}{l}{\textit{The five steps}}\\
\MoEtw & 0.9995 & 0 & 0 & 35.4 & 0.9744 & 0 & 0 & 35.6 & $-$25.16\\
\MoEtw{} with finer experts & 0.9953 & $-$4.22 & & 37.7 & 0.9735 & $-$0.86 & & 36.5 & $-$21.79\\
\MoEei & 0.9926 & $-$6.91 & \gapUMzeroAdamW & 40.2 & 0.9702 & $-$4.21 & \gapUMzeroMuon & 38.8 & $-$22.45\\
\MoEei{} with the recipe's expert layout & 0.9936 & $-$5.91 & & 38.9 & 0.9710 & $-$3.36 & & 38.7 & $-$22.61\\
Non-revisit control & 0.9906 & $-$8.89 & \seedDiffMfourMzeroAdamW & 40.5 & 0.9710 & $-$3.33 & \seedDiffMfourMzeroMuon & 40.4 & $-$19.60\\
Looped MoE & 0.9905 & $-$8.99 & \gapMthreeMzeroAdamW & 40.9 & 0.9707 & $-$3.70 & \gapMthreeMzeroMuon & 39.0 & $-$19.87\\
\grouprule
\multicolumn{10}{l}{\textit{Variants of the looped MoE}}\\
Experts revisited & 0.9942 & $-$5.34 & & 41.0 & 0.9712 & $-$3.19 & & 39.2 & $-$23.00\\
Experts and router revisited & 0.9903 & $-$9.20 & & 40.9 & 0.9695 & $-$4.88 & & 39.2 & $-$20.83\\
Looped MoE, 3 visits & 0.9927 & $-$6.83 & & 41.5 & 0.9709 & $-$3.52 & & 41.5 & $-$21.84\\
Looped MoE, 4 visits & 0.9925 & $-$7.08 & & 41.7 & 0.9726 & $-$1.73 & & 41.7 & $-$19.81\\
\grouprule
\multicolumn{10}{l}{\textit{Residual scaling on the middle six layers of \MoEtw{}}}\\
\MoEtw{} with residual scaling & 0.9955 & $-$4.01 & \gapMzeroRMzeroAdamW & 42.0 & 0.9745 & $+$0.10 & & 42.0 & $-$21.05\\
\MoEtw, learning-rate part & 0.9967 & $-$2.85 & & 35.5 & 0.9753 & $+$0.96 & & 37.3 & $-$21.35\\
\MoEtw, initialization part & 0.9972 & $-$2.30 & & 35.4 & 0.9735 & $-$0.91 & & 37.6 & $-$23.77\\
\grouprule
\multicolumn{10}{l}{\textit{Standard MoEs of 24 and 36 layers}}\\
MoE-24 & 0.9974 & $-$2.18 & & 63.4 & 0.9729 & $-$1.53 & & 62.1 & $-$24.50\\
MoE-36 & 1.0099 & $+$10.40 & & 55.9 & 0.9772 & $+$2.78 & & 53.9 & $-$32.78\\
\grouprule
\multicolumn{10}{l}{\textit{Compute scaling and expert scaling, not at matched budgets (\cref{sec:looptitan-compute})}}\\
\MoEloop & 0.9847 & $-$14.80 & & 45.6 & 0.9621 & $-$12.24 & \gapMzeroLoopMzeroMuon & 44.5 & $-$22.60\\
Deeper control & -- & -- & -- & -- & \testMzeroDeepMuon & \gapMzeroDeepMzeroMuonTwoDecSFortyTwo & \seedDiffMzeroDeepMzeroMuon & \gpuhMzeroDeepMuon & --\\
\MoEtwoE & 0.9852 & $-$14.29 & & 38.6 & 0.9645 & $-$9.85 & \gapMzeroTwoEMzeroMuon & 36.8 & $-$20.72\\
\bottomrule
\end{fittabular}}
\tablenote{Blank cells in 3 seeds have seed 42 only, except \MoEei{} with the recipe's expert layout under AdamW (three seeds in \cref{tab:results-seeds}), and the deeper control trains with Muon only. GPU hours are those of the run that completed, and \MoEloop{}, MoE-24, MoE-36 and the deeper control train with activation checkpointing.}
\end{table}

\needspace{7\baselineskip}
\subsection{Longer training and the 3.0B setting}
\label{sec:results-horizon}

Beyond 5B tokens and in the 3.0B setting, runs use one seed, and verdicts use the main setting's $s$ (\cref{tab:results-horizon,sec:results-seeds}).
The 20B run branches from the 5B run before decay (update 3,339), and the 10B run branches from the 20B run, all at the same peak learning rate (\cref{sec:setup-app-train}).
In the 3.0B setting, each 12B run branches from a peak-rate run and decays linearly to zero over its final 20\%.
In every setting, the looped MoE minus the \Ctrl{} lies between \hgapMthreeMfourAdamWHzTwoZero{} and \gapMthreeMfourMuonLarge\mbpb{}, within noise under both optimizers.

\begin{table}[!ht]
\caption{\lead{Beyond 5B tokens and in the 3.0B setting, the looped MoE still matches the \Ctrl{} under both optimizers (last column).} Test BPB at seed 42, and the looped MoE minus the \Ctrl{} in test mBPB. Adjacent test-BPB columns differ by one step of \cref{tab:settings}. The shape change yields \MoEei{} (1.3B stored parameters) and the 24-layer standard MoE of width 1672 (3.0B).}
\label{tab:results-horizon}
\centering
\begin{minipage}{146mm}
\begin{fittabular}{8}{lllNNNN>{\bfseries}N}
\toprule
 & & & & After the & Non-revisit & & Looped MoE $-$\\
Stored & Tokens & Optimizer & Standard MoE & shape change & control & Looped MoE & non-revisit control \colunit{mBPB}\\
\midrule
1.3B & 10B & AdamW & \hTestMzeroAdamWHzOneZero & \hTestUAdamWHzOneZero & \hTestMfourAdamWHzOneZero & \hTestMthreeAdamWHzOneZero & \hgapMthreeMfourAdamWHzOneZero\\
 & & Muon & \hTestMzeroMuonHzOneZero & \hTestUMuonHzOneZero & \hTestMfourMuonHzOneZero & \hTestMthreeMuonHzOneZero & \hgapMthreeMfourMuonHzOneZero\\
 & 20B & AdamW & \hTestMzeroAdamWHzTwoZero & \hTestUAdamWHzTwoZero & \hTestMfourAdamWHzTwoZero & \hTestMthreeAdamWHzTwoZero & \hgapMthreeMfourAdamWHzTwoZero\\
 & & Muon & \hTestMzeroMuonHzTwoZero & \hTestUMuonHzTwoZero & \hTestMfourMuonHzTwoZero & \hTestMthreeMuonHzTwoZero & \hgapMthreeMfourMuonHzTwoZero\\
\grouprule
3.0B & 12B & AdamW & \testMzeroAdamWLarge & \testUAdamWLarge & \testMfourAdamWLarge & \testMthreeAdamWLarge & \gapMthreeMfourAdamWLarge\\
 & & Muon & \testMzeroMuonLarge & \testUMuonLarge & \testMfourMuonLarge & \testMthreeMuonLarge & \gapMthreeMfourMuonLarge\\
 & 24B & AdamW & \testMzeroAdamWLargeLong & -- & \testMfourAdamWLargeLong & \testMthreeAdamWLargeLong & \gapMthreeMfourAdamWLargeLong\\
 & & Muon & \testMzeroMuonLargeLong & -- & \testMfourMuonLargeLong & \testMthreeMuonLargeLong & \gapMthreeMfourMuonLargeLong\\
\bottomrule
\end{fittabular}
\end{minipage}
\end{table}

\begin{figure}[!ht]
\fig{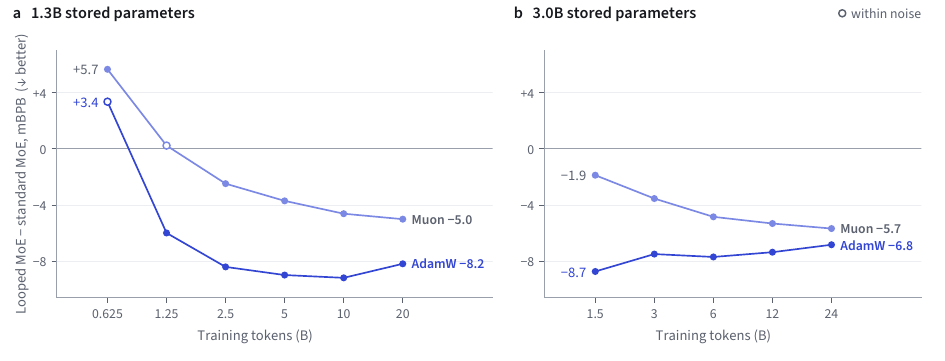}{Two line charts with a shared vertical axis, the looped MoE's test loss minus the standard MoE's in mBPB, against training tokens on a logarithmic axis. A filled dot is beyond noise and a hollow dot within noise, two standard errors of a one-seed difference, \seedTwoSeOneOneAdamW{} for AdamW and \seedTwoSeOneOneMuon{} for Muon. Panel a, 1.3B stored parameters, 0.625B to 20B tokens. AdamW: +3.37, -5.99, -8.41, -8.99, -9.19, -8.19. Muon: +5.68, +0.24, -2.47, -3.70, -4.62, -5.01. AdamW is within noise at 0.625B tokens and beyond noise from 1.25B. Muon is beyond noise at 0.625B, within noise at 1.25B and beyond noise from 2.5B. Panel b, 3.0B stored parameters, 1.5B to 24B tokens. AdamW: -8.74, -7.49, -7.70, -7.36, -6.83. Muon: -1.87, -3.53, -4.84, -5.31, -5.67. Every point of both lines is beyond noise, the Muon point at 1.5B tokens only just.}
\caption{\lead{The looped recipe has no gain at 0.625B tokens (\textbf{a}). Its gain appears with longer training and, under Muon, grows with length at both scales (\textbf{a}, \textbf{b}).} Looped MoE minus standard MoE in test mBPB under the same optimizer (learning rates in \cref{sec:results-lr}, one seed per point). Points on one curve share early training. Filled dots are beyond noise and hollow dots within noise (two standard errors of a one-seed difference, \seedTwoSeOneOneAdamW\mbpb{} under AdamW and \seedTwoSeOneOneMuon{} under Muon, \cref{sec:results-seeds}).}
\label{fig:results-measurable}
\end{figure}

\term{Training length}
The recipe's gain also depends on training length (\cref{fig:results-measurable}).
At calibrated learning rates and 0.625B tokens, the looped MoE has higher loss than \MoEtw{} by \hgapMthreeMzeroAdamWHzZeroPSixTwoFiveAbs\mbpb{} under AdamW (within noise, two standard errors \seedTwoSeOneOneAdamW) and \hgapMthreeMzeroMuonHzZeroPSixTwoFiveAbs{} under Muon (beyond noise).
Under AdamW, the looped MoE is lower from 1.25B tokens.
Under Muon, the two match at 1.25B tokens, and the looped MoE is lower from 2.5B.
In the 3.0B setting, the looped MoE has lower loss than the 16-layer standard MoE at every length from 1.5B tokens under both optimizers.
At 24B tokens, the 3.0B setting compares the standard MoE, the \Ctrl{} and the looped MoE (\cref{tab:results-horizon}).
In both panels, the runs of one curve share their early training, so its points are not independent.

\needspace{7\baselineskip}
\subsection{The shape change, step by step and by depth}
\label{sec:results-ladder}
\label{sec:results-depth}

\begin{figure}[!ht]
\fig{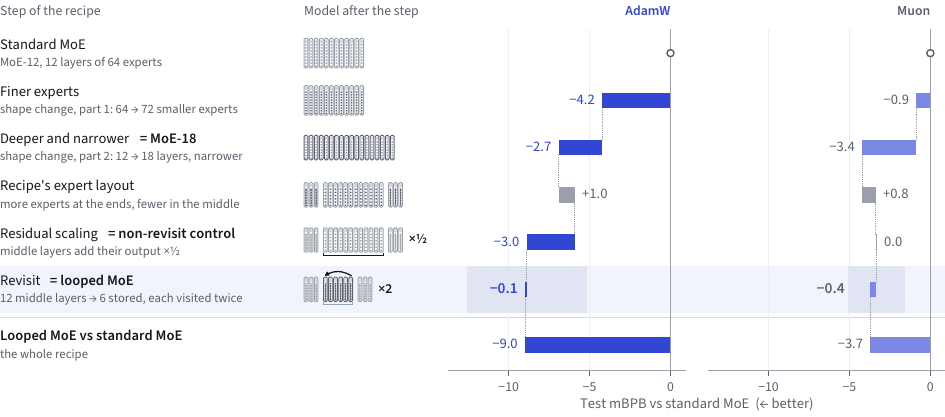}{Waterfall of the looped MoE's gain over MoE-12 under AdamW and Muon, one row per design decision, each bar starting where the row above ended: finer experts (-4.2, -0.9), deeper and narrower, reaching MoE-18 (-2.7, -3.4), the recipe's expert layout (+1.0, +0.8), residual scaling, reaching the non-revisit control (-3.0, 0.0), and parameter revisit, reaching the looped MoE (-0.1, -0.4). On the revisit row a band around the non-revisit control's loss, plus or minus \seedTwoSeOneOneAdamW{} in the AdamW column and plus or minus \seedTwoSeOneOneMuon{} in the Muon column, contains the end of the revisit bar in both columns. The whole recipe is -9.0 and -3.7.}
\caption{\lead{The shape change is beyond noise as a whole under both optimizers. Of its two steps alone, finer experts are beyond noise under AdamW and the deeper, narrower stack under Muon.} Bars use seed 42, with total recipe change \gapMthreeMzeroAdamWSFortyTwo{} (AdamW) and \gapMthreeMzeroMuonSFortyTwo\mbpb{} (Muon), versus \gapMthreeMzeroAdamW{} and \gapMthreeMzeroMuon{} between three-seed means (\cref{fig:gain}). Gray bars raise loss, and schematics follow \cref{fig:control}. The revisit band marks two standard errors around the \Ctrl{}.}
\label{fig:results-ladder}
\end{figure}

\term{Single steps}
At seed 42, the shape change's two steps gain differently across optimizers (\cref{fig:results-ladder}, models in \cref{tab:setup-arms}).
Finer experts gain \gapGMzeroAdamWAbs\mbpb{} under AdamW and \gapGMzeroMuonAbs{} under Muon, and the deeper, narrower stack gains \gapUGAdamWAbs{} and \gapUGMuonAbs.
Under AdamW, finer experts alone are \seedVerdictOneGMzeroAdamW{} (two standard errors \seedTwoSeOneOneAdamW\mbpb{}), and the deeper, narrower stack is within noise.
Under Muon, the deeper, narrower stack alone is \seedVerdictOneUGMuon{}, and finer experts are within noise.
The recipe's expert layout changes loss by \gapRUAdamW{} and \gapRUMuon\mbpb{} (within noise).
Residual scaling changes it by \gapMfourRAdamWSFortyTwo{} under AdamW (within noise, and \seedDiffMfourRAdamW{} over three seeds each, \seedVerdictMfourRAdamW) and $+$\gapMfourRMuonAbsTwoDec{} under Muon (\seedVerdictOneOwnMfourRMuon).
Revisit gains \gapMthreeMfourAdamWAbsSFortyTwo{} and \gapMthreeMfourMuonAbsSFortyTwo\mbpb{} at seed 42 and costs \seedDiffMthreeMfourAdamWAbs{} and \seedDiffMthreeMfourMuonAbs{} over three seeds, \seedVerdictMthreeMfourAdamW{} under both optimizers (\cref{tab:seeds}).

\begin{figure}[!ht]
\fig{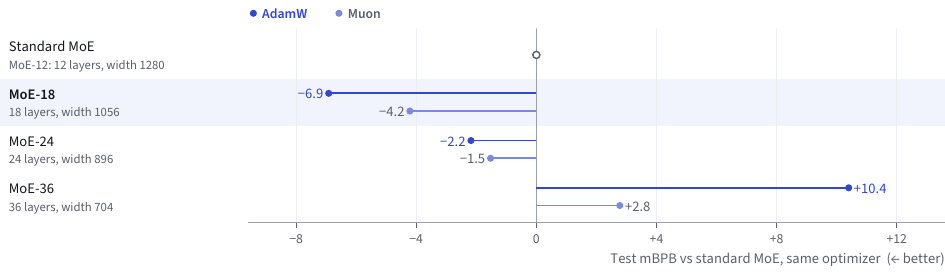}{Lollipop chart with four rows, AdamW above and Muon below in each row, in test mBPB relative to MoE-12 under the same optimizer: MoE-12 at 0, MoE-18 at -6.9 and -4.2 (the lowest), MoE-24 at -2.2 and -1.5, MoE-36 at +10.4 and +2.8.}
\caption{\lead{Of four depths at matched budgets, 18 layers give the lowest loss under both optimizers.} Standard MoEs of 12, 18, 24 and 36 layers in test mBPB relative to \MoEtw{} under the same optimizer (seed 42).}
\label{fig:results-depth}
\end{figure}

\term{Four depths}
Among four standard MoEs at matched budgets (seed 42), \MoEei{} has the lowest loss under both optimizers (\cref{fig:results-depth}, \cref{tab:setup-app-arch}).
MoE-24 is \gapDtwoFourUAdamWAbs\mbpb{} higher under AdamW and \gapDtwoFourUMuonAbs{} under Muon, and MoE-36 is \gapDthreeSixUAdamWAbs{} and \gapDthreeSixUMuonAbs\mbpb{} higher.
These four differences are beyond noise.

\subsection{Residual scaling, its two parts and the learning rate}
\label{sec:results-residual}

On \MoEtw{}, residual scaling and its two parts change only the middle six layers (\cref{fig:residual}a, raw losses in \cref{tab:results-5b}).
The learning-rate part halves their output projections' learning rate and doubles the weight-decay coefficient, so decay per update stays unchanged.
The initialization part halves those projections' initialization.
In the recipe, residual scaling lowers the \Ctrl{}'s loss by \seedDiffMfourRAdamWAbs\mbpb{} under AdamW over three seeds each (\seedVerdictMfourRAdamW, \cref{tab:seeds}).
On \MoEtw{} under AdamW, it changes the loss by \gapMzeroRMzeroAdamW\mbpb{} between three-seed means (\verdictGapMzeroRMzeroAdamW).
Its two parts have seed 42 only and change \MoEtw{}'s loss by \seedDiffMzeroLRMzeroAdamW{} and \seedDiffMzeroIMzeroAdamW\mbpb{} against its three-seed mean, each within noise.
Under Muon, residual scaling changes the \Ctrl{}'s loss by \seedDiffMfourRMuon\mbpb{} and \MoEtw{}'s by \seedDiffMzeroRMzeroMuon, and its two parts change \MoEtw{}'s by \seedDiffMzeroLRMzeroMuonTwoDec{} and \seedDiffMzeroIMzeroMuonTwoDec\mbpb{}, each with seed 42 against three-seed means.
Each is within noise, and the learning-rate part only just (two standard errors \seedTwoSeThreeOneMuonTwoDec\mbpb{}).

Under AdamW, halving the whole model's learning rate gives the \lr{6e-4} grid point (\cref{tab:results-lrgrid}).
The weight-decay coefficient is unchanged, so decay per update halves in every layer.
Global halving (seed 42) raises \MoEtw{}'s loss by \seedDiffMzeroHalfLrMzeroAdamWAbs\mbpb{} against its three-seed mean (\seedVerdictMzeroHalfLrMzeroAdamW) and the looped MoE's by \lrHalfGlobalMthreeAdamWAbs\mbpb{} against its seed-42 run.
On \MoEtw{}, the learning-rate part and global halving differ by \gapLrHalfLocalGlobalAdamWAbs\mbpb{} (\seedVerdictOneLrHalfLocalGlobalAdamW).
Muon's grid has no point at half the calibrated rate.
Its two neighbors change \MoEtw{}'s seed-42 loss by \gridDeltaMzeroMuonOneTwoEmThree\mbpb{} at \lr{1.2e-3} (within noise) and \gridDeltaMzeroMuonSixEmFour\mbpb{} at \lr{6e-4}.

\subsection{Adding compute}
\label{sec:results-compute}

Under Muon, the \DeepCtrl{} obtains \shareMzeroDeepofMzeroLoopMuon\% of one more visit's gain over three seeds and \shareMzeroDeepofMzeroLoopMuonSFortyTwo\% with seed 42 alone (\cref{tab:seeds}).
The loss of \MoEloop{} is \seedDiffMzeroLoopMzeroDeepMuonAbs\mbpb{} below the \DeepCtrl{}'s, lower in every seed and \seedVerdictMzeroLoopMzeroDeepMuon{} (two standard errors \seedTwoSeMzeroLoopMzeroDeepMuon\mbpb{}).
This difference is about a tenth of the visit's gain.
The two models also differ in experts per layer (64 against 42) and residual scaling.
Under AdamW, \MoEloop{} and \MoEtwoE{} match, \gapMzeroLoopMzeroTwoEAdamWAbs\mbpb{} apart with one seed each.

\MoEloop{} has residual scaling, which alone changes \MoEtw{}'s loss by \gapMzeroRMzeroAdamW\mbpb{} under AdamW (three seeds each, \verdictGapMzeroRMzeroAdamW) and \seedDiffMzeroRMzeroMuon{} under Muon (seed 42 against three seeds, \seedVerdictMzeroRMzeroMuon, \cref{sec:results-residual}).
With seed 42, one more visit gains \gapMzeroLoopMzeroAdamWAbs\mbpb{} over \MoEtw{} under AdamW and \gapMzeroLoopMzeroMuonAbsSFortyTwo{} under Muon.
Against \MoEtw{} with that residual scaling, the visit gains \gapMzeroLoopMzeroRAdamWAbs{} and \gapMzeroLoopMzeroRMuonAbs\mbpb{}.
We did not train the visit without residual scaling.

\needspace{7\baselineskip}
\subsection{Gains as training tokens}
\label{sec:results-tokens}

We convert each gain into equivalent training tokens on \MoEtw{} (\cref{tab:results-tokens}).
We interpolate \MoEtw{}'s test loss linearly in log training tokens between the 5B, 10B and 20B endpoints (one seed each) and never extrapolate.
The three runs share early training (\cref{sec:results-horizon}), so the three points are not independent.

\begin{table}[!ht]
\caption{\lead{By interpolation, Muon is worth about \tokEqMuonSwitchMain{} times \MoEtw{}'s training tokens at 5B. Under Muon, spending one more visit's compute on training tokens would gain at least as much as the visit.} Every gain uses seed 42 at 5B tokens. Tokens is the multiple of 5B tokens where \MoEtw{} reaches that test loss under the same optimizer. The lower block compares the two optimizers on \MoEtw{} at equal test loss.}
\label{tab:results-tokens}
\centering
\begin{minipage}{136mm}
\begin{fittabular}{5}{lNNNN}
\toprule
 & \multicolumn{2}{c}{AdamW} & \multicolumn{2}{c}{Muon}\\
\cmidrule(lr){2-3}\cmidrule(l){4-5}
Model & Gain \colunit{mBPB} & Tokens \colunit{$\times$} & Gain \colunit{mBPB} & Tokens \colunit{$\times$}\\
\midrule
\MoEei{} (shape change) & \gapUMzeroAdamWAbsSFortyTwo & \tokEqUAdamWTwoDec & \gapUMzeroMuonAbsSFortyTwo & \tokEqUMuonTwoDec\\
Non-revisit control & \gapMfourMzeroAdamWAbs & \tokEqMfourAdamWTwoDec & \gapMfourMzeroMuonAbs & \tokEqMfourMuonTwoDec\\
Looped MoE (the recipe) & \gapMthreeMzeroAdamWAbsSFortyTwo & \tokEqMthreeAdamWTwoDec & \gapMthreeMzeroMuonAbsSFortyTwo & \tokEqMthreeMuonTwoDec\\
\grouprule
\MoEloop & \gapMzeroLoopMzeroAdamWAbs & \tokEqMzeroLoopAdamWTwoDec & \gapMzeroLoopMzeroMuonAbsSFortyTwo & \tokEqMzeroLoopMuonTwoDec\\
Deeper control & -- & -- & \gapMzeroDeepMzeroMuonAbs & \tokEqMzeroDeepMuonTwoDec\\
\MoEtwoE & \gapMzeroTwoEMzeroAdamWAbs & \tokEqMzeroTwoEAdamWTwoDec & \gapMzeroTwoEMzeroMuonAbsSFortyTwo & \tokEqMzeroTwoEMuonTwoDec\\
\MoEtw{} on \flopsRatioMzeroLoopTwoDec{} times the tokens & \tokGainMzeroVisitFlopsAdamWAbs & \flopsRatioMzeroLoopTwoDec & \tokGainMzeroVisitFlopsMuonAbs & \flopsRatioMzeroLoopTwoDec\\
\bottomrule
\end{fittabular}

\vspace{8pt}
\begin{fittabular}{3}{lNN}
\toprule
\MoEtw{} at equal test loss & Tokens under Muon & Tokens under AdamW \colunit{$\times$ Muon's}\\
\midrule
At Muon's 5B loss (a gain of \testMzeroAdamWMuonDiff\mbpb{} over AdamW) & 5B & \tokEqMuonSwitchMainTwoDec\\
At AdamW's 10B loss & \tokMuonReachesAdamWHzOneZero{}B & \tokEqMuonSwitchHzOneZeroTwoDec\\
At AdamW's 20B loss & \tokMuonReachesAdamWHzTwoZero{}B & \tokEqMuonSwitchHzTwoZeroTwoDec\\
\bottomrule
\end{fittabular}
\tablenote{At \flopsRatioMzeroLoopTwoDec{} times the tokens (\MoEtw{} at the training FLOPs of \MoEloop{}), the gain is an interpolated estimate, not a run. A quadratic in log tokens changes each multiple by at most 0.03 and gives \tokGainMzeroVisitFlopsAdamWQuadAbs{} and \tokGainMzeroVisitFlopsMuonQuadAbs\mbpb{} for that row.}
\end{minipage}
\end{table}

\needspace{7\baselineskip}
\subsection{The optimizer and \texorpdfstring{\Base}{LoopTitan-Base} in every setting}
\label{sec:results-base}

Muon lowers every model's loss in every setting (\cref{tab:results-5b,tab:results-horizon}).
On the standard MoE, Muon's gain shrinks with training length and exceeds the recipe's gain under AdamW in every setting (\cref{tab:results-base-horizon}).
Under Muon, the looped MoE is within \rangeMthreeUMuonMaxAbs\mbpb{} of \Base{} in every setting.
Every difference is within noise (two standard errors \seedTwoSeMthreeUMuon\mbpb{} for three seeds each at 5B tokens, \seedTwoSeOneOneMuon{} for one seed each elsewhere).

\begin{table}[!ht]
\caption{\lead{In every setting, changing the optimizer gains more than the looped recipe gains under AdamW.} Differences $A-B$ in test mBPB at calibrated learning rates. Ratio divides the first difference by the second. In the 3.0B setting, the standard MoE has 16 layers of width 2048, and \Base{} is the 24-layer standard MoE of width 1672 trained with Muon.}
\label{tab:results-base-horizon}
\begin{fittabular}{7}{llNNNNN}
\toprule
 & & Muon $-$ AdamW, & Looped MoE $-$ standard & & \Base{} $-$ standard & \best{Looped MoE $-$ \Base,}\\
Stored & Tokens & standard MoE & MoE, both with AdamW & Ratio & MoE with AdamW & \best{both with Muon}\\
\midrule
1.3B & 5B & \leadMzero & \gapMthreeMzeroAdamW & \ratioLeadMzeroGainAdamW & \xgapUMuonMzeroAdamW & \best{\seedDiffMthreeUMuon}\\
 & 10B & \hleadMzeroHzOneZero & \hgapMthreeMzeroAdamWHzOneZero & \ratioLeadMzeroGainAdamWHzOneZero & \xgapUMuonMzeroAdamWHzOneZero & \best{\hgapMthreeUMuonHzOneZero}\\
 & 20B & \hleadMzeroHzTwoZero & \hgapMthreeMzeroAdamWHzTwoZero & \ratioLeadMzeroGainAdamWHzTwoZero & \xgapUMuonMzeroAdamWHzTwoZero & \best{\hgapMthreeUMuonHzTwoZero}\\
\grouprule
3.0B & 12B & \leadMzeroLarge & \gapMthreeMzeroAdamWLarge & \ratioLeadMzeroGainAdamWLarge & \xgapUMuonMzeroAdamWLarge & \best{\gapMthreeUMuonLarge}\\
\bottomrule
\end{fittabular}
\tablenote{The 5B row gives differences of three-seed means (\cref{tab:settings}), and other rows use seed 42.}
\end{table}

\term{Training time}
At matched compute and 5B tokens, \MoEei{} took \gpuhUMuon{} GPU hours under Muon and \gpuhUAdamW{} under AdamW, against \gpuhMzeroMuon{} and \gpuhMzeroAdamW{} for \MoEtw{} (\gpuhPctUMzeroMuon\% and \gpuhPctUMzeroAdamW\%, every model in \cref{tab:results-5b}).
These GPU hours come from training runs on 8 H200 GPUs, not a controlled timing benchmark.
Two seeds of one model can differ by up to 11\% (\gpuhSeedMinMthreeAdamW{} and \gpuhSeedMaxMthreeAdamW{} hours for the looped MoE under AdamW), so only the difference under AdamW exceeds between-seed spread.

\needspace{7\baselineskip}
\subsection{Nine published looped recipes under one protocol}
\label{sec:results-recipes}

\begin{table}[!ht]
\caption{\lead{Under Muon, \Base{} matches three of the looped MoE's four configurations.} Rows are the twelve configurations of the nine published designs that \moeloop{} implements today, each our adaptation under one protocol, not the design in its original setting. Seed 42 at 5B tokens, with AdamW at \lr{1.2e-3} and Muon at \lr{1.7e-3} unless noted. The last three columns give test mBPB minus \MoEtw{} under the same optimizer and minus \Base{} under Muon. \Cref{fig:base}b subtracts the three-seed mean of \Base{} instead of its seed 42, which lowers every one-seed value by the same amount, and gives the looped MoE between three-seed means (\seedDiffMthreeUMuon\mbpb, \cref{tab:seeds}).}
\label{tab:results-recipes-config}
\label{tab:results-recipes}
{\addtolength{\tabcolsep}{-0.5pt}%
\newcommand{\keeps}[1]{\multicolumn{9}{@{}p{\linewidth}@{}}{\leftskip=1em\relax\leavevmode\color{hlMid}#1}\\\addlinespace[4pt]}%
\newcommand{\keepsv}[1]{\multicolumn{9}{@{}p{\linewidth}@{}}{\leftskip=2em\relax\leavevmode\color{hlMid}#1}\\\addlinespace[4pt]}%
\begin{tabular*}{\linewidth}{@{\extracolsep{\fill}}lllNNNNNN@{}}
\toprule
 & & & \multicolumn{3}{c}{Budgets \colunit{\%}} & \multicolumn{2}{c}{vs \MoEtw} & vs \Base\\
\cmidrule(lr){4-6}\cmidrule(lr){7-8}\cmidrule(l){9-9}
Design & Source & Layers & Stored & FLOPs & KV & AdamW & Muon & Muon\\
\midrule
Middle half run twice (the looped MoE) & \citep{smelt} & 3 + 6 $\times$ 2 + 3 & $-$0.3 & $+$2.4 & $+$3.1 & $-$8.99 & $-$3.70 & $+$0.51\\
\keeps{All core parameters shared, core residuals $\times$ 1/2}
\hspace*{1em}three visits & & 3 + 4 $\times$ 3 + 3 & $+$0.1 & $+$2.5 & $+$3.1 & $-$6.83 & $-$3.52 & $+$0.69\\
\keepsv{Core residuals $\times$ 1/3, 18 executed layers}
\hspace*{1em}four visits & & 3 + 3 $\times$ 4 + 3 & 0.0 & $+$2.7 & $+$3.1 & $-$7.08 & $-$1.73 & $+$2.48\\
\keepsv{Core residuals $\times$ 1/4}
\hspace*{1em}per-visit residual split & \citep{latentloop} & 3 + 6 $\times$ 2 + 3 & $-$0.3 & $+$2.4 & $+$3.1 & $-$7.51 & $-$4.09 & $+$0.12\\
\keepsv{Residual scales (0.25, 1) in place of (0.5, 0.5), not the complete method of its source}
\grouprule
Double-expert core visited twice & \citep{foil} & 2 + 4 $\times$ 2 + 2 & 0.0 & $+$0.1 & 0.0 & $-$2.29 & $-$0.83 & $+$3.38\\
\keeps{Core with twice the experts and half the layers, attention per visit, shared experts, router and norms}
Short middle loop, four visits & \citep{metaloop} & 2 + 2 $\times$ 4 + 2 & $-$0.1 & $+$0.2 & 0.0 & $+$2.72 & $+$3.88 & $+$8.09\\
\keeps{Independent prelude and coda around a shared middle loop, fixed depth, full backpropagation through time}
All layers twice, LoRA per visit & \citep{relaxed} & 6 $\times$ 2 & $+$0.3 & $+$3.1 & 0.0 & $-$2.19 & $+$3.93 & $+$8.14\\
\keeps{Layers shared cyclically, zero-initialized LoRA of rank 160 on the attention projections, experts not adapted}
Experts shared by four layers & \citep{tying} & 3 groups $\times$ 4 & $+$0.2 & $+$0.6 & 0.0 & $+$4.42 & $+$8.07 & $+$12.29\\
\keeps{Consecutive layers share their experts only, with attention, router and norms per layer and expert learning rate $\times$ 0.5}
Sampled depth, random start & \citep{huginn} & 1 + 2 $\times$ 4 + 1 & 0.0 & $-$0.1 & $+$4.2 & $+$435.61 & $+$10.68 & $+$14.89\\
\multicolumn{9}{@{}p{\linewidth}@{}}{\leftskip=1em\relax\leavevmode\color{hlMid}Depth sampled per micro-batch with mean 4, gradient through the last 2 visits, evaluated at depth 4}\\
\hspace*{1em}with AdamW at \lr{3e-4} & & & & & & $+$49.57 & -- & --\\
\addlinespace[4pt]
Tied core $\times$8, input injected & \citep{sandwich} & 2 + 1 $\times$ 8 + 2 & $-$0.3 & $+$2.4 & 0.0 & $+$15.90 & $+$16.33 & $+$20.54\\
\keeps{Prelude and coda around the core, the state decays and the input is injected before each pass, top-2 routing}
Loop with two EMA memories & \citep{loom} & 4 $\times$ 3 & 0.0 & $+$0.4 & 0.0 & $+$7.96 & $+$17.47 & $+$21.68\\
\keeps{Two exponential-moving-average (EMA) memories, written then read, per-visit routers, the embedding re-injected}
Loop with a learned exit gate & \citep{ouro} & 4 $\times$ 3 & $-$0.1 & 0.0 & 0.0 & $+$33.75 & $+$39.26 & $+$43.47\\
\multicolumn{9}{@{}p{\linewidth}@{}}{\leftskip=1em\relax\leavevmode\color{hlMid}Stage~I, a shared exit gate trained with the expected per-token loss minus an entropy term, all visits executed, width halved to 640}\\
\bottomrule
\end{tabular*}}
\tablenote{Rows follow \cref{fig:base}b, with retained mechanisms in gray. Layers are prefix + core $\times$ visits + suffix. Budgets are stored parameters, training FLOPs per token and KV cache relative to \MoEtw, with expected FLOPs over depth for sampled depth. Reference test BPB is \testMzeroAdamW{} (\MoEtw{} with AdamW), \testMzeroMuon{} (\MoEtw{} with Muon) and \testUMuon{} (\Base). The three-seed mean of \Base{} in \cref{fig:base}b changes no verdict on the nine recipes.}
\end{table}

\moeloop{} implements nine published looped recipes as twelve configurations of one model under our training setup (\cref{tab:results-recipes-config}).
The looped MoE recipe~\citep{smelt} has four configurations (two, three and four visits, plus another paper's per-visit residual split~\citep{latentloop}), and the other eight recipes have one each.
These runs record the present state of our implementations, not the designs in their original settings.

\term{The looped MoE's four configurations}
Under Muon, \Base{} matches the two-visit, three-visit and residual-split configurations of the looped MoE at seed 42.
The four-visit configuration is \gapSmeltVfourUMuonAbs\mbpb{} above \Base{} (beyond noise).
The per-visit residual-split variant matches the looped MoE under both optimizers (\gapLatentLoopResidualMthreeAdamWAbs{} and \gapLatentLoopResidualMthreeMuonAbs\mbpb{} apart).
The two-visit, three-visit and residual-split configurations are also below \MoEtw{} under both optimizers, each beyond noise.
With four visits, this holds under AdamW only, and under Muon the 1.73\mbpb{} gain is just within noise (two standard errors \seedTwoSeOneOneMuonTwoDec).

\term{Adaptation}
Each configuration keeps its source's mechanism, adapting depth, width and layout to stay within 5\% of \MoEtw{} on the three budgets.
Where a source uses a dense or pretrained model, we apply its loop mechanism to an MoE trained from scratch.
Beyond the summaries in \cref{tab:results-recipes-config}, the repository lists every adaptation in full.
We evaluate each configuration at its inference depth.
This comparison tests whole recipes under one protocol and does not rank loop mechanisms.

\term{Learning rate}
We calibrate learning rates once per optimizer on \MoEtw{} and the looped MoE, and other configurations inherit them (\cref{sec:results-lr}).
The one exception is our adaptation of the sampled-depth design~\citep{huginn} under AdamW.
At our protocol's AdamW rate (\lr{1.2e-3}), this adaptation does not train (\gapHuginnMzeroAdamWAbs\mbpb{} above \MoEtw{}).
Checking four learning rates selects \lr{3e-4}, \gapHuginnMzeroAdamWUpdatedAbs\mbpb{} above \MoEtw.

\section{Proofs}
\label{sec:proof-app}

We prove \cref{prop:residual} for the two implemented update rules, with every condition the proof uses.
We also prove \cref{lem:proof-step} on the size of one update under each optimizer (\cref{sec:residual}) and measure that size in training (\cref{sec:results-stepsize}).

\subsection{Setting and update rules}
\label{sec:proof-setting}

\term{Model}
Let $\mathcal{B}$ be the set of layers that carry the residual scale $a > 0$.
One visit of a layer $\ell \in \mathcal{B}$ maps the residual stream $h$ to $h''$ by
\begin{align}
\label{eq:proof-attn}
h' &= h + a\,W_{\mathrm{att}}^{\ell}\;c_\ell(x;\omega),\\
\label{eq:proof-ffn}
h'' &= h' + a \sum_{j \in \mathcal{T}_\ell(y;\omega)} p_{\ell,j}(y;\omega)\;W_{2,j}^{\ell}\,\bigl(\operatorname{SiLU}(W_{1,j}^{\ell}\,y) \odot W_{3,j}^{\ell}\,y\bigr),
\end{align}
with the normalized inputs $x = \mathrm{Norm}(h)$ and $y = \mathrm{Norm}(h')$.
Here $c_\ell$ is the concatenated output of the attention heads, $\mathcal{T}_\ell$ is the set of experts that the router selects (those with the largest router probabilities), and $p_{\ell,j}$ are these probabilities renormalized over $\mathcal{T}_\ell$.
No bias and no normalization follows $W_{\mathrm{att}}^{\ell}$ or $W_{2,j}^{\ell}$.
A layer may be visited several times in one forward pass, as the revisited layers of a looped MoE are, and every visit uses the same $a$.
Layers outside $\mathcal{B}$ have the same form with $a = 1$.

We split the parameters into $(W, \omega)$.
The tuple $W$ collects the output projections of the layers in $\mathcal{B}$, that is, every $W_{\mathrm{att}}^{\ell}$ and every $W_{2,j}^{\ell}$ with $\ell \in \mathcal{B}$.
Everything else is $\omega$, namely the embedding, the head, the routers, the normalization scales, the query, key and value projections, the expert matrices $W_{1,j}$ and $W_{3,j}$, and all parameters of the layers outside $\mathcal{B}$.
Let $L_t(a, W, \omega)$ be the training objective on the batch of update $t$ (the cross-entropy plus the router balance term).
In \eqref{eq:proof-attn} and \eqref{eq:proof-ffn} the scale $a$ and an output projection occur only as a product, and nothing else in the network depends on $a$.
Hence, for every $a > 0$,
\begin{equation}
\label{eq:proof-reparam}
L_t(a, W, \omega) = L_t(1, aW, \omega),
\end{equation}
and the same identity holds for the function that the network computes on any input.

\term{Update rules}
Both optimizers update a parameter $\Theta$ from its own gradients $G_1, \dots, G_t$ and have the form
\begin{equation}
\label{eq:proof-step}
\Theta_t = (1 - \eta_t\lambda)\,\Theta_{t-1} - \eta_t\,U_t ,
\end{equation}
where $\eta_t$ is the scheduled learning rate, $\lambda$ the coefficient of the decoupled weight decay, and $U_t$ a direction that depends on $G_1, \dots, G_t$ only.
Optimizer states start at zero.
AdamW \citep{loshchilov2019adamw} has $m_0 = v_0 = 0$ and, with all operations elementwise,
\begin{equation}
\label{eq:proof-adamw}
m_t = \beta_1 m_{t-1} + (1-\beta_1)\,G_t, \quad
v_t = \beta_2 v_{t-1} + (1-\beta_2)\,G_t^{2}, \quad
U^{\mathrm{A}}_t(\varepsilon) = \frac{m_t / (1-\beta_1^{t})}{\sqrt{v_t / (1-\beta_2^{t})} + \varepsilon}.
\end{equation}
Our runs use $\beta_1 = 0.9$, $\beta_2 = 0.95$, $\varepsilon = 10^{-8}$ and $\lambda = 0.1$.

Muon \citep{jordanmuon}, in the distributed implementation of TorchTitan \citep{torchtitan}, updates a matrix $\Theta \in \mathbb{R}^{p \times q}$ with $B_0 = 0$ by
\begin{align}
\label{eq:proof-muon-mom}
B_t &= \mu B_{t-1} + (1-\mu)\,G_t, \qquad N_t = (1-\mu)\,G_t + \mu B_t,\\
\label{eq:proof-muon-ns}
X_0 &= \frac{N_t}{\max(\normF{N_t},\, \varepsilon_{\mathrm{M}})}, \qquad
X_{j+1} = c_1 X_j + \bigl(c_2\,X_j X_j^{\top} + c_3\,(X_j X_j^{\top})^{2}\bigr) X_j,\\
\label{eq:proof-muon-dir}
U^{\mathrm{M}}_t &= \rho\,X_J, \qquad \rho = 0.2\sqrt{\max(p, q)} .
\end{align}
$N_t$ is the Nesterov momentum and $X_J$ is the result of $J$ Newton--Schulz iterations, which approximate the orthogonal polar factor of $N_t$.
The normalization and the iteration run in 16-bit floating point, on the transpose when $p > q$.
Our runs use $\mu = 0.95$, $J = 5$, $(c_1, c_2, c_3) = (3.4445, -4.775, 2.0315)$, $\varepsilon_{\mathrm{M}} = 10^{-7}$ and $\lambda = 0.1$.
The factor $\rho$ depends on the shape of $\Theta$ only, and the decay in \eqref{eq:proof-step} uses $\eta_t$ without it.
A bank of experts is updated one expert matrix at a time, so each $W_{2,j}$ has its own norm in \eqref{eq:proof-muon-ns} and its own iteration.

\subsection{The equivalence}
\label{sec:proof-main}

The statements here hold in exact arithmetic (\cref{sec:proof-breaks} covers floating point).

\begin{lemma}[Gradients]
\label{lem:proof-grad}
Let $V = aW$. The gradients of the two sides of \eqref{eq:proof-reparam} satisfy
\[
\nabla_W L_t(a, W, \omega) = a\,\nabla_V L_t(1, V, \omega), \qquad
\nabla_\omega L_t(a, W, \omega) = \nabla_\omega L_t(1, V, \omega).
\]
\end{lemma}
\begin{proof}
Apply the chain rule to $W \mapsto L_t(1, aW, \omega)$. Two details of the implementation do not affect the identity. First, an output projection that is used at several visits receives the sum of the gradients of its visits. This sum is the total derivative, to which the chain rule applies. Second, the selection $\mathcal{T}_\ell$ is piecewise constant, and backpropagation treats it as a constant. The two forward passes have identical activations by \eqref{eq:proof-reparam}, so they select the same experts, and the identity holds for the gradients that are actually computed.
\end{proof}

\begin{lemma}[Rescaled gradients]
\label{lem:proof-scale}
Let $\gamma > 0$. Write $U_t[G]$ for the direction computed from the history $G = (G_1, \dots, G_t)$ with optimizer states that start at zero, and $\gamma G$ for the history $(\gamma G_1, \dots, \gamma G_t)$.
\begin{enumerate}[nosep]
\item[\textup{(a)}] The AdamW direction of $\gamma G$ with $\varepsilon$ is the direction of $G$ with $\varepsilon / \gamma$, that is, $U^{\mathrm{A}}_t(\varepsilon)[\gamma G] = U^{\mathrm{A}}_t(\varepsilon/\gamma)[G]$.
\item[\textup{(b)}] The Muon direction does not change, $U^{\mathrm{M}}_t[\gamma G] = U^{\mathrm{M}}_t[G]$, if $N_t = 0$ or if $\min(1, \gamma)\,\normF{N_t} \ge \varepsilon_{\mathrm{M}}$, where $N_t$ is the momentum of the history $G$. The second condition says that neither $\normF{N_t}$ nor $\normF{\gamma N_t}$ is below $\varepsilon_{\mathrm{M}}$.
\end{enumerate}
\end{lemma}
\begin{proof}
(a) The recursions in \eqref{eq:proof-adamw} are linear in $G$ and in $G^2$, so $m_t$ becomes $\gamma\,m_t$ and $v_t$ becomes $\gamma^2 v_t$. With $\hat m_t = m_t/(1-\beta_1^t)$ and $\hat v_t = v_t/(1-\beta_2^t)$,
\[
\frac{\gamma\,\hat m_t}{\gamma\sqrt{\hat v_t} + \varepsilon} = \frac{\hat m_t}{\sqrt{\hat v_t} + \varepsilon/\gamma}.
\]
(b) The recursions in \eqref{eq:proof-muon-mom} are linear, so $N_t$ becomes $\gamma N_t$. If $N_t = 0$, both directions are zero. Otherwise the condition says that the maximum in \eqref{eq:proof-muon-ns} is attained by the norm for $N_t$ and for $\gamma N_t$, so $X_0 = N_t/\normF{N_t}$ in both cases, and the iteration and \eqref{eq:proof-muon-dir} depend on $X_0$ only.
\end{proof}

\begin{proposition}[Full statement of \cref{prop:residual}]
\label{prop:proof-full}
Let $a > 0$. Consider two training runs on the same sequence of batches.
\begin{enumerate}[nosep]
\item[\textup{(R)}] \emph{Residual scaling}, the tuple $(a;\ W_0,\ \eta_t,\ \lambda)$. The layers in $\mathcal{B}$ have residual scale $a$. The output projections start at $W_0$ and follow \eqref{eq:proof-step} with learning rate $\eta_t$ and weight-decay coefficient $\lambda$.
\item[\textup{(P)}] \emph{The optimizer setting}, the tuple $(1;\ aW_0,\ a\,\eta_t,\ \lambda/a)$. The layers in $\mathcal{B}$ have residual scale $1$. The output projections start at $V_0 = aW_0$ and follow \eqref{eq:proof-step} with learning rate $a\,\eta_t$ and weight-decay coefficient $\lambda/a$.
\end{enumerate}
The other parameters start at the same $\omega_0$ and have the same optimizer and hyperparameters in both runs. Assume the following three conditions.
\begin{enumerate}[nosep]
\item[\textup{(A1)}] The gradients that reach the optimizers are the backpropagated gradients of $L_t$, without gradient-norm clipping.
\item[\textup{(A2)}] The output projections are updated either by Muon in both runs, with $N_t = 0$ or $\min(1, a)\,\normF{N_t} \ge \varepsilon_{\mathrm{M}}$ for every output projection of run \textup{(P)} and every $t$, or by AdamW with $\varepsilon$ in \textup{(R)} and $\varepsilon/a$ in \textup{(P)}.
\item[\textup{(A3)}] All optimizer states start at zero.
\end{enumerate}
Then, in exact arithmetic, $V_t = a\,W_t$ and $\omega^{\mathrm{P}}_t = \omega^{\mathrm{R}}_t$ for every $t \ge 0$. Consequently the two runs compute the same function after every update, have the same training loss at every update, and have the same loss on any evaluation set.
\end{proposition}
\begin{proof}
Write $G^{W}_r$ and $G^{V}_r$ for the gradients of the output projections at update $r$ in runs (R) and (P), and $G^{\omega,\mathrm{R}}_r$ and $G^{\omega,\mathrm{P}}_r$ for those of the other parameters. By induction on $t$ we show the statement $H_t$, namely that $V_t = aW_t$, that $\omega^{\mathrm{P}}_t = \omega^{\mathrm{R}}_t$, and that $G^{W}_r = a\,G^{V}_r$ and $G^{\omega,\mathrm{R}}_r = G^{\omega,\mathrm{P}}_r$ for every $r \le t$. $H_0$ holds by the choice of the initial values, and there are no gradients yet.

Assume $H_{t-1}$. The gradients of update $t$ are taken at $(W_{t-1}, \omega_{t-1})$ in (R) and at $(V_{t-1}, \omega_{t-1}) = (aW_{t-1}, \omega_{t-1})$ in (P). By \cref{lem:proof-grad}, $G^{W}_t = a\,G^{V}_t$ and $G^{\omega,\mathrm{R}}_t = G^{\omega,\mathrm{P}}_t$, and by (A1) these are the gradients that the optimizers receive.

\emph{Other parameters.} Each parameter in $\omega$ is updated by the same rule with the same hyperparameters, from the same value $\omega_{t-1}$, the same initial state by (A3) and the same gradient history in both runs. Hence $\omega^{\mathrm{P}}_t = \omega^{\mathrm{R}}_t$.

\emph{Output projections.} Let $U^{\mathrm{R}}_t$ and $U^{\mathrm{P}}_t$ be the directions of an output projection in the two runs. The gradient history of (R) is $a$ times that of (P), and the states start at zero by (A3), as \cref{lem:proof-scale} assumes. Under AdamW, \cref{lem:proof-scale}(a) with $\gamma = a$ gives $U^{\mathrm{R}}_t = U^{\mathrm{A}}_t(\varepsilon)[aG^{V}] = U^{\mathrm{A}}_t(\varepsilon/a)[G^{V}] = U^{\mathrm{P}}_t$. Under Muon, \cref{lem:proof-scale}(b) with $\gamma = a$ and (A2) give $U^{\mathrm{R}}_t = U^{\mathrm{P}}_t$. Multiplying the update \eqref{eq:proof-step} of run (R) by $a$,
\[
a\,W_t = (1 - \eta_t\lambda)\,aW_{t-1} - a\,\eta_t\,U^{\mathrm{R}}_t
= \Bigl(1 - (a\,\eta_t)\,\frac{\lambda}{a}\Bigr) V_{t-1} - (a\,\eta_t)\,U^{\mathrm{P}}_t = V_t ,
\]
where the last equality is the update \eqref{eq:proof-step} of run (P). This proves $H_t$. The consequences follow from \eqref{eq:proof-reparam}.
\end{proof}

Prior work equates a weight multiplier with rescaling that weight's initial value and learning rate~\citep{yang2022tp5,blake2024umup}.
\citet{yang2022tp5} state this symmetry for Adam without $\varepsilon$ or weight decay.
\Cref{prop:proof-full} extends it to AdamW with decoupled weight decay and to Muon, introducing $\lambda/a$ and condition (A2).
Weight decay is decoupled, so it acts through the product $\eta_t\lambda$.
Rescaling leaves this product unchanged, changing the weight-decay coefficient in (P) and not the decay per update.
A residual scale is therefore not additional weight decay.

The learning-rate schedule, weight sharing across visits and the remaining architecture play no role.
With a separate scale $a_\ell$ per layer, the same proof gives equivalence layer by layer.
Under (A1) to (A3), a looped MoE with $k$ visits and residual scale $1/k$ is thus equivalent to the same model without the scale, whose revisited output projections start at $W_0/k$ and train with learning rate $\eta_t/k$ and weight-decay coefficient $k\lambda$.
Under Muon, (A2) constrains the momentum along the run, so it can be checked only during training (\cref{sec:proof-breaks}).

\begin{corollary}[One AdamW update with the same $\varepsilon$]
\label{cor:proof-eps}
In the setting of \cref{prop:proof-full} with \textup{(A1)} and \textup{(A3)}, let both runs update the output projections by AdamW with the same $\varepsilon > 0$. Let the two runs be equivalent before update $t$. That is, $V_{t-1} = a\,W_{t-1}$ and $\omega^{\mathrm{P}}_{t-1} = \omega^{\mathrm{R}}_{t-1}$, and at every earlier update the gradient of each output projection in run \textup{(R)} was $a$ times its gradient in run \textup{(P)} and the gradients of the other parameters agreed. Then $\omega^{\mathrm{P}}_t = \omega^{\mathrm{R}}_t$, and every coordinate $i$ of every output projection satisfies
\[
\bigl|(V_t - aW_t)_i\bigr| \;\le\; \bigl|(V_t - (1-\eta_t\lambda)V_{t-1})_i\bigr| \cdot \min\Bigl(|1 - a|,\; \Bigl|\frac{1}{a} - 1\Bigr| \frac{\varepsilon}{\sqrt{\hat v_{t,i}}}\Bigr),
\]
where $\hat v_t$ is the bias-corrected second moment of run \textup{(P)}. The relative difference of the two updates vanishes as $\varepsilon \to 0$ in every coordinate with $\hat v_{t,i} > 0$, and a coordinate with $\hat v_{t,i} = 0$ is not moved by the direction in either run.
\end{corollary}
\begin{proof}
The hypothesis is the statement $H_{t-1}$ of the proof of \cref{prop:proof-full}. As there, the gradients of update $t$ satisfy $G^{W}_t = a\,G^{V}_t$ and agree for $\omega$, which gives $\omega^{\mathrm{P}}_t = \omega^{\mathrm{R}}_t$. Write $\hat m$ and $\hat v$ for the bias-corrected moments of one coordinate in run (P). By \cref{lem:proof-scale}(a) the update of that coordinate, in the variable $V$ and without the decay, is $-a\,\eta_t\,\hat m/(\sqrt{\hat v} + \varepsilon/a)$ in run (R) and $-a\,\eta_t\,\hat m/(\sqrt{\hat v} + \varepsilon)$ in run (P). The decay factors agree. The difference of the two is the update of run (P) times $\varepsilon\,(1/a - 1)/(\sqrt{\hat v} + \varepsilon/a)$, whose absolute value is at most $|1-a|$ and at most $|1/a - 1|\,\varepsilon/\sqrt{\hat v}$. If $\hat v = 0$ then $\hat m = 0$ and both updates are zero.
\end{proof}

For $a = \half$, the factor is $\min\bigl(\half,\, \varepsilon/\sqrt{\hat v_{t,i}}\bigr)$, so one update has a small relative difference in every coordinate with root-mean-square gradient large against $\varepsilon$.
The initial values satisfy the corollary's hypothesis at the first update.
Once the two runs differ, the hypothesis fails, so the corollary bounds one update from equivalent states, not the distance between two runs (\cref{sec:proof-breaks}).

\subsection{What breaks exactness}
\label{sec:proof-breaks}

The proof of \cref{prop:proof-full} uses each condition named here, and without it the two runs (R) and (P) can differ.

\term{Optimizer}
Under AdamW, residual scaling is exactly the optimizer setting with $\varepsilon/a$.
With the same $\varepsilon > 0$ in both runs, $a \neq 1$ and $\eta_t > 0$, one update from equivalent states already differs in every coordinate with nonzero first moment $\hat m_{t,i}$ (proof of \cref{cor:proof-eps}).
Under Muon with $a \neq 1$, the two runs start the Newton--Schulz iteration from different matrices if $0 < \min(1, a)\,\normF{N_t} < \varepsilon_{\mathrm{M}}$ for some output projection, with $N_t$ the momentum in (P).
With $\varepsilon_{\mathrm{M}} = 10^{-7}$, this requires nonzero momentum with Frobenius norm below $\varepsilon_{\mathrm{M}}/\min(1, a)$.
An expert never selected has $N_t = 0$, which the proposition covers.
An expert that stops being selected receives zero gradients, so its momentum $B_t$ shrinks by the factor $\mu$ each update and can enter this range.

\Cref{lem:proof-scale} uses (A3), zero initial optimizer states.
From nonzero states, such as a checkpoint of run (R) continued as run (P), its proof requires rescaling output-projection states like the gradients ($m$ and $B$ divided by $a$, $v$ divided by $a^2$).
The proof of the proposition also uses decoupled weight decay, which enters \eqref{eq:proof-step} only through the factor $1 - \eta_t\lambda$ common to both runs.

\term{Gradient clipping}
Our runs clip the global gradient norm $g_t$ at $\tau = \cfgClipNorm$, multiplying all gradients by $\min(1, \tau/g_t)$.
At equivalent states, \cref{lem:proof-grad} gives
\[
\bigl(g^{\mathrm{R}}_t\bigr)^2 = \lVert G^{\omega}_t \rVert^2 + a^2\,\lVert G^{V}_t \rVert^2,
\qquad
\bigl(g^{\mathrm{P}}_t\bigr)^2 = \lVert G^{\omega}_t \rVert^2 + \lVert G^{V}_t \rVert^2 ,
\]
where $G^{\omega}_t$ is the common gradient of $\omega$.
For $a \neq 1$ and $G^{V}_t \neq 0$, the two runs therefore clip by different factors whenever either clips, whether run (P) uses $\varepsilon$ or $\varepsilon/a$.
Neither optimizer is invariant to a factor that varies across updates.
The proposition holds for runs where clipping never activates.
In our runs, clipping activates in a small share of updates with or without residual scaling (\cref{tab:proof-clip}).

\begin{table}[htbp]
\caption{\lead{With residual scaling (upper rows), clipping activates in \clipPctScaleMinAdamW{} to \clipPctScaleMaxAdamW\% of updates under AdamW and \clipPctScaleMinMuon{} to \clipPctScaleMaxMuon\% under Muon. Without it (lower rows), the share is also small.}
Gradient-norm clipping counts and shares of the 4,769 updates in seed-42 main-setting runs (5B tokens).}
\label{tab:proof-clip}
\centering
\begin{fittabular}{5}{lNNNN}
\toprule
 & \multicolumn{2}{c}{AdamW} & \multicolumn{2}{c}{Muon}\\
\cmidrule(lr){2-3}\cmidrule(lr){4-5}
Run & Updates & Share \colunit{\%} & Updates & Share \colunit{\%}\\
\midrule
\MoEtw{} with residual scaling & \clipStepsMzeroRAdamW & \clipPctMzeroRAdamWTwoDec & \clipStepsMzeroRMuon & \clipPctMzeroRMuonTwoDec\\
Non-revisit control & \clipStepsMfourAdamW & \clipPctMfourAdamWTwoDec & \clipStepsMfourMuon & \clipPctMfourMuonTwoDec\\
Looped MoE & \clipStepsMthreeAdamW & \clipPctMthreeAdamWTwoDec & \clipStepsMthreeMuon & \clipPctMthreeMuonTwoDec\\
\grouprule
\MoEtw, learning-rate part & \clipStepsMzeroLRAdamW & \clipPctMzeroLRAdamWTwoDec & \clipStepsMzeroLRMuon & \clipPctMzeroLRMuonTwoDec\\
\MoEtw, initialization part & \clipStepsMzeroIAdamW & \clipPctMzeroIAdamWTwoDec & \clipStepsMzeroIMuon & \clipPctMzeroIMuonTwoDec\\
\MoEei{} with the recipe's expert layout & \clipStepsRAdamW & \clipPctRAdamWTwoDec & \clipStepsRMuon & \clipPctRMuonTwoDec\\
\bottomrule
\end{fittabular}
\end{table}

\term{Model and arithmetic}
Branch linearity fails if normalization sits between the output projection and the residual stream, or if a low-rank adapter is added to the projection without being rescaled.
For the proof to apply, any bias after the output projection must enter $W$ with the same rescaling of initial value, learning rate and weight-decay coefficient (and $\varepsilon/a$ in (P) where AdamW updates it).
Linearity also fails if a shared projection uses different scales across visits, because no single $V$ then represents it.
None of these occurs in the models of \cref{sec:matched,sec:residual,sec:depth} with residual scaling.
The per-visit residual split in \cref{tab:results-recipes} scales its two visits differently, and the proposition therefore does not apply to it.

When $a$ is a power of two, multiplying by $a$ is exact in binary floating point unless it overflows or underflows.
Under (A1) to (A3) with computed norms in (A2), $V_t = aW_t$ then holds bit for bit when both runs otherwise perform identical operations in the same order, including in the reduced-precision Newton--Schulz iteration.
For other $a$, the two runs round differently.
One update from equivalent states then agrees up to rounding while both select the same experts, and nothing here bounds the difference over a run.

\begin{table}[htbp]
\caption{\lead{Runs (R) and (P) agree bit for bit under the conditions of \cref{prop:proof-full}, and dropping a condition ends the agreement.}
Both checks train (R) and (P) with $a = \half$ for 30 updates.
The second runs without clipping except in its last two rows.}
\label{tab:proof-check}
\centering
\begin{fittabular}{3}{lll}
\toprule
Runs compared & Optimizer & Difference\\
\midrule
\multicolumn{3}{l}{\textit{First check, maximum logit difference after 30 updates (four layers, middle two scaled, width 32, 8 experts, top-2 routing)}}\\
$\varepsilon$ in (R), $\varepsilon/a$ in (P), no clipping & AdamW & none, bit for bit at all 30 updates\\
The same $\varepsilon = 10^{-8}$ in (R) and (P) & AdamW & $7.6 \times 10^{-7}$\\
Weight-decay coefficient left unscaled & AdamW & $7.4 \times 10^{-4}$\\
Initial values left unscaled & AdamW & $0.28$\\
$\varepsilon/a$ in (P), clipping at norm \cfgClipNorm & AdamW & none until the first update where only (P) clips\\
\grouprule
\multicolumn{3}{l}{\textit{Second check, training-loss difference over 30 updates (three layers, middle one scaled, width 64, 4 experts, top-2 routing)}}\\
(R) against (P) & Muon & none, bit for bit\\
The same $\varepsilon = 10^{-8}$ in (R) and (P) & AdamW & largest $3.5 \times 10^{-4}$\nats\\
Learning-rate part (with $\lambda/a$) or initialization part alone, against (R) & both & 0.103 to 0.194\nats\\
(R) against (P), clipping at norm \cfgClipNorm{} active at every update & AdamW & about $8 \times 10^{-3}$\nats\\
 & Muon & about $2 \times 10^{-2}$\nats\\
\bottomrule
\end{fittabular}
\end{table}

\term{Numerical check}
Two checks on small models train (R) and (P) with $a = \half$ for 30 updates (\cref{tab:proof-check}).
The first compares logits and tests the AdamW case of \cref{prop:proof-full} and the same-$\varepsilon$ setting of \cref{cor:proof-eps}, on CPU with our training code's model class, objective and optimizer.
The second compares training loss and tests Muon and the two parts of residual scaling, with optimizers and model built by our training code and 64-bit floating-point parameters.
Under the proposition's conditions, the runs of the first check agree bit for bit in 32-bit and 64-bit floating point across three AdamW implementations (the power-of-two case above).
With the same $\varepsilon$ in both runs, they differ from the first update, and after 30 updates the relative difference on the scaled output projections is $2.3 \times 10^{-4}$.
The corollary does not bound these measured distances.
Leaving the weight-decay coefficient or the initial values unscaled changes the logits, so the check can fail.
In the second check, Muon's bound was not reached.
We did not run the first check on GPU, in our runs' 16-bit mixed precision, or for $a$ not a power of two.

\subsection{The step size of one matrix under the two optimizers}
\label{sec:proof-stepsize}

We measure one update of a matrix $\Theta \in \mathbb{R}^{p \times q}$ by the spectral norm $\lVert U_t \rVert_2$ of its direction in \eqref{eq:proof-step}, that is, by the size of the update per unit learning rate.
Let $f(x) = c_1 x + c_2 x^3 + c_3 x^5$ be the Newton--Schulz polynomial of \eqref{eq:proof-muon-ns} with the coefficients of our runs.
On $(0, \infty)$ its derivative vanishes at $x_a \approx 0.5545$ and $x_b \approx 1.0501$, and we write
\[
M_* = f(x_a) \approx 1.2024, \qquad m_* = f(x_b) \approx 0.6818, \qquad
C = \frac{1-\beta_1}{\sqrt{(1-\beta_2)\,(1-\beta_1^2/\beta_2)}} \approx 1.165 ,
\]
where the value of $C$ is for our $\beta_1 = 0.9$ and $\beta_2 = 0.95$.

\begin{lemma}[Size of one update]
\label{lem:proof-step}
Let $\Theta \in \mathbb{R}^{p \times q}$ and let $U_t$ be its direction at update $t$.
\begin{enumerate}[nosep]
\item[\textup{(a)}] \emph{Muon.} Let $\normF{N_t} \ge \varepsilon_{\mathrm{M}}$, $J = 5$ and $\min(p, q) \le 2.7 \times 10^5$. Then, in exact arithmetic,
\[
m_*\,\rho \;\le\; \lVert U^{\mathrm{M}}_t \rVert_2 \;\le\; M_*\,\rho, \qquad \rho = 0.2\sqrt{\max(p, q)} ,
\]
whatever the gradients $G_1, \dots, G_t$.
\item[\textup{(b)}] \emph{AdamW, any gradient.} Let $\beta_1^2 < \beta_2$, as in our settings. Every entry satisfies $\lvert (U^{\mathrm{A}}_t)_{ij} \rvert \le C$, so $\lVert U^{\mathrm{A}}_t \rVert_2 \le C\sqrt{pq}$. If $G_r = \xi_r\,u v^{\top}$ for $r \le t$, with $\xi_r > 0$ and no zero entry in $u$ or $v$, then $\lVert U^{\mathrm{A}}_t \rVert_2 \to \kappa_t \sqrt{pq}$ as $\varepsilon \to 0$, for some $\kappa_t \in (0, C]$, and $\kappa_t = 1$ when $\xi_r$ is constant.
\item[\textup{(c)}] \emph{AdamW, incoherent gradient.} Let the histories $(G_{1,ij}, \dots, G_{t,ij})$ of the entries be independent and identically distributed across $(i, j)$, and let each have the same law as its negative. Let $\varsigma^2 = \mathbb{E}\,(U^{\mathrm{A}}_t)_{ij}^2 > 0$ and $n = \max(p, q)$. Then, with a universal constant $K$,
\[
\frac{\varsigma\sqrt{n}}{\sqrt{1 + C^2/(\varsigma^2 n)}} \;\le\; \mathbb{E}\,\lVert U^{\mathrm{A}}_t \rVert_2 \;\le\; K\sqrt{C \varsigma}\,\bigl(\sqrt{p} + \sqrt{q}\bigr).
\]
If the law of a history does not change with $p$ and $q$, then $\lVert U^{\mathrm{A}}_t \rVert_2 / \bigl(\varsigma(\sqrt{p} + \sqrt{q})\bigr) \to 1$ almost surely as $p, q \to \infty$ with $p/q$ fixed.
\end{enumerate}
\end{lemma}

\begin{proof}
(a) By \eqref{eq:proof-muon-ns} and $\normF{N_t} \ge \varepsilon_{\mathrm{M}}$, $X_0 = N_t/\normF{N_t}$. Its nonzero singular values $\sigma_1 \ge \sigma_2 \ge \dots$ lie in $(0, 1]$, and their squares sum to 1, so $\sigma_1 \ge 1/\sqrt{\min(p, q)}$. If $X = Q_1 \Sigma Q_2^{\top}$ is a singular value decomposition, then $(XX^{\top})X = Q_1\Sigma^3 Q_2^{\top}$, so one iteration maps $X$ to $Q_1 f(\Sigma) Q_2^{\top}$, and $X_J = Q_1 f^J(\Sigma) Q_2^{\top}$. Running the iteration on the transpose changes nothing. The quadratic $c_1 + c_2 y + c_3 y^2$ in $y = x^2$ has no real root, because $c_2^2 < 4c_1c_3$, so $f(x) = x(c_1 + c_2x^2 + c_3x^4) > 0$ on $(0, \infty)$ and $\lVert X_J \rVert_2 = \max_i f^J(\sigma_i)$.

The function $f$ increases on $[0, x_a]$, decreases on $[x_a, x_b]$ and increases after $x_b$. Two intervals are invariant. On $[0, M_*]$, this monotonicity gives $f \le \max(f(x_a), f(M_*)) = M_*$, because $f(x_a) = M_*$ and $f(M_*) \approx 0.9465$. On $[m_*, M_*]$, $f \ge m_*$, because $m_* > x_a$ and $f$ attains its minimum over $[x_a, \infty)$ at $x_b$. Every $\sigma_i \le 1 < M_*$, so $\lVert X_J \rVert_2 \le M_*$.

For the lower bound, let $x_* \approx 0.2107$ be the point of $[0, x_a]$ with $f(x_*) = m_*$. On $[0, x_*]$ the ratio $f(x)/x = c_1 + c_2x^2 + c_3x^4$ decreases, because $x^2 < -c_2/(2c_3)$, so $f(x) \ge \alpha x$ with $\alpha = m_*/x_* \approx 3.237$. On $[x_*, M_*]$, $f \ge m_*$. Hence the iterates $x_j = f^j(\sigma_1)$ grow at least by the factor $\alpha$ while they are below $x_*$, and from the first iterate at or above $x_*$ on, every later iterate lies in $[m_*, M_*]$. Therefore $x_J \ge m_*$ whenever $\alpha^{J-1}\sigma_1 \ge x_*$. For $J = 5$ this holds if $\sigma_1 \ge 0.00192$, which $\sigma_1 \ge 1/\sqrt{\min(p,q)}$ guarantees for $\min(p, q) \le 2.7 \times 10^5$. Thus $\lVert X_J \rVert_2 \ge f^J(\sigma_1) \ge m_*$, and \eqref{eq:proof-muon-dir} multiplies both bounds by $\rho$.

(b) Fix an entry with gradients $g_1, \dots, g_t$. By \eqref{eq:proof-adamw}, $m_t = (1-\beta_1)\sum_{r \le t} \beta_1^{t-r} g_r$ and $v_t = (1-\beta_2)\sum_{r \le t} \beta_2^{t-r} g_r^2$. If $v_t = 0$, every $g_r$ is zero and so is the entry of $U_t$. Otherwise, write $\beta_1^{t-r} = b^{(t-r)/2}\,\beta_2^{(t-r)/2}$ with $b = \beta_1^2/\beta_2 < 1$. The Cauchy--Schwarz inequality gives
\[
\lvert m_t \rvert \le (1-\beta_1)\sqrt{\frac{1-b^t}{1-b}}\,\sqrt{\frac{v_t}{1-\beta_2}},
\qquad\text{so}\qquad
\lvert (U^{\mathrm{A}}_t)_{ij} \rvert \le \frac{\lvert m_t \rvert\,\sqrt{1-\beta_2^t}}{(1-\beta_1^t)\sqrt{v_t}} \le C\,\frac{\sqrt{(1-\beta_2^t)(1-b^t)}}{1-\beta_1^t}.
\]
The product of $\beta_2^t$ and $b^t$ is $\beta_1^{2t}$, so their sum is at least $2\beta_1^t$, and $(1-\beta_2^t)(1-b^t) \le (1-\beta_1^t)^2$. Hence every entry is at most $C$ in absolute value, and $\lVert U \rVert_2 \le \normF{U} \le C\sqrt{pq}$. For $G_r = \xi_r\,uv^{\top}$ the entry $(i,j)$ has gradients $u_iv_j\xi_r$. Its first moment is $u_iv_j$ times that of $(\xi_r)$, and its second moment is $u_i^2v_j^2$ times that of $(\xi_r)$. As $\varepsilon \to 0$, the entry therefore tends to $\operatorname{sign}(u_iv_j)\,\kappa_t$, where $\kappa_t$ is the ratio for the scalar history $(\xi_r)$, which is positive and at most $C$. The limit is $\kappa_t\operatorname{sign}(u)\operatorname{sign}(v)^{\top}$, with spectral norm $\kappa_t\sqrt{pq}$. A constant $\xi_r = \xi$ has bias-corrected moments $\xi$ and $\xi^2$, so $\kappa_t = 1$.

(c) Each entry of $U^{\mathrm{A}}_t$ is the same function of its own history. The function is odd, because $m_t$ is odd and $v_t$ is even in the history. The entries are therefore independent, identically distributed, centered and bounded by $C$, with variance $\varsigma^2$, and $\mathbb{E}(U^{\mathrm{A}}_t)_{ij}^4 \le C^2 \varsigma^2$. By the theorem of \citet{latala2005}, $\mathbb{E}\lVert U \rVert_2$ is at most a universal constant times $\varsigma\sqrt{q} + \varsigma\sqrt{p} + (pq\,C^2\varsigma^2)^{1/4}$. Since $(pq)^{1/4} \le (\sqrt{p} + \sqrt{q})/2$ and $\varsigma \le \sqrt{C\varsigma}$, this gives the upper bound. For the lower bound, $\lVert U \rVert_2$ is at least the norm of one column or one row. Let $Y$ be the squared norm of a column or row of length $n$. Then $\mathbb{E}Y = n\varsigma^2$ and $\mathbb{E}Y^2 \le n^2\varsigma^4 + nC^2\varsigma^2$. Hölder's inequality gives $\mathbb{E}Y \le (\mathbb{E}\sqrt{Y})^{2/3}(\mathbb{E}Y^2)^{1/3}$, so $\mathbb{E}\sqrt{Y} \ge (\mathbb{E}Y)^{3/2}/(\mathbb{E}Y^2)^{1/2}$, which is the lower bound. The limit is the theorem of \citet{yin1988} on the largest singular value of a matrix with independent, identically distributed, centered entries of finite fourth moment.
\end{proof}

\begin{figure}[tbp]
\fig{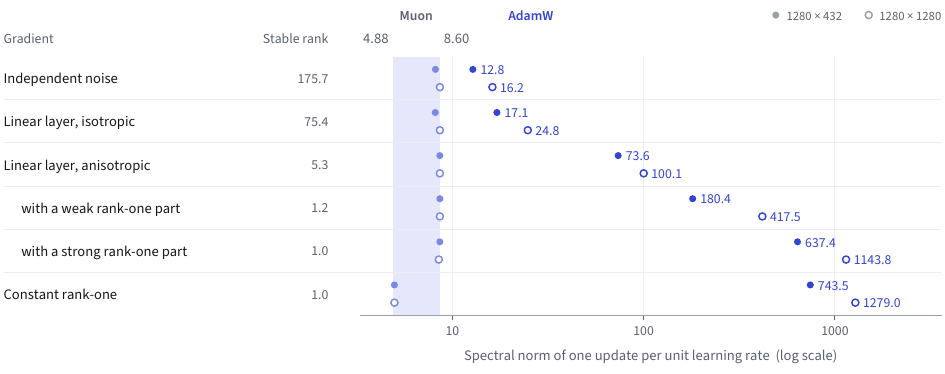}{Dot chart on one logarithmic axis, the spectral norm of one update per unit learning rate after 300 updates. Six rows, one per kind of gradient, each with the stable rank of the Nesterov momentum for the 1280 by 432 matrix and with two lines of marks, filled for the 1280 by 432 matrix above and hollow for the 1280 by 1280 matrix below. In every pair of values the first is for the 1280 by 432 matrix and the second for the 1280 by 1280 matrix. Independent noise, stable rank 175.7, AdamW 12.8 and 16.2, Muon 8.16 and 8.61. Linear layer, isotropic, stable rank 75.4, AdamW 17.1 and 24.8, Muon 8.14 and 8.61. Linear layer, anisotropic, stable rank 5.3, AdamW 73.6 and 100.1, Muon 8.60 and 8.61. Anisotropic with a weak rank-one part, stable rank 1.2, AdamW 180.4 and 417.5, Muon 8.61 and 8.61. Anisotropic with a strong rank-one part, stable rank 1.0, AdamW 637.4 and 1143.8, Muon 8.60 and 8.50. Constant rank-one, stable rank 1.0, AdamW 743.5 and 1279.0, Muon 4.98 and 4.98. A shaded band from 4.88 to 8.60 marks the interval of the lemma under Muon for both shapes, and every Muon mark lies in it or at its upper end. The AdamW marks carry their values and move to the right from row to row.}
\caption{\lead{Under Muon, the size of one update stays in a narrow interval fixed by matrix shape. Under AdamW, it spans a factor of 58 to 79 across gradients.}
Spectral norm of update direction $U_t$ after 300 updates for both output-projection shapes in \MoEtw{}.
The band shows the interval of \cref{lem:proof-step}(a).
Rows are ordered by stable rank of Nesterov momentum $N_t$ (for $1280 \times 432$).}
\label{fig:proof-step}
\end{figure}

Under the conditions of (a), in particular $\normF{N_t} \ge \varepsilon_{\mathrm{M}}$, Muon therefore gives every matrix of one shape and learning rate a step whose spectral norm lies within the factor $M_*/m_* \approx 1.76$, whatever its gradients.
Under AdamW, (b) gives the upper bound $C\sqrt{pq}$ for any gradient.
No lower bound growing with shape holds for all gradients, because gradients nonzero in only one fixed entry give $\lVert U^{\mathrm{A}}_t \rVert_2 \le C$.
A constant rank-one gradient without zero entries gives $\sqrt{pq}$ as $\varepsilon \to 0$, and incoherent gradients in (c) give $\varsigma(\sqrt{p} + \sqrt{q})$ in the large-matrix limit.
At fixed $\varsigma$, their ratio grows like $\min(\sqrt{p}, \sqrt{q})$.
For Gaussian gradients independent across entries and updates, $\varsigma \approx 0.226$ numerically at large $t$, so the incoherent limit is about $0.23(\sqrt{p} + \sqrt{q})$.
The lemma concerns step size alone, not which size is best (\cref{sec:residual} addresses this by experiment).
It also does not say which matrices of our models receive coherent gradients (\cref{sec:results-stepsize} measures the step in training).

\term{Numerical check}
Output projections in \MoEtw{} have two shapes, expert $W_{2,j}$ ($1280 \times 432$) and attention $W_{\mathrm{att}}$ ($1280 \times 1280$), both with $\rho \approx 7.155$.
We computed $\lVert U_t \rVert_2$ for both shapes after 300 updates under our optimizer settings across six gradient types (\cref{fig:proof-step}).
These include independent Gaussian noise and linear-layer gradients (mean of $\delta z^{\top}$ over 64 tokens with Gaussian $z$ and $\delta$, isotropic or with covariance eigenvalues inversely proportional to index).
The remaining cases add a weak or strong fixed rank-one matrix to the anisotropic gradient, or use that rank-one matrix alone.
Under Muon, five types give 8.14 to 8.61 and the constant rank-one gradient gives $f^5(1)\,\rho \approx 4.98$.
All lie in $[m_*\rho, M_*\rho] = [4.88, 8.60]$, except that the 16-bit rounding of the iteration, which the check imitates, puts six of the twelve values up to about $0.1\%$ above the upper end.
Under AdamW, values span a factor of 58 for $W_{2,j}$ and 79 for $W_{\mathrm{att}}$, staying below $C\sqrt{pq}$ from (b) (866 and 1491).
Independent noise gives $0.226(\sqrt{p} + \sqrt{q})$ and constant rank-one gradients give $\sqrt{pq}$, matching the lemma.

\subsection{The step size measured in training}
\label{sec:results-stepsize}

We measure $\lVert U_t \rVert_2$ of \cref{sec:proof-stepsize} on the output projections of \MoEtw{} during training (\cref{tab:results-stepsize}).
From the saved state of a seed-42 run after 477, 954, 1,908 and 3,339 of its 4,769 updates, we replay the optimizer step on eight consecutive training batches, remove the weight-decay term and divide by the learning rate.
A replayed Muon update equals the training update up to 16-bit nondeterminism.

\begin{table}[!ht]
\caption{\lead{Under Muon, every output projection takes nearly the same step per unit learning rate. Under AdamW, the step varies across matrices and between the two types.} Spectral norm $\lVert U_t \rVert_2$ of one training update per unit learning rate in \MoEtw{} (seed 42), mean over eight batches. Median and P90/P10 (90th over 10th percentile) are over the matrices of a type. Middle is the median over the middle six layers divided by the median over all layers.}
\label{tab:results-stepsize}
\centering
\begin{fittabular}{7}{lNNNNNN}
\toprule
 & \multicolumn{3}{c}{AdamW} & \multicolumn{3}{c}{Muon}\\
\cmidrule(lr){2-4}\cmidrule(lr){5-7}
After update & Median & P90/P10 & Middle & Median & P90/P10 & Middle\\
\midrule
\multicolumn{7}{l}{\textit{Attention output projections (12 matrices, $1280 \times 1280$)}}\\
\quad 477 & 95.0 & 1.56 & 1.01 & 8.61 & 1.00 & 1.00\\
\quad 954 & 54.7 & 1.20 & 0.99 & 8.61 & 1.00 & 1.00\\
\quad 1,908 & 56.7 & 1.26 & 1.02 & 8.61 & 1.00 & 1.00\\
\quad 3,339 & 56.4 & 1.25 & 0.94 & 8.61 & 1.00 & 1.00\\
\grouprule
\multicolumn{7}{l}{\textit{Expert output projections (768 matrices, $1280 \times 432$)}}\\
\quad 477 & 35.9 & 2.03 & 0.94 & 8.56 & 1.06 & 0.97\\
\quad 954 & 31.6 & 1.64 & 0.96 & 8.14 & 1.06 & 1.00\\
\quad 1,908 & 30.0 & 1.65 & 0.96 & 8.14 & 1.06 & 1.00\\
\quad 3,339 & 29.1 & 1.69 & 0.95 & 8.14 & 1.05 & 1.00\\
\bottomrule
\end{fittabular}
\end{table}

Under Muon, the median step is 8.61 for attention and 8.14 to 8.56 for expert output projections, and the 90th percentile is at most 1.06 times the 10th.
The largest value over all matrices that Muon updates is 8.63, against the upper end 8.60 of \cref{lem:proof-step}(a) in exact arithmetic.
Under AdamW, the median step is 54.7 to 95.0 for attention and 29.1 to 35.9 for expert output projections, and the 90th percentile is 1.20 to 2.03 times the 10th.
The step of the middle six layers is 0.94 to 1.02 times the median over all layers under AdamW and 0.97 to 1.00 under Muon, so these layers are not singled out.
With residual scaling on the middle six layers under AdamW, the median step of their expert output projections stays within 5\% of the step without it at all four checkpoints (33.9 against 33.8 after update 477, 28.0 against 27.8 after update 3,339).
This agrees with \cref{lem:proof-scale}(a), by which rescaling the gradients leaves AdamW's direction unchanged up to $\varepsilon$.

\section{Further observations}
\label{sec:observations-app}

Unless stated, models store 1.3B parameters and train on 5B tokens with seed 42 at calibrated learning rates.
Differences are in test mBPB.
A one-seed difference is within noise below two standard errors, which are \seedTwoSeOneOneAdamW\mbpb{} under AdamW and \seedTwoSeOneOneMuon{} under Muon (\cref{sec:looptitan-setup}).

\subsection{Revisit taken apart}
\label{sec:obs-parts}

The revisit step shares experts, router and attention with norms between the two visits.
At 5B tokens, its three parts nearly cancel.
Two models between the \Ctrl{} and the looped MoE add sharing step by step.
With \emph{experts revisited}, visits keep separate attention, routers and norms.
With \emph{experts and router revisited}, visits keep only separate attention and norms.
Both compute the same 18 layers at matched budgets (\cref{tab:setup-app-budgets}).

\term{At 5B tokens}
Under AdamW, revisiting the experts alone costs \gapMsevenMfourAdamWAbsTwoDec\mbpb{}, adding router revisit gains \gapMfiveMsevenAdamWAbsTwoDec{}, and adding attention and norm revisit costs \gapMthreeMfiveAdamWAbsTwoDec{} (\cref{tab:obs-parts}).
Against the threshold of \seedTwoSeOneOneAdamWTwoDec, the router part is beyond noise and the other two are within noise.
Under Muon, each part is within noise, led by the router part (\gapMfiveMsevenMuonAbsTwoDec{} against the threshold of \seedTwoSeOneOneMuonTwoDec).
Under each optimizer, they sum to the seed-42 revisit step (three seeds in \cref{tab:settings}).
The model with experts and router revisited matches the \Ctrl{} under both optimizers.
Under Muon, this model has \gapMfiveUMuonAbsTwoDec\mbpb{} lower loss than \MoEei{} (\Base{}), also within noise.

\begin{table}[t]
\caption{\lead{At 5B tokens, the three parts of revisit nearly cancel under both optimizers.}
Later minus earlier model in test mBPB (negative is a gain), seed 42.
Raw values are in \cref{tab:results-5b}.}
\label{tab:obs-parts}
\centering
\begin{fittabular}{3}{lNN}
\toprule
Step & AdamW & Muon\\
\midrule
Revisit the experts (\Ctrl{} to experts revisited) & \gapMsevenMfourAdamWTwoDec & \gapMsevenMfourMuonTwoDec\\
Revisit the router (to experts and router revisited) & \gapMfiveMsevenAdamWTwoDec & \gapMfiveMsevenMuonTwoDec\\
Revisit attention and norms (to the looped MoE) & \gapMthreeMfiveAdamWTwoDec & \gapMthreeMfiveMuonTwoDec\\
\oursrow Sum (looped MoE $-$ \Ctrl) & \gapMthreeMfourAdamWTwoDecSFortyTwo & \gapMthreeMfourMuonTwoDecSFortyTwo\\
\grouprule
\multicolumn{3}{l}{\textit{Experts and router revisited, against two models}}\\
Minus the \Ctrl & \gapMfiveMfourAdamWTwoDec & \gapMfiveMfourMuonTwoDec\\
Minus \MoEei{} (\Base{} under Muon) & \gapMfiveUAdamWTwoDec & \gapMfiveUMuonTwoDec\\
\bottomrule
\end{fittabular}
\end{table}

\begin{figure}[t]
\fig{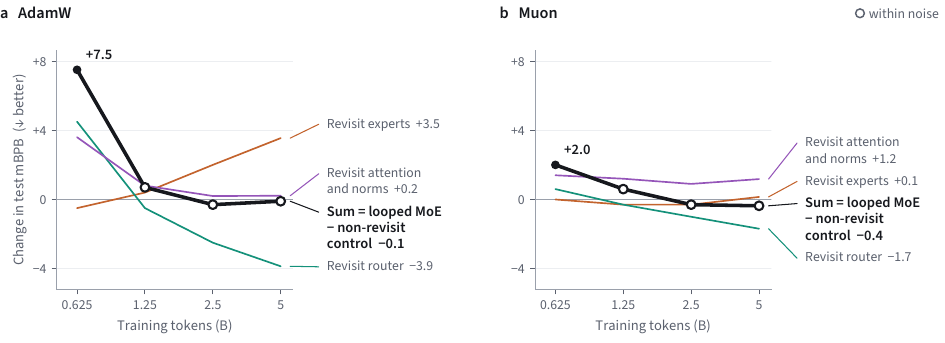}{Two panels, AdamW on the left and Muon on the right, with a shared vertical axis in test mBPB and training tokens of 0.625B, 1.25B, 2.5B and 5B on the horizontal axis. Each panel has three thin lines for the three parts of the revisit step and one thick line with dots for their sum (the looped MoE minus the non-revisit control). A filled dot of the sum is beyond noise and a hollow dot within noise, two standard errors of a one-seed difference, \seedTwoSeOneOneAdamW{} in the AdamW panel and \seedTwoSeOneOneMuon{} in the Muon panel. In both panels the sum is beyond noise at 0.625B tokens and within noise from 1.25B on. AdamW. Revisit the experts, minus \hstepExpertsAdamWHzZeroPSixTwoFiveAbs, plus \hstepExpertsAdamWHzOnePTwoFiveAbs, plus \hstepExpertsAdamWHzTwoPFiveAbs, plus \gapMsevenMfourAdamWAbs. Revisit the router, plus \hstepRouterAdamWHzZeroPSixTwoFiveAbs, minus \hstepRouterAdamWHzOnePTwoFiveAbs, minus \hstepRouterAdamWHzTwoPFiveAbs, minus \gapMfiveMsevenAdamWAbs. Revisit attention and norms, plus \hstepAttnAdamWHzZeroPSixTwoFiveAbs, plus \hstepAttnAdamWHzOnePTwoFiveAbs, plus \hstepAttnAdamWHzTwoPFiveAbs, plus \gapMthreeMfiveAdamWAbs. Sum, plus \hstepAllShareAdamWHzZeroPSixTwoFiveAbs, plus \hstepAllShareAdamWHzOnePTwoFiveAbs, minus \hstepAllShareAdamWHzTwoPFiveAbs, minus \gapMthreeMfourAdamWAbsSFortyTwo. Muon. Revisit the experts, \hstepExpertsMuonHzZeroPSixTwoFiveAbs, minus \hstepExpertsMuonHzOnePTwoFiveAbs, minus \hstepExpertsMuonHzTwoPFiveAbs, plus \gapMsevenMfourMuonAbs. Revisit the router, plus \hstepRouterMuonHzZeroPSixTwoFiveAbs, minus \hstepRouterMuonHzOnePTwoFiveAbs, minus \hstepRouterMuonHzTwoPFiveAbs, minus \gapMfiveMsevenMuonAbs. Revisit attention and norms, plus \hstepAttnMuonHzZeroPSixTwoFiveAbs, plus \hstepAttnMuonHzOnePTwoFiveAbs, plus \hstepAttnMuonHzTwoPFiveAbs, plus \gapMthreeMfiveMuonAbs. Sum, plus \hstepAllShareMuonHzZeroPSixTwoFiveAbs, plus \hstepAllShareMuonHzOnePTwoFiveAbs, minus \hstepAllShareMuonHzTwoPFiveAbs, minus \gapMthreeMfourMuonAbsSFortyTwo.}
\caption{\lead{At 0.625B tokens, revisit costs \hstepAllShareAdamWHzZeroPSixTwoFiveAbs\mbpb{} under AdamW and \hstepAllShareMuonHzZeroPSixTwoFiveAbs{} under Muon. From 1.25B on, it is at most \hstepAllShareAdamWHzOnePTwoFiveAbs{} from zero.}
Thin lines show the three parts and the thick line their sum (looped MoE minus \Ctrl{}), in test mBPB with seed 42 (log token axis).
Filled dots of the sum are beyond noise and hollow dots within noise (two standard errors of a one-seed difference, \seedTwoSeOneOneAdamW\mbpb{} under AdamW and \seedTwoSeOneOneMuon{} under Muon, \cref{sec:results-seeds}).
Three-seed 5B sums are in \cref{tab:settings}.}
\label{fig:obs-forks}
\end{figure}

\term{During training}
Runs of 0.625B, 1.25B and 2.5B tokens branch from the 5B runs' constant phase and decay the learning rate to zero over their last 20\% (\cref{fig:obs-forks}).
At 0.625B tokens, the looped MoE's loss exceeds the \Ctrl{}'s by \hstepAllShareAdamWHzZeroPSixTwoFiveAbs\mbpb{} under AdamW and \hstepAllShareMuonHzZeroPSixTwoFiveAbs{} under Muon (against the threshold of \seedTwoSeOneOneMuon), both beyond noise.
From 1.25B tokens on, the sum is at most \hstepAllShareAdamWHzOnePTwoFiveAbs\mbpb{} from zero under both optimizers.
Under AdamW, the router part is \hstepRouterAdamWHzZeroPSixTwoFive\mbpb{} at 0.625B tokens against \gapMfiveMsevenAdamWTwoDec{} at 5B, and the expert part \hstepExpertsAdamWHzZeroPSixTwoFive{} against \gapMsevenMfourAdamWTwoDec.
Under Muon, no part exceeds \gapMfiveMsevenMuonAbsTwoDec\mbpb{} at any length.
At 0.625B tokens under Muon, the looped MoE's loss exceeds \MoEtw{}'s by \hgapMthreeMzeroMuonHzZeroPSixTwoFiveAbs\mbpb{}, with \hstepShapeMuonHzZeroPSixTwoFiveAbs{} from the four steps to the \Ctrl{} and \hstepAllShareMuonHzZeroPSixTwoFiveAbs{} from revisit.

\subsection{The second visit at inference}
\label{sec:obs-readout}

The trained looped MoE uses its second visit.
A fixed-visit model is best at its trained count.

\begin{figure}[t]
\fig{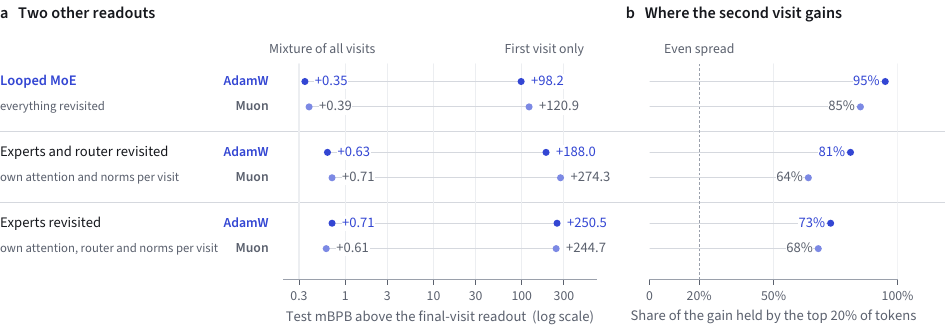}{(a) For three models under AdamW and Muon, the loss of a mixture-of-visits readout (0.35 to 0.71 mBPB) and of a first-visit readout (98 to 274 mBPB) relative to the final-visit readout, on a log axis. (b) The fraction of the second visit's gain held by the top 20 percent of tokens, between 64 and 95 percent.}
\caption{\lead{Reading out after the first visit costs \roFirstMinusNativeMinOneDec{} to \roFirstMinusNativeMaxOneDec\mbpb{}, and mixing visits does not beat the final state.}
\textbf{a}~Test mBPB above the final-visit readout (log axis) for a mixture of all visits and for the first visit alone, by model and optimizer.
The mixture is compared with a fitted readout of the final state alone.
\textbf{b}~Share of the second visit's gain in the top 20\% of tokens, against an even spread (dashed).}
\label{fig:obs-readout}
\end{figure}

\term{Visit readouts}
We freeze the trained looped MoE and the models with experts revisited and with experts and router revisited (\cref{sec:obs-parts}), and compare test-split readouts (\cref{fig:obs-readout}).
Reading out after the first visit (straight to the last three layers and output head) costs the looped MoE \roFirstMinusNativeMthreeAdamW\mbpb{} under AdamW and \roFirstMinusNativeMthreeMuon{} under Muon.
It costs the other two models, whose visits keep separate attention, \roFirstMinusNativeMfiveAdamW{} to \roFirstMinusNativeMfiveMuon\mbpb{}.
Therefore, the second visit carries a large part of the computation, and more when attention is not revisited.
Mixing the states after the first three layers and each visit (three scalar weights fitted on validation sequences) puts weight \roWeightTwoMin{} to \roWeightTwoMax{} on the final state.
This mixture is \roControlMinusMixtureMin{} to \roControlMinusMixtureMax\mbpb{} worse than a comparison readout that uses the final state alone and is fitted the same way.
The 20\% of tokens that gain most hold \roTopShareMin{} to \roTopShareMax\% of the second visit's gain.
The second visit's residual update has \roURatioMin{} to \roURatioMax{} times the norm of the first's.

\begin{figure}[t]
\fig{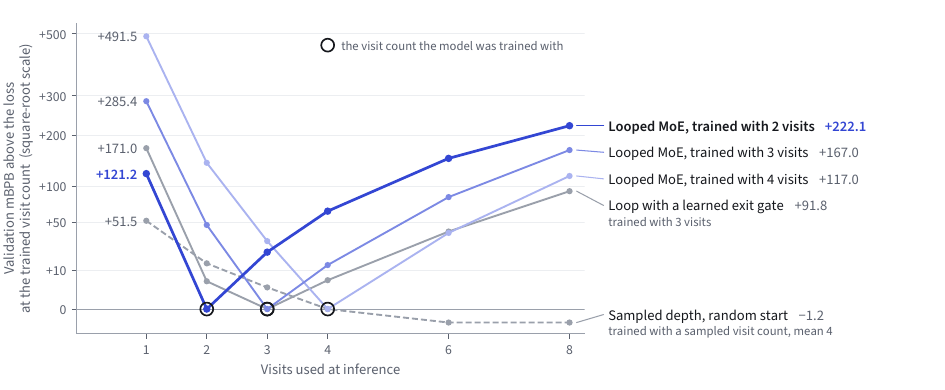}{Line chart. Horizontal axis, visits at inference, 1, 2, 3, 4, 6 and 8. Vertical axis, validation mBPB above the same model at its trained visit count, on a square-root scale. One line per model, all trained with Muon, with a ring at the trained count of each fixed-visit model. Looped MoE trained with two visits, plus \depthDeltaMthreeMuonVisOneAbs{} at one visit, zero at two, plus \depthDeltaMthreeMuonVisThreeAbs{} at three, plus \depthDeltaMthreeMuonVisFourAbs{} at four, plus \depthDeltaMthreeMuonVisSixAbs{} at six, plus \depthDeltaMthreeMuonVisEightAbs{} at eight. Looped MoE trained with three visits, plus \depthDeltaSmeltVthreeMuonVisOneAbs, plus \depthDeltaSmeltVthreeMuonVisTwoAbs, zero, plus \depthDeltaSmeltVthreeMuonVisFourAbs, plus \depthDeltaSmeltVthreeMuonVisSixAbs, plus \depthDeltaSmeltVthreeMuonVisEightAbs. Looped MoE trained with four visits, plus \depthDeltaSmeltVfourMuonVisOneAbs, plus \depthDeltaSmeltVfourMuonVisTwoAbs, plus \depthDeltaSmeltVfourMuonVisThreeAbs, zero, plus \depthDeltaSmeltVfourMuonVisSixAbs, plus \depthDeltaSmeltVfourMuonVisEightAbs. Loop with a learned exit gate, trained with three visits, plus \depthDeltaOuroMuonVisOneAbs, plus \depthDeltaOuroMuonVisTwoAbs, zero, plus \depthDeltaOuroMuonVisFourAbs, plus \depthDeltaOuroMuonVisSixAbs, plus \depthDeltaOuroMuonVisEightAbs. Sampled depth from a random start, referenced to four visits, plus \depthDeltaHuginnMuonVisOneAbs, plus \depthDeltaHuginnMuonVisTwoAbs, plus \depthDeltaHuginnMuonVisThreeAbs, zero, minus \depthDeltaHuginnMuonVisSixAbs, minus \depthDeltaHuginnMuonVisEightAbs. The four fixed-visit models each form a V with its low point at the trained count, and the sampled-depth line is flat from three visits on.}
\caption{\lead{Fixed-visit models are best at their trained visit count and degrade on both sides.}
Validation mBPB at each inference visit count minus loss at the trained count (ring), on a square-root axis.
All models train with Muon.
Gray lines show our adaptations (\cref{sec:results-recipes}) of two designs, a loop with a learned exit gate~\citep{ouro} and sampled depth from a random start~\citep{huginn}.
The second is referenced to four visits.}
\label{fig:obs-visits}
\end{figure}

\term{Other visit counts}
Trained with three visits at matched budgets and the same 18 computed layers, the looped MoE matches its two-visit form (\gapSmeltVthreeMthreeAdamW\mbpb{} under AdamW, \gapSmeltVthreeMthreeMuon{} under Muon).
With four trained visits, loss rises by \gapSmeltVfourMthreeAdamWAbs\mbpb{} under AdamW (within noise) and \gapSmeltVfourMthreeMuonAbs{} under Muon (beyond noise, \cref{tab:results-recipes}).
At inference, we run five Muon-trained models at 1, 2, 3, 4, 6 and 8 visits and report loss on the full validation split (\cref{fig:obs-visits}).
Every fixed-visit model is best at its trained count and degrades on both sides.
The two-visit looped MoE is \depthDeltaMthreeMuonVisOneAbs\mbpb{} worse at one visit and \depthDeltaMthreeMuonVisThreeAbs{} worse at three.
Our adaptation of sampled depth from a random start~\citep{huginn} instead trains with a sampled visit count.
From three visits on, it stays within \depthDeltaHuginnMuonVisThreeAbs\mbpb{} of its four-visit loss, but its best loss exceeds the two-visit looped MoE's by \depthHuginnBestMinusMthreeAbs\mbpb{}.
In these runs, adjustable inference depth is not a by-product of revisit.

\subsection{Gradients and curvature}
\label{sec:obs-curvature}

We measure gradients and curvature in five models (\MoEtw{} and the four 18-layer models from the \Ctrl{} to the looped MoE) under both optimizers at five checkpoints, each on 64 validation sequences of 4096 tokens.

\term{Gradients of the two visits}
The two visits' gradients on revisited parameters point the same way.
On revisited experts, their cosine is positive at every checkpoint and falls during training, from \cosExpertsMthreeAdamWStFourSevenSeven{} to \cosExpertsMthreeAdamWStFourSevenSixNine{} in the looped MoE under AdamW and \cosExpertsMthreeMuonStFourSevenSeven{} to \cosExpertsMthreeMuonStFourSevenSixNine{} under Muon.
With separate routers per visit (experts revisited), this cosine is \cosExpertsMinMseven{} to \cosExpertsMaxMseven{}, nearly orthogonal.
Adding the two gradients does not reduce noise.
The summed gradient's signal-to-noise ratio exceeds both single-visit ratios in \nSnrAbove{} of 70 combinations of model, optimizer, checkpoint and revisited component, never by more than 5\%.
The median ratio of summed to single-visit signal-to-noise is \snrRatioMedian{}.

\begin{figure}[t]
\fig{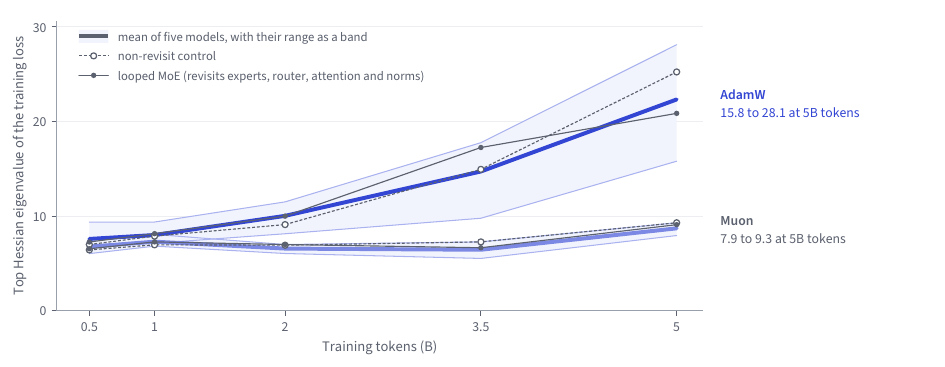}{Line chart of the top Hessian eigenvalue against training tokens (0.5B, 1B, 2B, 3.5B, 5B), with one band per optimizer that spans five models (MoE-12, the non-revisit control, experts revisited, experts and router revisited, the looped MoE), a thick line for their mean, a dashed line with rings for the non-revisit control and a thin line with dots for the looped MoE. Under AdamW the band rises from \lamMinAdamWStFourSevenSeven{} to \lamMaxAdamWStFourSevenSeven{} at 0.5B tokens to \lamMinAdamWStFourSevenSixNine{} to \lamMaxAdamWStFourSevenSixNine{} at 5B. Under Muon the band stays between \lamMinAllMuon{} and \lamMaxAllMuon{} at every checkpoint.}
\caption{\lead{Curvature rises under AdamW and stays in one range under Muon, in looped and standard models alike.}
Top Hessian eigenvalue of training loss (20 power iterations on 64 validation sequences of 4096 tokens).
Each band spans five models (\MoEtw{}, the \Ctrl{}, experts revisited, experts and router revisited and the looped MoE), with a thick mean line.
Thin lines mark the \Ctrl{} (dashed, rings) and the looped MoE (dots).}
\label{fig:obs-curvature}
\end{figure}

\term{Top Hessian eigenvalue}
Training-loss curvature rises during training under AdamW and stays in one range under Muon, in all five models (\cref{fig:obs-curvature}).
Under Muon, the top eigenvalue stays between \lamMinAllMuon{} and \lamMaxAllMuon{} across all five models and checkpoints.
Under AdamW, it starts between \lamMinAdamWStFourSevenSeven{} and \lamMaxAdamWStFourSevenSeven{} at 0.5B tokens and reaches \lamMinAdamWStFourSevenSixNine{} to \lamMaxAdamWStFourSevenSixNine{} at 5B.
At 0.5B tokens, the four 18-layer models, including the \Ctrl{}, are flatter than \MoEtw{} (\lamMfourAdamWStFourSevenSeven{} to \lamMthreeAdamWStFourSevenSeven{} against \lamMzeroAdamWStFourSevenSeven{} under AdamW, \lamMfiveMuonStFourSevenSeven{} to \lamMsevenMuonStFourSevenSeven{} against \lamMzeroMuonStFourSevenSeven{} under Muon).
By 5B tokens, no ordering between looped and standard models remains.
Whether shared layers are easier to optimize than an equally deep independent stack, as SMELT proposes~\citep{smelt}, remains an open question (\cref{sec:matched}).

\par\penalty0\relax

\section{\texorpdfstring{\moeloop}{LoopTitan}: protocol, records and released models}
\label{sec:system-app}

We release the \moeloopname{} code base, seventeen trained models and the record of every study and run in this paper.

\subsection{One model for every recipe}
\label{sec:system-code}

\begin{figure}[ht]
  \fig{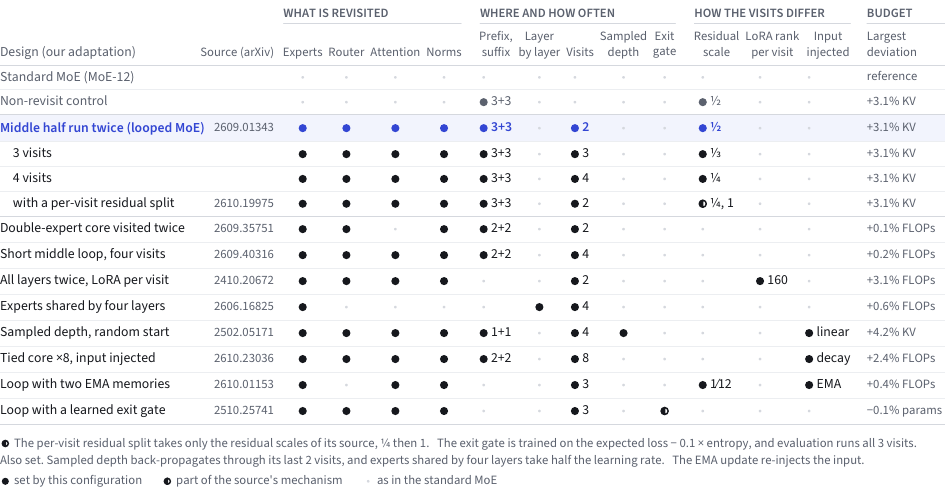}{A dot matrix. Rows: MoE-12 and the non-revisit control for reference, then the twelve configurations of nine published looped recipes, each named by its design with its source in gray, the looped MoE tinted. Columns in three groups: what is revisited (experts, router, attention, norms), where and how often (prefix and suffix, layer by layer, visits, sampled depth, exit gate), how the visits differ (residual scale, LoRA rank per visit, input injected), and a last column with the largest resource deviation. Filled dots mark the keys each configuration sets, and half dots mark that only part of the source's mechanism is used.}
  \caption{\lead{Each published looped recipe is a configuration of one model.}
    Rows show \MoEtw{}, the \Ctrl{} and the twelve configurations of the nine published recipes, named by design as in \cref{fig:base}b with sources in gray (looped MoE tinted).
    Filled dots mark keys changed from \MoEtw{}, and half dots mark partial use of a source mechanism.
    The last column shows the largest of the three resource deviations from \MoEtw{} (every configuration is within 5\% on all three).}
  \label{fig:system-designspace}
\end{figure}

One model class expresses every architecture in this paper, so two recipes differ only in configuration keys (\cref{fig:system-designspace}).

\term{Configurations}
Each weight block is stored once.
Between prefix and suffix, core blocks are visited several times, as a whole core (A\,B, A\,B) or block by block (A\,A, B\,B).
Per visit, a key sets whether attention, router and norms are revisited or stored anew.
Each of the nine published recipes (\cref{sec:results-recipes}) and LatentLoop's residual split is built by a constructor, a function that returns its configuration and records the paper version, implemented sections and every adaptation.
A new recipe is one configuration file, built by a constructor.
Six conformance checks test it, among them that the parameter and call counts behind its computed resources equal those of an executed forward pass and that cached decoding equals recomputation.
Ten of the twelve configurations in \cref{sec:results-recipes} were added in one change without model-code changes.

\term{Resources}
Before training, one \moeloop{} function computes a configuration's stored parameters, training FLOPs and KV cache on CPU.
A second, \code{compare\_recipes}, labels each resource between two configurations as matched, controlled, measured different or unknown (which never counts as matched).
Sweeping a recipe's widths, expert counts and structural settings finds shapes within 5\% of a reference on all three resources (12 of 17,850 in the repository's recorded sweep).
The repository also holds a benchmark protocol under which a new looped design is admitted and ranked.
It admits only submissions within 5\% of their reference on all three resources and ranks only runs that complete their schedule.

\term{Training}
\moeloop{} trains on TorchTitan~\citep{torchtitan} with AdamW or distributed Muon.
The optimizer groups parameters by the block that stores them, so each revisited parameter receives one update per step from its gradient sum across visits.
The model has a plain implementation, and training uses native GPU kernels for attention and experts.
The plain implementation agrees with the official Transformers Qwen3-MoE modules in logits, loss and gradients on the attention settings the runs use.
The native kernels agree with the plain implementation in routing, logits, loss and gradients on the same settings.

\subsection{Released models and how to reproduce a result}
\label{sec:system-app-models}
\label{sec:system-app-repro}

We release seventeen trained models on Hugging Face (\cref{tab:system-app-models}).
They are the six models of the five steps, \MoEloop{} and \MoEtwoE{}, each trained on 5B tokens with AdamW and with Muon, and the \DeepCtrl{} with Muon.
Each is the seed-42 run from \cref{tab:results-5b}, whereas the main text reports three-seed means for the main setting.
The evaluation files of each model reproduce its test BPB in the repository's results file to six decimals (\cref{tab:results-5b} prints four).
Each model is a standalone export that bundles its model code, loads via the Transformers auto classes (\code{trust\_remote\_code=True}) without the repository and decodes with an exact per-visit KV cache.

\begin{table}[ht]
\caption{\lead{Paper, record and Hugging Face names for the seventeen released models.}
  All are seed-42 runs in the \code{NaturalScale} organization.
  Released names encode design, stored parameters, active parameters per token (\code{A}, counting revisited layers at every visit) and optimizer.
  Code \code{U} denotes \MoEei{}, and the \Ctrl{} is released as \code{Unrolled}.}
\label{tab:system-app-models}
\begin{fittabular}{3}{lll}
\toprule
Model & Code in the record & Released model \\
\midrule
\MoEtw & \code{M0} & \code{LoopTitan-MoE12-1.3B-A0.2B-\{AdamW,Muon\}} \\
\MoEtw{} with finer experts & \code{G} & \code{LoopTitan-MoE12-FineExperts-1.3B-A0.2B-\{AdamW,Muon\}} \\
\MoEei & \code{U} & \code{LoopTitan-MoE18-1.3B-A0.2B-AdamW} \\
\oursrow \Base{} (\MoEei{} with Muon) & \code{U} & \code{LoopTitan-Base-1.3B-A0.2B} \\
\MoEei{} with the recipe's expert layout & \code{R} & \code{LoopTitan-MoE18-Layout-1.3B-A0.2B-\{AdamW,Muon\}} \\
\oursrow Non-revisit control & \code{M4} & \code{LoopTitan-Unrolled-1.3B-A0.2B-\{AdamW,Muon\}} \\
Looped MoE & \code{M3} & \code{LoopTitan-Looped-1.3B-A0.2B-\{AdamW,Muon\}} \\
\MoEloop & \code{M0-Loop} & \code{LoopTitan-MoE12-Revisit-1.3B-A0.3B-\{AdamW,Muon\}} \\
\MoEtwoE & \code{M0-2E} & \code{LoopTitan-MoE12-2xExperts-2.6B-A0.2B-\{AdamW,Muon\}} \\
Deeper control & \code{M0-Deep} & \code{LoopTitan-MoE12-Deep-1.3B-A0.3B-Muon} \\
\grouprule
\multicolumn{3}{l}{\textit{Not released (configurations in \cref{tab:setup-app-arch})}}\\
Experts revisited & \code{M7} & \\
Experts and router revisited & \code{M5} & \\
Looped MoE with 3 visits & \code{V3} & \\
Looped MoE with 4 visits & \code{V4} & \\
\MoEtw{} with residual scaling & \code{M0R} & \\
\quad its learning-rate part & \code{M0LR} & an optimizer setting on \MoEtw{} (that table's note) \\
\quad its initialization part & \code{M0I} & an initialization setting on \MoEtw{} (that table's note) \\
MoE-24 & \code{D24} & \\
MoE-36 & \code{D36} & \\
\bottomrule
\end{fittabular}
\tablenote{Codes identify models across run records, configurations and study specs.
In the 3.0B setting, the four models carry the codes of their main-setting counterparts.}
\end{table}

\term{\Base{} and the \Ctrl{}}
\Base{} is the released \MoEei{} trained with Muon, and the \Ctrl{} is released as \code{Unrolled} with either optimizer.
The repository stores both models' frozen configurations, learning rates and results, so a comparison with \Base{} needs no retraining of it.
For a new looped design, a researcher writes its configuration and that of a \Ctrl{} built for it, checks the three resources before training (\cref{sec:system-code}) and trains in the main setting (\cref{tab:setup-app-train}) at the optimizer's learning rate.

\term{Rerunning a study}
The record holds each batch of a study's runs as one GitHub issue, which lists the runs, one command per line, and the commit from which they launch.
To rerun them, check out that commit and install the locked environment.
The first command launches one line and records the run, and the second reads the study's spec, prints the results table and checks the runs against the spec (\cref{sec:system-app-checker}).
\begingroup\csname @beginparpenalty\endcsname=10000
\begin{quote}
\code{python scripts/run.py -\kern0pt-issue N -\kern0pt-gpus K -\kern0pt- <command>}\\
\code{python scripts/results.py -\kern0pt-issue N -\kern0pt-spec <file>}
\end{quote}
\endgroup
A checkpoint holds the model, the optimizer, the scheduler, the random state of every rank and the data cursor, so a run resumes exactly.
In a 16-rank resume test on 22 October 2026, 4 steps plus 4 resumed steps reproduced every model weight and every optimizer state of 8 uninterrupted steps (maximum absolute error 0 across all 103 model and all 309 optimizer tensors).
Under Muon, resumption reproduces the next 32 updates' losses bit for bit.

\subsection{An auditable collaboration with agents}
\label{sec:system-protocol}

\begin{figure}[ht]
  \fig{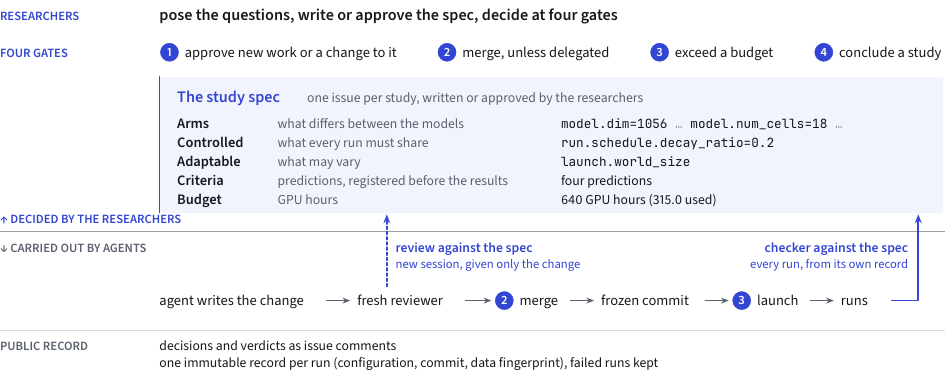}{Diagram in two zones separated by a line. Above the line, decided by the researchers: their row
    (pose the questions, write or approve the spec, decide at four gates), a row of four numbered gates (approve new
    work or a change to it, merge unless delegated, exceed a budget, conclude a study) and a card for the study spec, one
    issue per study, written or approved by the researchers, with its sections Arms, Controlled, Adaptable, Criteria and Budget, each with its
    meaning and a real example line. Below the line, carried out by agents: agent writes the
    change, fresh reviewer, gate 2 merge, frozen commit, gate 3 launch, runs. Two arrows cross the line into the
    spec, a dashed one from the reviewer (review against the spec) and a solid one from the runs (checker against
    the spec, every run from its own record). At the bottom, the public record: decisions and verdicts as issue
    comments, one immutable record per run with failed runs kept.}
  \caption{\lead{Researchers pose the questions, write or approve each spec and decide at four gates.
    Agents work within the spec, with changes and runs recorded and checked against it.}
    Researchers decide above the line (numbered circles mark the four gates) and agents work below it.
    Gate~3 stands at the launch because used GPU hours are summed before each launch, and the researchers decide only when a launch would exceed the budget.
    The dashed arrow marks an agent rule and the solid arrow a code-enforced check.
    Spec lines come from the study of the shape change (\cref{sec:matched}).}
  \label{fig:system-overview}
\end{figure}

Researchers and AI agents conducted this work in the issues of one GitHub repository, whose record is released with this paper (\cref{fig:system-overview}).
The researchers posed the questions and wrote or approved each study's spec.
A spec names the compared models and their sole differences, the settings every run shares (each with a reason) and the settings that may vary (each with a rule).
It also fixes the criteria for an answer and a GPU-hour budget.
Eleven of the \recordStudies{} studies also registered predictions before results.

\term{Four gates}
Four decisions rest with the researchers, who took or delegated each one.
The first approves new work and changes to approved work, including any change to what a study compares, holds fixed or lets vary.
The second merges a change.
An agent may merge only when the issue delegates it, the newest review passes the latest commit and no protected path changes.
Every other merge waits for the researchers.
Configurations, data, evaluation and the rules themselves are protected.
The third approves a batch of runs exceeding its budget, with used GPU hours (including failed runs) summed before each launch.
The fourth concludes a study.
An agent stops at each gate and at any deviation outside the spec, such as a controlled setting unattainable without changing the comparison.
It then returns the question to the researchers with evidence, at most three options and a recommendation.

\term{Agents within the spec}
Within each spec, agents wrote code, reviewed changes and ran experiments.
A guard rejects changes touching paths outside their issue's scope.
Changes are reviewed against the spec in a fresh session that receives only the change, and a checker tests every run's own record (\cref{sec:system-app-checker}).

\term{The record}
Every result in this paper traces to a public record of its study, its runs and the decisions that released them.
Decisions are recorded as issue comments with their reasons, and every run leaves a record.
Each run record is created before waiting for GPUs.
It fixes the commit, working-tree state, resolved configuration and hashes of environment and data, and its parameters cannot be changed afterwards.
Started runs are never stopped to save budget, failed runs remain recorded and runs repeat only after infrastructure failures, never because a result looks wrong.
Failed predictions also stay recorded.

The released record spans \recordStudies{} studies and \recordRuns{} run records over \recordDays{} days, including 31 failed or killed runs.
Its issues and comments are in Chinese.
The protocol, CVSM (Controlled-Variable State Machine), is independent of looped models and public at \cvsmURL{} (Apache-2.0 license).

\subsection{The checker}
\label{sec:system-app-checker}

The checker matches each run to a spec model using the run's own recorded configuration, commit and hashes of environment and data, never an agent's report.
It reports every undeclared difference between runs.
It also reports unmet or unrecorded controlled values, runs matching no model or several and models without a run.
It checks all runs of a study together, so each study is checked as one comparison.

The checker only reports, and agent-applied run rules govern the response.
Before requesting GPUs, the agent checks CPU-generated configurations.
Once the launched runs have recorded their configurations, the agent reruns the check and stops every run that violates a controlled value, omits a controlled key or matches no model or several.
The agent returns undeclared differences to the researchers, first stopping the runs if results could be affected.
Stopped runs stay recorded.

\subsection{One study and its registered predictions}
\label{sec:system-app-trace}

In the study of the shape change (\cref{sec:matched}), every residual-scaling prediction was registered before the runs that test it, and the failed one stays recorded.
The study spec and four predictions were published on 3 November 2026 at 10:28 UTC.
Its eight configurations were merged on the researchers' instruction, and by 13:42 UTC the eight runs were launched and checked against the spec.
All eight finished with none failed, using 315.0 of the 640 budgeted GPU hours, and the checker printed ``no deviation''.
Two of those four predictions concern residual scaling (\cref{sec:residual}), as do three of the five added later (\cref{tab:system-app-predictions}).

\begin{table}[ht]
\caption{\lead{Of the five registered residual-scaling predictions in the study of the shape change, three held, one held in part and one failed.}
  The first two were registered with the study and the last three after its first results, each before the runs that test it (dates in 2026).
  The registered tests use seed 42, and the Measured column also gives each difference with all seeds of its models.
  Signed values are test mBPB differences (negative is a gain).
  The last column judges the difference with all seeds by its optimizer's own seed spread, with two standard errors in parentheses (\cref{sec:results-seeds}).}
\label{tab:system-app-predictions}
\begin{tabularx}{\linewidth}{@{}>{\RaggedRight\arraybackslash}X l >{\RaggedRight\arraybackslash}p{45mm} l l@{}}
\toprule
Prediction, with its threshold & Registered & Measured & As registered & By seed spread \\
\midrule
Residual scaling on \MoEtw{} lowers the loss by more than 2\mbpb{} under AdamW & 3 Nov & \gapMzeroRMzeroAdamWSFortyTwo{} at seed 42, \gapMzeroRMzeroAdamW{} over three seeds & held & \verdictGapMzeroRMzeroAdamW{} (\twoSeGapMzeroRMzeroAdamW) \\
\addlinespace
The recipe's residual-scaling step is the largest of the four steps from \MoEtw{} to the \Ctrl{} under AdamW, and its gain is at least 3\mbpb{} smaller under Muon & 3 Nov & Second largest at seed 42, after finer experts (\gapMfourRAdamWSFortyTwo{} against \gapGMzeroAdamW). Under Muon the gain is \shrinkMfourRAbsTwoDec{} smaller at seed 42 and \seedGapResidualStepMuonAdamWAbs{} smaller with all seeds & held in part & \seedVerdictResidualStepMuonAdamW{} (\seedTwoSeResidualStepMuonAdamW) \\
\addlinespace
The learning-rate part alone lowers the loss by at least 2\mbpb{} under AdamW and stays within 2\mbpb{} of residual scaling & 4 Nov & \gapMzeroLRMzeroAdamW, and \gapMzeroLRMzeroRAdamW{} from residual scaling, at seed 42, \seedDiffMzeroLRMzeroAdamW{} against \MoEtw{}'s three seeds & held & \seedVerdictMzeroLRMzeroAdamW{} (\seedTwoSeMzeroLRMzeroAdamW) \\
\addlinespace
The initialization part alone stays within 2\mbpb{} of \MoEtw{} under AdamW & 4 Nov & \gapMzeroIMzeroAdamW{} at seed 42, \seedDiffMzeroIMzeroAdamW{} against \MoEtw{}'s three seeds & \textbf{failed} & \seedVerdictMzeroIMzeroAdamW{} (\seedTwoSeMzeroIMzeroAdamW) \\
\addlinespace
Each part alone stays within 2\mbpb{} of \MoEtw{} under Muon & 13 Nov & \gapMzeroLRMzeroMuon{} (learning-rate part) and \gapMzeroIMzeroMuon{} (initialization part) at seed 42, \seedDiffMzeroLRMzeroMuon{} and \seedDiffMzeroIMzeroMuon{} against \MoEtw{}'s three seeds & held & both \seedVerdictMzeroLRMzeroMuon{} (\seedTwoSeMzeroLRMzeroMuon) \\
\bottomrule
\end{tabularx}
\tablenote{The second row's step goes from \MoEei{} with the recipe's expert layout to the \Ctrl{}.
Over three seeds each, the step lowers the loss by \gapMfourRAdamWAbs\mbpb{} under AdamW, \seedVerdictMfourRAdamW{} (\cref{sec:residual}).
Under Muon, \MoEei{} with the recipe's expert layout has seed 42 only.
\Cref{tab:settings} merges this step with the layout step, so its column shows other differences.
The third row's verdict judges the difference from \MoEtw{}.}
\end{table}

\term{Registered thresholds}
The study's criteria fixed these thresholds before its runs, at 2\mbpb{} between two models and $\sqrt{2}\times 2\approx 3$\mbpb{} between optimizers.
Both came from the seed spread of an earlier code base and configuration.
Our seed runs later measured the spread of the current code, and the paper judges every difference by its own optimizer's spread (\cref{sec:results-seeds}).
Under that rule, the second row's difference between optimizers is beyond noise, and every other difference in the last column of \cref{tab:system-app-predictions} is within noise.
The first prediction held as registered at seed 42, yet over three seeds its difference is \gapMzeroRMzeroAdamWAbs\mbpb{} and \verdictGapMzeroRMzeroAdamW{} (\cref{sec:results-residual}).

\subsection{Sources of the capability matrix}
\label{sec:system-app-public}

For the capability matrix (\cref{fig:system-capabilities}), we surveyed sixteen related papers on 11 November 2026 and read each public repository at the commit in \cref{tab:system-app-repos}.

\begin{table}[!ht]
\caption{\lead{Repositories and commits of the ten public code bases in the capability matrix.}
  MoEUT is graded across its model and training repositories, and Loop the Loopies!\ on its Megatron-LM fork with its Transformers and vLLM forks for the export column.}
\label{tab:system-app-repos}
\begin{tabularx}{\linewidth}{@{}l >{\RaggedRight\arraybackslash}X@{}}
\toprule
Code base & Repository at commit \\
\midrule
Huginn~\citep{huginn} & \code{seal-rg/recurrent-pretraining@1ea7220} \\
Mixture-of-Recursions~\citep{mor} & \code{raymin0223/mixture\_of\_recursions@82c97f2} \\
MoEUT~\citep{moeut} & \code{RobertCsordas/moeut@87f0497} \\
 & \code{RobertCsordas/moeut\_training\_code@521093a} \\
Foil~\citep{foil} & \code{SR-A-W/how-to-loop-moe@0f9cc73} \\
Expert Tying~\citep{tying} & \code{epfml/looped-moe@833fa17} \\
LOOM~\citep{loom} & \code{hed-ucas/LOOM@a2ee5a7} \\
Sandwich~\citep{sandwich} & \code{ifm-ai/xllm-loop@755c7b3} \\
LatentLoop~\citep{latentloop} & \code{lcqysl/LatentLoop@deef911} \\
Sparse Layers~\citep{sparselayers} & \code{ml-ryanlee/looped-moe-scaling@426f632} \\
Loop the Loopies!~\citep{loopie} & \code{zitian-gao/Megatron-LM@cb2daa2} \\
 & \code{zitian-gao/transformers@90ce33b} \\
 & \code{zitian-gao/vllm@9a9703d} \\
\bottomrule
\end{tabularx}
\end{table}

A mark is \emph{yes} when the capability is present, \emph{partly} when something narrower is present, \emph{no} when the README, file tree and keyword searches find nothing, and \emph{unclear} when public code is inconclusive.
Marks grade public code on that date, not the papers or their experiments.
Searches of arXiv, GitHub, the web, two curated lists and the Hugging Face Hub found no official training code for the other six papers, Ouro~\citep{ouro} (public weights and inference file), Relaxed Recursive Transformers~\citep{relaxed}, SMELT~\citep{smelt}, scaling laws for looped MoE~\citep{metaloop}, LoopMoE~\citep{loopmoe} and ELR~\citep{elr}.

\end{document}